\documentclass[12pt]{article}
\usepackage{amsmath}
\usepackage{graphicx}
\usepackage{enumerate}
\usepackage{natbib}
\usepackage{url} % not crucial - just used below for the URL 

\newcommand{\blind}{1}

\usepackage[utf8]{inputenc}
\usepackage{amsfonts}
\usepackage{amsgen,amsmath,amstext,amsbsy,amsopn,amssymb,subcaption,mathrsfs}
\usepackage{comment}
\usepackage{bbm}
\usepackage{times}

\usepackage[shortlabels]{enumitem}
\usepackage{natbib}
\usepackage{booktabs}
\usepackage{amsthm}
\usepackage{hyperref}
\usepackage{blkarray}
\usepackage{algorithm}
\usepackage{algorithmic}
\usepackage{url}
\usepackage{longtable}
\usepackage{stmaryrd}
\usepackage{mwe}
\usepackage{mathtools}
\DeclareMathAlphabet\mathbfcal{OMS}{cmsy}{b}{n}

\makeatletter
\newcommand*{\rom}[1]{\expandafter\@slowromancap\romannumeral #1@}
\makeatother

\usepackage[dvipsnames]{xcolor}
\newcommand{\red}[1]{{\color{red}#1}}

\newcommand{\inp}[2]{\langle #1,#2\rangle}

\renewcommand{\baselinestretch}{1}
\newcommand{\PP}{\mathbb{P}}

\def\uoff{\textsf{u-off}}
\def\uipw{\textsf{IPW}}
\def\uaw{\textsf{AW}}

\def\calX{{\mathcal X}}

\def\EE{{\mathbb E}}

\def\PP{{\mathbb P}}

\def\RR{{\mathbb R}}

\def\indicator{\mathbf{1}}
\def\hat{\widehat}
\def\eps{\varepsilon}

\def\uoff{\textsf{u-off}}
\def\uipw{\textsf{IPW}}
\def\uaw{\textsf{AW}}

\newtheorem{Theorem}{Theorem}
\newtheorem{Assumption}{Assumption}

\newtheorem{Lemma}{Lemma}
\newtheorem{Remark}{Remark}
\theoremstyle{plain}
\newtheorem{Corollary}{Corollary}

\begin{document}

\def\spacingset#1{\renewcommand{\baselinestretch}%
{#1}\small\normalsize} \spacingset{1}

%%%%%%%%%%%%%%%%%%%%%%%%%%%%%%%%%%%%%%%%%%%%%%%%%%%%%%%%%%%%%%%%%%%%%%%%%%%%%%

\if1\blind
{
  \title{\bf Regret Minimization and Statistical Inference in Online Decision Making with High-dimensional Covariates}
  \author{Congyuan Duan\\\bigskip 
   Department of Mathematics, Hong Kong University of Science and Technology\\  
     Wanteng Ma \\\bigskip
    Department of Statistics and Data Science, University of Pennsylvania\\
     Jiashuo Jiang$^\ast$ \\\bigskip
    Department of Industrial Engineering and Decision Analytics,\\\bigskip Hong Kong University of Science and Technology\\
    Dong Xia\footnote{Co-corresponding authors.} \\\bigskip
    Department of Mathematics, Hong Kong University of Science and Technology
    }
  \maketitle
} \fi

\if0\blind
{
  \bigskip
  \bigskip
  \bigskip
  \begin{center}
    {\LARGE\bf Regret Minimization and Statistical Inference in Online Decision
Making with High-dimensional Covariates}
\end{center}
  \medskip
} \fi

\bigskip
\begin{abstract}
This paper investigates regret minimization, statistical inference, and their interplay in high-dimensional online decision-making based on the sparse linear contextual bandit model. We integrate the $\varepsilon$-greedy bandit algorithm for decision-making with a hard thresholding algorithm for estimating sparse bandit parameters and introduce an inference framework based on a debiasing method using inverse propensity weighting. Under a margin condition, our method achieves either 
$O(\sqrt{T})$ regret or classical $O(\sqrt{T})$-consistent inference, indicating an unavoidable trade-off between exploration and exploitation. If a diverse covariate condition holds, we demonstrate that a pure-greedy bandit algorithm—i.e., exploration-free—combined with a debiased estimator based on average weighting can simultaneously achieve optimal 
$O(\log T)$ regret and $O(\sqrt{T})$-consistent inference. We also show that a simple sample mean estimator can provide valid inference for the optimal policy's value. Numerical simulations and experiments on Warfarin dosing data validate the effectiveness of our methods.
\end{abstract}

\noindent%
{\it Keywords:} linear contextual bandit, sparsity, $\eps$-greedy, inference, regret  
\vfill

\newpage
\spacingset{1.68} % DON'T change the spacing!

\section{Introduction}\label{sec:intro}

In today's digital age, the vast amount of user-specific data generated by online platforms and services has created unprecedented opportunities for personalized online decision-making. This approach involves designing policies based on previously collected data and making the best personalized decisions and recommendations for each future individual based on their unique features (i.e., covariates). Compared to implementing general actions for the entire population, personalization allows decision-makers to improve overall prediction accuracy. As a result of these benefits, personalized online decision-making has found a wide range of applications in clinical trials and healthcare services (\citep{murphy2003optimal, bertsimas2017personalized}), online advertising and marketing (\citep{bottou2013counterfactual, bertsimas2020predictive}), revenue management (\citep{chen2022statistical}), and online news recommendation (\citep{li2010contextual, ban2019big}).

In personalized online decision-making, covariates can be ultrahigh-dimensional when platforms collect a large number of personalized features. For instance, an e-commerce platform may collect a user’s identity information, browsing history, and purchase history, which together constitute a high-dimensional feature space that characterizes each user’s preferences. % Consequently, the platform can recommend different products to different users based on their personalized information. 
Similarly, in healthcare services, electronic medical records consist of a patient’s medical history, genetic information, and biomarkers, enabling doctors to optimize treatment strategies for each patient based on these medical characteristics. Specially, high-dimensionality refers to the case when the dimension of covariates $d$ is comparable to or substantially larger than the time horizon of the experiment, denoted by $T$.

\subsection{High-dimensional Contextual Bandit}

Driven by the proliferation of high-dimensional personalized data and the increasing demand for online personalized decision-making, high-dimensional sparse linear contextual bandits (LCBs) have emerged as prominent models to address these challenges. Formally, the problem is defined as follows: a decision-maker is presented with a finite set of available actions, commonly referred to as \emph{arms}. At each time step, an individual arrives bearing a set of covariates, known as \emph{contextual information}. The decision-maker selects one arm and subsequently receives a stochastic \emph{reward}. Specifically, the expected reward is modeled as a \emph{sparse} linear function of the covariates, meaning that only $s_0$ out of $d$ covariates contribute to the expected reward. Typically, the sparsity level $s_0$ is substantially smaller than the ambient dimension $d$, underscoring the high-dimensional nature of the problem.

Various sparse-LCB algorithms have been developed, and their performances are commonly evaluated using upper bounds on  \emph{regret}, defined as the difference in total expected reward between an online decision-making algorithm and an oracle policy. Existing works typically achieve ideal regret upper bounds of $O(\log T)$ or $O(\sqrt{T})$  under certain conditions (\cite{wang2018minimax, kim2019doubly, bastani2020online, oh2021sparsity, ren2024dynamic}). Although these studies have fruitfully focused on regret analysis and algorithm efficiency, uncertainty quantification, statistical inference of bandit parameters, and the evaluation of optimal policy values remain largely underexplored in the context of sparse-LCB. While regret analysis assesses the overall performance of a bandit algorithm in selecting the correct actions, it is often necessary for decision-makers to assign a confidence level to each recommended action and to quantify the uncertainty in both parameter and policy value estimates. This task is particularly crucial in variable selection when covariates are high-dimensional. 
Common upper bounds on the estimation error fail to provide a precise characterization of these uncertainties, a problem that becomes even more challenging when sparse estimators are employed for policy evaluation. Therefore, it is imperative to answer the following questions: (i) \emph{How confident are we that variables with small coefficients are truly irrelevant?}; (ii)\emph{How confident are we that a low-regret algorithm achieves the expected optimal reward?}; (iii) \emph{How confident are we that one action is better than another?} Addressing these questions requires a distributional characterization of the estimated bandit parameters and empirical rewards, which, to the best of our knowledge, has not been previously explored.

Quantifying the uncertainty of an estimator requires careful analysis of its bias, variance, and asymptotic distribution. Indeed, conducting statistical inference for high-dimensional bandit parameters is particularly challenging and far from straightforward. First, data in bandit algorithms is collected adaptively, which can introduce implicit bias and may lead to a substantial underestimation of the final estimator’s variance if the inherent dependence is not properly addressed. Moreover, adaptively collected data render crucial technical tools, such as the central limit theorem, inapplicable. Second, high-dimensional data poses significant challenges for parameter estimation and inference because the estimators typically lack explicit analytical forms and must be computed via iterative algorithms. The implicit or explicit regularization applied to these estimators introduces additional bias that is difficult to characterize analytically..

Existing literature on the inference of contextual bandits primarily focuses on low-dimensional setting (\cite{bibaut2021post, chen2021statistical, zhan2021off, zhang2021statistical, chen2022online, shen2024doubly}). For example, \cite{chen2021statistical} investigated the $\varepsilon$-greedy algorithm and proposed an online weighted OLS with inverse propensity weight (IPW) for bias correction, establishing its asymptotic normality. The follow-up work \cite{shen2024doubly} developed a doubly robust method to infer low-dimensional parameters and optimal policy values. In these studies, estimators for bandit parameters and policy values, such as OLS estimates, often possess explicit analytical forms, making it relatively straightforward to derive their estimation errors and asymptotic distributions. However, these methods cannot be directly generalized to the high-dimensional LCB setting due to the lack of explicit forms and the unknown asymptotic behavior of the sparse estimators used in bandit algorithms.

Another line of research primarily focuses on minimizing regret in sparse-LCB algorithms by employing regularization techniques such as LASSO. Specifically, \cite{bastani2020online} introduced an efficient forced-sampling bandit algorithm based on the LASSO estimator, establishing a regret upper bound of $O\left(s_0^2(\log T+\log d)^2\right)$.  \cite{wang2018minimax} followed the same forced-sampling approach but incorporated the minimax concave penalty estimator (\cite{zhang2010nearly}), which improved the regret bound to $O\left(s_0^2(\log d+s_0)\log T\right)$. Additionally, \cite{kim2019doubly} developed the doubly robust LASSO bandit algorithm by integrating techniques from the missing data literature, achieving a regret of $O(s_0\log(dT)\sqrt{T})$ without assuming arm separability conditions.  While these methods effectively control the regret in high-dimensional LCB, their sparse estimators do not provide asymptotic distributional guarantees. To enable statistical inference for high-dimensional LCB, it is necessary to develop novel algorithms that simultaneously achieve nontrivial regret bounds and facilitate tractable asymptotic analysis of the sparse estimators. Interestingly, we found that a smart integration of online iterative hard thresholding (IHT) and  $\eps$-greedy algorithms offers a promising solution.

%Conducting effective uncertainty quantification for high dimensional bandit parameter is, however, far from straightforward. Valid statistical inference necessitates an accurate understanding of the bias, variance, and asymptotic distribution of bandit parameter estimators.  A non-zero \emph{bias} of the estimator indicates a non-negligible difference between the expectation of the estimator and the true effect of the factor, which would not disappear asymptotically. This bias can result in incorrect analyses, such as exaggerating, underestimating, or even reversing the true effect of a specific factor. Meanwhile, the \emph{variance} of the estimator impacts the sensitivity of evaluations. For instance, high variance results in wider confidence intervals and less significant $p$-values in hypothesis tests, causing significant factors fail to be recognized. Even in low-dimensional contextual bandit settings, characterizing the asymptotic distribution of estimators is challenging due to the non-i.i.d. nature of observed data resulting from adaptive bandit policies, which introduces additional bias and nontrivial variance. Moreover, the non-i.i.d. data renders the standard central limit theorem inapplicable, necessitating a more careful analysis of asymptotic variance. Furthermore, high dimensional data introduces additional technical challenges since the regularization method for sparse linear model such as LASSO imposes additional bias and sparse estimators lack a tractable limiting distribution.  

More importantly, balancing the trade-off between regret performance and inference efficiency, as characterized by the order of asymptotic variance, presents significant challenges. In fact, these two objectives inherently conflict with each other. Efficient inference typically requires a sufficiently large sample size for each arm. For instance, implementing random exploration or forced sampling can ensure that a constant fraction of the total samples is allocated to each arm, resulting in an effective sample size of $O(T)$ per arm and enabling optimal estimates of each arm's parameters. However, these approaches lead to a trivial regret performance of $O(T)$. In contrast, most bandit algorithms prioritize minimizing overall regret, often at the expense of the effective sample size for suboptimal arms. By selecting suboptimal arms with vanishing probability, these algorithms can reduce the effective sample size for such arms, resulting in suboptimal estimates of their parameters. A similar trade-off is highlighted in \cite{simchi2023multi}, which demonstrates that the product of the estimator's variance and the regret is of constant order in the worst case. \emph{Is it possible to design a learning policy and inference strategy that on one hand achieves the optimal regret, whereas on the other hand enjoys optimal statistical inference efficiency?} Motivated by this question, we summarize our contribution in Section \ref{sec:contribution}.

\subsection{Main Contributions} \label{sec:contribution}
To the best of our knowledge, this work is the first to investigate regret minimization, statistical inference, and their interplay in high-dimensional online decision-making based on the sparse-LCB model. Our contributions are summarized as follows:  \\
\textit{General Inference Framework and Tradeoff with Regret}. We propose a novel statistical inference framework for adaptively collected high-dimensional data. Our approach integrates the $\varepsilon$-greedy bandit algorithm with hard-thresholding (HT), resulting in a biased estimator due to the adaptive data collection and implicit regularization introduced by the HT algorithm. To mitigate this bias, we introduce an online debiasing technique based on IPW that maintains low computational and storage complexity. Under a margin condition with parameter $\nu$, the debiased estimator is asymptotically normal, enabling the construction of confidence intervals and hypothesis tests for both individual arm parameters and their differences. Additionally, we identify a trade-off between regret performance and the estimator's asymptotic variance, which affects inference efficiency by determining the width of confidence intervals and the p-values of test statistics. Specifically, when the algorithm achieves a regret upper bound of $O(T^{1-\gamma} + T^{(\gamma-1)(1+\nu)/2})$ with margin parameter $\nu$, and some user-specified $\gamma \in [0, 1)$—which characterizes the exploration probability, the estimator's asymptotic variance is $O(T^{-(1-\gamma)})$. For example, when $\nu=1$, setting $\gamma = \frac{1}{2}+o(1)$ yields a regret bound of $O(T^{1/2})$ and an estimator variance of $O(T^{-1/2})$, which does not attain the classic $\sqrt{T}$-consistency; setting $\gamma=0$ yields a trivial regret bound of $O(T)$ and an asymptotically normal estimator which is $\sqrt{T}$-consistent. While IPW is effective for debiasing, it unfortunately inflates the variance of the final estimator. 
\begin{comment}
    \red{(delete?)} We also emphasize that, despite its simplicity, our $\varepsilon$-greedy-based algorithm provides two notable advantages that are particularly valuable in this setting. First, the known selection probabilities under this framework facilitate reliable statistical inference for bandit parameters through IPW. Second, the exploration parameter $\gamma$ directly characterizes and balances the inherent trade-off between regret and inference efficiency. 
\end{comment}

\noindent\textit{Simultaneous Optimal Inference and Regret}. We demonstrate that optimal inference efficiency and regret performance can be simultaneously achieved under an additional \emph{covariate diversity} assumption, commonly employed in high-dimensional bandit literature (\cite{bastani2021mostly, ren2024dynamic} and references therein). This assumption is motivated by the observation that when covariates are sufficiently diverse, an exploration-free algorithm (i.e., setting the exploration probability $\varepsilon=0$ in the $\varepsilon$-greedy algorithm) can still adequately explore each arm. This automatic exploration facilitates debiasing through a simple average weighting (AW) approach, bypassing IPW and thereby avoiding variance inflation. Specifically, our approach achieves an optimal $O(\log T)$ regret upper bound, and the resulting estimators of arm parameters are asymptotically normal with a variance of $O(T^{-1})$, thereby attaining the classic $\sqrt{T}$-consistency and optimal inference efficiency. Additionally, we introduce an inference procedure for the optimal policy's value, often referred to as the Q-value, within this framework. We provide a straightforward method to assess the maximum total reward achievable by the optimal policy.\\
\textit{Empirical Result}. We evaluate the empirical performance of our algorithm and inference framework through numerical simulations and a real-world data experiment. Specifically, we apply this framework to the aforementioned Warfarin dosing problem. Our approach identifies several significant variables that determine the appropriate dosage, with findings that are consistent with existing medical literature while also offering novel insights.

\subsection{Other Related Works}
Our work is closely related to several fields. We provide a comprehensive review of the related literature and a comparison with our results.

\textit{High dimensional linear regression and statistical inference}. High-dimensional linear regression and statistical inference have been extensively studied. Pioneering works \cite{javanmard2014confidence, javanmard2014hypothesis, f29b4853-f902-378f-99ba-4ec108f6737e, zhang2014confidence,10.1214/17-AOS1630} developed debiased LASSO estimators, enabling the construction of confidence intervals for individual regression coefficients with nearly optimal widths and testing power. \cite{belloni2014high} introduced a two-stage approach combining post-double-selection with LSE. Additionally, \cite{liu2013asymptotic, liu2020bootstrap} explored bootstrapping techniques for modified LSE and LASSO with a partial ridge estimator, respectively. While these studies primarily address i.i.d. offline data, there is recent research on advanced inference methods for online and adaptively collected data. \cite{10.1214/18-AOS1801} integrated the Regularization Annealed epoch Dual Averaging algorithm (RADAR; \cite{agarwal2012stochastic}), a variant of stochastic gradient descent (SGD), with debiased techniques. \cite{shi2021statistical} proposed a recursive algorithm for model selection that reduces dimensionality and constructs score equations based on selected variables. \cite{deshpande2023online} examined online debiasing LASSO for time series and batched data. Recently, \cite{han2024online} developed debiased SGD and RADAR as one-pass algorithms that further reduce the space complexity of previous methods. However, the above inference procedures do not account for reward and regret control, which is critical in online decision making.

%\textit{Statistical inference in contextual bandit}. Existing literature on the inference of contextual bandits primarily focuses on standard low-dimensional setting (\cite{bibaut2021post, chen2021statistical, zhan2021off, zhang2021statistical, chen2022online, shen2024doubly}). For example, \cite{chen2021statistical} investigated the $\varepsilon$-greedy algorithm and proposed an online weighted LSE with IPW for bias correction, establishing its asymptotic normality. The follow-up work \cite{shen2024doubly} developed a doubly robust method to infer low-dimensional parameters and optimal policy values. However, their studies are unable to be directly generalized to high-dimensional problems due to inefficient regret control of simple exploration algorithms in \cite{chen2021statistical}, \cite{shen2024doubly},  and the unknown asymptotic behavior of sparse estimators in bandits update.

%Additionally, \cite{han2022online} and \cite{duan2024online} studied inference for low-rank matrix contextual bandits, where the covariate information is given. However, in the absence of covariate information,  \cite{han2022online} and \cite{duan2024online} fail to infer parameters even in vector cases.

\textit{Variance stabilization for adaptive experiment}. Inference for adaptively collected data is challenging due to potential bias in estimators. In causal inference and economics, IPW technique (\cite{imbens2004nonparametric, imbens2009recent}) is commonly used for bias correction. However, IPW often inflates variance because of diminishing propensities. To stabilize variance, two strategies have been proposed: shrinking importance weights to balance the bias-variance tradeoff (\cite{bottou2013counterfactual, wang2017optimal, su2020doubly}) and constructing adaptive weights for IPW estimators (\cite{luedtke2016statistical, bibaut2021post, hadad2021confidence, zhan2021off, zhang2021statistical, chen2022online}). Unfortunately, these methods are unsuitable for sparse-LCB, as the randomness from adaptive weights can introduce significant bias. By contrast, we propose an exploration-free approach that bypasses IPW, thereby stabilizing variance and achieving optimal inference efficiency.

%\textit{A/B testing}. A/B testing is a widely used technique for the online and adaptive evaluation of strategies or services. It involves assigning users to control or treatment groups, collecting outcomes, and performing statistical inference on the expected differences between groups. The literature primarily focuses on designing estimators for the average treatment effect with minimal bias and variance, as well as characterizing their asymptotic distributions to enable efficient and robust inference. A growing body of work has developed various A/B testing methods (\cite{tang2010overlapping, johari2017peeking, yang2017framework, kohavi2020trustworthy, shi2023dynamic, wu2024nonstationary}). However, A/B testing generally does not emphasize policy learning algorithms and regret performance, thereby avoiding the need to address the trade-off between regret performance and inference efficiency.

\section{Algorithm Framework with Regret Performance}
%We first formulate the two-arm sparse-LCB problem. Next, we present the $\varepsilon$-greedy bandit and hard-thresholding algorithm proposed by \cite{ma2024high} for high-dimensional bandits.

\subsection{Sparse Linear Contextual Bandit}
We first introduce the notations used throughout this paper. Let $\|\cdot\|$ and $\|\cdot\|_{\ell 1}$ denote the $\ell_2$ and $\ell_1$ norms for vectors, respectively. The matrix infinity norm $\|\cdot\|_{\infty}$ represents the maximum of the absolute row sums for matrices. The $\ell_0$ norm, $\|\cdot\|_{0}$, counts the number of nonzero elements, and the max-norm, $\|\cdot\|_{\max}$, is defined as the maximum absolute entry of a matrix or vector. For a vector $a \in \mathbb{R}^d$, $a_{(i)}$ denotes its $i$th entry for $i \in [d]$, and for a matrix $A \in \mathbb{R}^{m \times n}$, $A_{(ij)}$ denotes its $(i,j)$ entry for $(i,j)\in [m]\times [n]$. Let $I_d$ represent the $d \times d$ identity matrix. For a square matrix $\Sigma$, $\lambda_{\min}(\Sigma)$ and $\lambda_{\max}(\Sigma)$ denote its smallest and largest eigenvalues, while $\sigma_{\min}(\Sigma)$ and $\sigma_{\max}(\Sigma)$ denote its smallest and largest singular values. We use $\indicator(\cdot)$ for the indicator function and $\lesssim$ to indicate inequalities up to a constant factor.

We focus on two-arm bandit for simplicity, though our framework easily extends to $K$-arm bandits. At each time step $t \in [T]$, we observe a contextual covariate $X_t \in \mathcal{X}$ drawn independently from the distribution $\mathcal{P}_{\mathcal{X}}$, with potentially high dimensionality, i.e., $d \gg T$. An action $a_t \in \{0, 1\}$ is selected by an online policy based on the current context $X_t$ and historical data. Upon taking action $a_t$, we receive a noisy reward $y_t$ modeled linearly as
\begin{align}\label{eq:reward}
y_t = \langle \beta_{a_t}, X_t \rangle + \xi_t,
\end{align}
where $\beta_1$ and $\beta_0$ are sparse parameter vectors with $\max\{\|\beta_1\|_0, \|\beta_0\|_0\} \leq s_0 \ll d$. The noise terms $\xi_t$ are independent, mean-zero sub-Gaussian with variances $\sigma_i^2$ conditioned on $a_t = i$ for $i \in \{0, 1\}$. We define a filtration $\{\mathcal{F}_t\}_{t=0}^T$, where $\mathcal{F}_t: = \sigma(X_\tau, a_\tau, y_\tau; 1 \leq \tau \leq t)$ consists of all information up to time $t$, and $\mathcal{F}_0 = \emptyset$. The observations $(X_t, a_t, y_t), t\in[T]$ are neither independent nor identically distributed, as each depends on the past history $\{(X_\tau, a_\tau, y_\tau)\}_{\tau=1}^{t-1}$.

Our primary tasks—regret performance and statistical inference, are characterized as follows. Consider real-world applications such as evaluating the treatment effect of a Warfarin dosage plan, a company's marketing strategy profit, or the reduction in losses from ticket cancellations. Regret of an online algorithm is defined as the expected difference between its achieved total reward and the optimal one:
\begin{align}
R_T = \EE\left[\sum_{t=1}^{T} \max_{i \in \{0,1\}} \langle \beta_i, X_t \rangle - \langle \beta_{a_t}, X_t \rangle\right].
\label{eq:regret}
\end{align}
To assess whether a specific variable, such as the first variable with the associated arm parameters $\beta_{1(1)}$ and $\beta_{0(1)}$, affects the choice between actions, we can conduct the following test:
\begin{align}\label{eq:test-H3}
H_0: \beta_{1(1)} - \beta_{0(1)} = 0 \quad \text{vs.} \quad H_1: \beta_{1(1)} - \beta_{0(1)} \neq 0.
\end{align}
In subsequent sections, we develop a computationally efficient decision-making algorithm, derive an upper bound for its regret, construct estimators for $\beta_1$ and $\beta_0$, establish their central limit theorems, and explore the relationship between regret and inference efficiency.

\subsection{\texorpdfstring{$\varepsilon$}{epsilon}-Greedy Bandit with Hard Thresholding}
The primary objective is to develop online estimators for $\beta_1$ and $\beta_0$, denoted as $\widehat{\beta}_{1,t}$ and $\widehat{\beta}_{0,t}$, respectively, during the learning process. They serve as the foundation of decision making and statistical inference. Under the $\ell_0$-norm constraint and action $a_t = i$, the population LSE at time $t$ is
\begin{align}
\min_{\|\beta\|_0 \leq s_0} \EE\left[(y_t - X_t^\top \beta)^2 \mid a_t = i \right] = \|\beta - \beta_i\|_{\Sigma}^2 + \sigma_i^2, \quad \forall i \in \{0,1\},
\label{eq:optimizationbeta}
\end{align}
where $\Sigma:={\rm Cov}(X_t), \forall t\in[T]$. Convex relaxation using LASSO is commonly employed to solve the sample LSE. However, LASSO's statistical properties rely on a sample of data and necessitate solving a non-smooth convex optimization at each step, which is computationally intensive. Instead, we adopt a fully online, non-convex approach called hard thresholding (HT) to estimate the sparse arm parameters. Compared to online LASSO, online HT is computationally faster and allows for more flexible debiasing methods for statistical inference. 

The detailed steps are presented in Algorithm~\ref{alg:onlineHT}. We define the sparse projection operator $\mathcal{H}_s(x)$ as the hard-thresholding operator that retains the $s$ largest (in absolute value) entries of $x$ and zeros the rest. Our learning strategy employs an $\varepsilon$-greedy policy, a classical and straightforward method for online decision-making in bandit problems (\cite{sutton2018reinforcement}). Given a non-increasing and non-negative sequence $\{\varepsilon_t\}_{t=1}^{T}$ representing exploration probabilities, at each time step $t$, we select the action with the higher estimated reward with probability $1 - \varepsilon_t/2$ (exploitation) and choose the alternative arm with probability $\varepsilon_t/2$ (exploration). Notably, the probability $\pi_t: = \PP(a_t = 1 \mid X_t, \mathcal{F}_{t-1})$ is influenced by the history due to the adaptive data collection process. In Algorithm \ref{alg:onlineHT}, the inverse propensities $(\pi_t)^{-1}$ and $(1 - \pi_t)^{-1}$ serve to adjust the empirical covariance matrix and the gradient at each iteration, ensuring consistent  gradient estimation. Throughout the procedure, we need to maintain the empirical covariance matrix $\widehat{\Sigma}_{i,t}$, which requires $O(d^2)$ storage. However, all updates to $\widehat{\Sigma}_{i,t}$ and the stochastic gradients $g_{i,t}$ can be computed in linear time, resulting in an overall computational complexity of $O(d^2T)$.

\begin{algorithm}[t]
    \caption{$\varepsilon$-greedy policy with online hard thresholding}\label{alg:onlineHT}
\begin{algorithmic}
\STATE{\textbf{Input}: exploration probabilities $\{\varepsilon_t\}_{t\geq 1}$; gradient descent step size $\eta$; sparsity level $s$; \\
\hspace{1.4cm} initialization by $\widehat{\beta}_{1,0}=\widehat{\beta}_{0,0}:=\mathbf{0}$; $\widehat{\Sigma}_{1,0}=\widehat{\Sigma}_{0,0}:=\mathbf{0}_{d\times d}$.}
        \FOR{$t= 1,2,\cdots, T$}  
    \STATE Observe a new request with covariate $X_t$; \\
    \STATE Calculate $\pi_t:=(1-\varepsilon_t)\indicator\big(\inp{\widehat{\beta}_{1,t-1}-\widehat{\beta}_{0,t-1}}{X_t}>0\big)+\frac{\varepsilon_t}{2}$; \\
    \STATE Sample an action $a_t\sim \text{Bernoulli}(\pi_t)$ and observe the reward $y_t$; \\
    
    \FOR{$i=0,1$}
    \IF{$i\pi_t+(1-i)(1-\pi_t)=0$}
    \STATE Treat $\frac{\indicator(a_t=i)}{i\pi_t+(1-i)(1-\pi_t)}=0$ in the following computation;
    \ENDIF  
    \STATE Update the covariance matrix by $\widehat{\Sigma}_{i,t}:=\frac{1}{t}\left((t-1)\widehat{\Sigma}_{i,t-1}+\frac{\indicator(a_t=i)}{i\pi_t+(1-i)(1-\pi_t)}X_tX_t^{\top}\right);$  
    \STATE Update average stochastic gradient: ${g}_{i,t}:=2\widehat{\Sigma}_{i,t}\widehat{\beta}_{i,t-1}-\frac{2}{t}\sum_{\tau=1}^t \frac{\indicator(a_\tau=i)}{i\pi_\tau+(1-i)(1-\pi_\tau)}X_{\tau}y_{\tau};$
    \STATE Gradient descent with hard thresholding: $\widehat{\beta}_{i,t}:=\mathcal{H}_s(\widehat{\beta}_{i,t-1}-\eta {g}_{i,t});$
    \ENDFOR  
    \ENDFOR
    \STATE{\textbf{Output}: $\widehat{\beta}_{0,T}$ and $\widehat{\beta}_{1,T}$.}
\end{algorithmic}
\end{algorithm}

\begin{Remark} In Algorithm \ref{alg:onlineHT}, the sparsity level $s$ should be set to an integer exceeding the true $s_0$. Hard thresholding methods primarily face the challenge of gradient information loss due to truncation, as the hard thresholding operator zeros out minor signals that may contain valuable gradient information for subsequent updates (\cite{murata2018sample, zhou2018efficient}). Additionally, errors from missing gradients can accumulate over iterations, hindering the recovery of a sparse structure. Therefore, selecting a slightly larger sparsity level helps retain more gradient information, and averaging gradients at each step enables a more accurate characterization of the stochastic gradient.
\end{Remark}

\subsection{Convergence and Regret Performance}

In this section, we derive the estimation error rate of the online estimators $\widehat{\beta}_{1,t}$ and $\widehat{\beta}_{0,t}$ and the regret upper bound of Algorithm \ref{alg:onlineHT}. The following conditions are standard and essential in the high-dimensional linear regression and bandit literature. 

\begin{Assumption}
    \label{assump:basic}
    (a) There exists a constant $D$ such that  $\|X_t\|_{\max}\leq D$ \emph{almost surely}. \\
    %(b) The population covariance $\Sigma$ satisfies $\sigma_{\max}(\Sigma)\leq C_{\max}<\infty$ and $\sigma_{\min}(\Sigma)\geq C_{\min}=\Omega(1)$. \\
    (b) For any positive integer $s$, the $2s$-sparse minimal eigenvalue $\phi_{\min}(s)$ and $2s$-sparse maximal eigenvalue $\phi_{\max}(s)$ of the population covariance $\Sigma:=\EE[X_tX_t^{\top}]$ are
    \begin{align*}
        \phi_{\min}(s):=\min_{\beta:\|\beta\|_{\ell 1}\leq \sqrt{2s}\|\beta\|} \frac{\beta^{\top}\Sigma\beta}{\beta^{\top}\beta} \quad and \quad \phi_{\max}(s):=\max_{\beta:\|\beta\|_{\ell 1}\leq \sqrt{2s}\|\beta\|} \frac{\beta^{\top}\Sigma\beta}{\beta^{\top}\beta},
    \end{align*}
    where $0<\phi_{\min}(s)\leq \phi_{\max}(s)<\infty$. \\
    (c)There exists a constant $c>0$ such that for any $2s$-sparse vector $u$,
    \begin{align*}
        \EE[\exp(u^{\top}X_t)]\leq \exp(c\phi_{\max}^2(s)\|u\|^2/2).
    \end{align*}
\end{Assumption}
Assumption \ref{assump:basic}(b), commonly referred to as the sparse minimal eigenvalue condition or the Sparse Riesz Condition, is a widely employed concept in high-dimensional bandit studies (\cite{chakraborty2023thompson, ren2024dynamic}). It is analogous to the restrictive eigenvalue condition introduced in \cite{buhlmann2011statistics, javanmard2014confidence}.

Theorem \ref{thm:estimation-nodc} establishes the convergence of the estimators $\widehat{\beta}_{i,t}$ produced by Algorithm \ref{alg:onlineHT}. Notably, when $\gamma = 0$, the algorithm reduces to the standard online HT algorithm and achieves the well-known statistically optimal, up to a logarithmic factor, error rate of $O\left(\frac{s_0 \sigma_i^2 \log d}{\phi_{\min}^2(s)  t}\right)$ (\citep{ye2010rate, rigollet2011exponential}). Hereafter, we assume $T\leq d^{100}$ for simplicity, where the constant in the exponent can by replaced by any large constants.

\begin{Theorem}
    \label{thm:estimation-nodc}
    Suppose Assumption~\ref{assump:basic} holds.  Let $\varepsilon_t:=c_{\varepsilon}t^{-\gamma}$ for some constant $c_{\varepsilon}$ and $\gamma\in [0,1)$, the step size $\eta:=\big(4\kappa_{*} \phi_{\max}(s)\big)^{-1}$ in Algorithm \ref{alg:onlineHT}, where $\kappa_{*}$ is the condition number $\kappa_{*}:=\phi_{\max}(s)/\phi_{\min}(s)$. Moreover, we take $\rho:=\frac{s_0}{s}=\frac{1}{9\kappa_{*}^4}$. Then, with probability at least $1-d^{-10}$, the online estimators $\widehat{\beta}_{i,t}$ for $i\in\{0,1\}$ and all $t\in[T]$ satisfy
    \begin{align*}
        \|\widehat{\beta}_{i,t} - \beta_i\|^2\lesssim \frac{\sigma_i^2s_0}{\phi_{\min}^2(s)}\frac{\log d}{t^{1-\gamma}} \quad {\rm and}\quad 
        \|\widehat{\beta}_{i,t} - \beta_i\|_{\ell 1}^2\lesssim \frac{\sigma_i^2s_0^2}{\phi_{\min}^2(s)}\frac{\log d}{t^{1-\gamma}}.
    \end{align*}
\end{Theorem}

We establish an upper bound on regret under the margin condition, a widely adopted assumption in the contextual bandit literature (\cite{bastani2020online, chen2021statistical}). It was initially introduced by \cite{tsybakov2004optimal} for investigating the optimal rate in classification. The margin condition controls the probability of encountering difficult requests near the decision boundary defined by $X^{\top}(\beta_1-\beta_0)=0$. By regulating these challenging instances, it reduces the likelihood of making incorrect decisions due to estimation errors in $\widehat{\beta}_{1,t}$ and $\widehat{\beta}_{0,t}$. Note that this condition is not required for online estimation.    

\begin{Assumption}[Margin condition]
    \label{assump:margincondition}
    There exist constants $C_0>0$ and $0\leq \nu\leq 1$ such that $\PP\left(|\inp{\beta_1-\beta_0}{X}|\leq L\right)\allowbreak\leq C_0L^{\nu}$ for all $L\in \RR^{+}$.
\end{Assumption}

\begin{Theorem}
    \label{thm:regret-nodc}
    Under the conditions in Theorem \ref{thm:estimation-nodc} and Assumption \ref{assump:margincondition}, the regret of Algorithm~\ref{alg:onlineHT} is upper bounded by
    \begin{align*}
        R_{T}\lesssim R_{\max}T^{1-\gamma} + \phi_{\min}(s)^{-(1+\nu)}s_0^{(1+\nu)/2}T^{(\gamma-1)(1+\nu)/2+1}\log^2d,
    \end{align*}
    where $R_{\max}:=\sup_{X\in\mathcal{X},a\in \{0,1\}}|\inp{X}{\beta_i}|$. By choosing $\gamma = \frac{1+\nu}{3+\nu} + \frac{2}{3+\nu}\log\big(\frac{R_{\max}}{(s_0^{(1+\nu)/2}\phi_{\min}^{-(1+\nu)}(s) \log^2 d}\big)/\log T$, Algorithm \ref{alg:onlineHT} achieves the best regret upper bound
    \begin{equation*}
     R_{T}\lesssim \big(R_{\max}s_0/\phi_{\min}^2(s)\big)^{(1+\nu)/(3+\nu)}T^{2/(3+\nu)}\log^{4/(3+\nu)}d  .
    \end{equation*}
\end{Theorem}

Theorem \ref{thm:regret-nodc} indicates that the regret of Algorithm \ref{alg:onlineHT} arises from two distinct sources: (i) the regret due to exploration, which is of order $\widetilde{O}(T^{1-\gamma})$ and is unaffected by the margin condition, and (ii) the regret stemming from incorrect decisions caused by estimation errors, which is of order $\widetilde{O}(T^{(\gamma-1)(1+\nu)/2+1})$. When $\nu$ is close to 0, implying a weaker margin condition, the second source of regret tends to be larger as more samples are close to the decision boundary, resulting in undistinguishable actions.  Algorithm \ref{alg:onlineHT} achieves a regret upper bound of $\widetilde{O}(T^{2/(3+\nu)})$ provided that the exploration probability is appropriately scheduled. In particular, when $\nu=1$, the best regret upper bound is $\widetilde{O}(\sqrt{s_0 T})$, which is comparable to the regret lower bound established in \cite{ren2024dynamic}; when $\nu=0$ corresponding to the scenario without a margin condition, the best regret upper bound is $\widetilde{O}(T^{2/3})$, which is also well known in the literature on high-dimensional bandits and $\varepsilon$-greedy algorithm (\cite{hao2020high}).  

Furthermore, Theorems \ref{thm:estimation-nodc} and \ref{thm:regret-nodc} indicate that selecting the parameter $\gamma$ involves a trade-off between the estimation error rate and the regret upper bound. When $\gamma = 0$, Algorithm \ref{alg:onlineHT} produces estimators that are classically $\sqrt{T}$-consistent, i.e., $\|\widehat{\beta}_{i,T} - \beta_i\|^2=\widetilde{O}(T^{-1})$. However, this setting maintains a constant probability of exploration, resulting in a trivial $O(T)$ regret even when $\nu=1$. When $\gamma$ is chosen as recommended in Theorem \ref{thm:regret-nodc}, 
Algorithm \ref{alg:onlineHT} attains a regret of $\widetilde{O}(\sqrt{T})$. However, the estimation error bound becomes $\|\widehat{\beta}_{i,T} - \beta_i\|^2=\widetilde{O}(T^{-1/2})$.  

\begin{Remark} Compared with existing studies, our results improve upon previous work in several key aspects: (i) While most prior analyses employing $\varepsilon$-greedy policies focus exclusively on low-dimensional bandits (e.g., \cite{chen2021statistical}), we extend these results to the high-dimensional setting, obtaining the same regret upper bound; (ii) Existing results for high-dimensional bandits, such as those by \cite{hao2020high, ma2024high,ren2024dynamic}, typically require more restrictive covariate or signal conditions to achieve a regret bound of $\widetilde{O}(\sqrt{T})$. In contrast, we demonstrate that under an appropriate margin condition, we can achieve the same regret bound without these stronger assumptions. Furthermore, we relax the margin parameter restriction to $\nu<1$ accommodating scenarios that lack a clear separation between optimal and suboptimal actions, as commonly encountered in practice.
\end{Remark}

%In Section \ref{sec:para-dc}, we demonstrate that, under an additinonal covariate diversity condition,  an exploration-free algorithm, i.e., when $\varepsilon_t\equiv 0$, achieves an optimal regret of $\widetilde{O}(s_0 \log T)$ while yields classically $\sqrt{T}$-consistent estimators.

\section{Inference by Debiasing and Tradeoff with Regret}\label{sec:para-nodc}
The estimators $\widehat{\beta}_{0,T}$ and $\widehat{\beta}_{1,T}$ produced by the hard-thresholding algorithm are usually biased and cannot be immediately used for statistical inference. The bias arises from two aspects: the implicit regularization imposed by hard-thresholding that is also common in sparse learning and the correlation inherent in the adaptively collected data $\{(X_t,a_t,y_t)\}_{t=1}^T$. Our proposed inference framework starts with a debiasing procedure to $\widehat{\beta}_{0, T}$ and $\widehat{\beta}_{1,T}$. 

Debiasing methods for (\emph{offline}) LASSO were initially introduced by \cite{javanmard2014confidence, f29b4853-f902-378f-99ba-4ec108f6737e, zhang2014confidence}. The key idea, imagining $\{(X_t, a_t, y_t): a_t=1, t\in[T]\}$ are independent for narrative simplicity, is to find a de-correlation matrix and construct the estimator   
\begin{align}\label{eq:offline-debias}
    \widehat{\beta}_{i}^{\uoff}:=\widehat{\beta}_{i,T} + \frac{1}{\big|\{t\in [T]:a_t=i\}\big|}\sum_{t=1}^{T}\indicator(a_t=i)M_{T}^{(i)}X_t(y_t-X_t^{\top}\widehat{\beta}_{i,T}),\quad \forall i=0,1, 
\end{align}
 where $M_{T}^{(i)}\in \RR^{d\times d}$ is an approximate inverse of the sample covariance matrix $\big|\big\{t\in [T]:a_t=i\big\}\big|^{-1}\sum_{t=1}^{T} \indicator(a_t=i)X_tX_t^{\top}$. The rationale behind this estimator is that the difference $\widehat{\beta}_{i}^{\uoff}-\beta_i$ can be decomposed as
\begin{align}
    \widehat{\beta}_{i}^{\uoff}-\beta_i&=\frac{1}{\big|{t\in [T]:a_t=i}\big|}\sum_{t=1}^{T} \indicator(a_t=i)M_{T}^{(i)}X_t\xi_t \label{eq:offline-diff}\\ 
    &\quad +  \frac{1}{\big|{t\in [T]:a_t=i}\big|}\sum_{t=1}^{T} \indicator(a_t=i)\left(I_d-M_{T}^{(i)}X_tX_t^{\top}\right)(\widehat{\beta}_{i,T} - \beta_i),\notag
\end{align}
where the first term on the RHS of (\ref{eq:offline-diff}) is asymptotically normal conditioned on $X_1,\cdots, X_T$, under the assumption that the noise $\xi_t$'s are i.i.d. By carefully constructing the de-correlation matrix $M_T^{(i)}$, the second term on the RHS can be asymptotically dominated by the first term. 

However, the debiasing procedure outlined in equation \eqref{eq:offline-debias} is not appropriate for adaptively collected data, where the chosen action $a_t$ heavily depends on all historical data $\{(X_\tau, a_{\tau}, y_{\tau})\}_{\tau=1}^{t-1}$. This interdependence introduces significant challenges in characterizing the distribution of the sum $\sum_{t=1}^{T} \mathbb{I}(a_t=i) M_{T}^{(i)} X_t \xi_t$. Furthermore, constructing an approximate inverse of the sample covariance matrix $\widehat{\Lambda}_{i,T}:=\big|\big\{t\in [T]:a_t=i\big\}\big|^{-1}\sum_{t=1}^{T} \indicator(a_t=i)X_tX_t^{\top}$ becomes problematic due to this interdependence and may even be infeasible because of the potential non-invertibility of $\mathbb{E}[\mathbb{I}(a_t=i) X_t X_t^{\top}]$.

We propose an online debiased estimator based on the inverse probability weight (IPW), defined as, 
\begin{align}\label{eq:debias-ipw}
    \widehat{\beta}_{i}^{\uipw}:=\widehat{\beta}_{i,T} + \frac{1}{T}\sum_{t=1}^{T}\frac{\indicator(a_t=i)}{i\pi_t+(1-i)(1-\pi_t)}M_TX_t(y_t-X_t^{\top}\widehat{\beta}_{i,T}),\quad \forall i=0,1. 
\end{align}
The de-correlation matrix $M_T$, detailed later, is constructed at time $T$ and is identical for both arms, which serves as an estimate of $\Sigma^{-1}$. The rationale for our estimator is as follows: the difference  $\widehat{\beta}_{i}^{\uipw}-\beta_i$ can be decomposed as
\begin{align*}
    \widehat{\beta}_{i}^{\uipw} - \beta_i&= \underbrace{\frac{1}{T}\sum_{t=1}^{T} \frac{\indicator(a_t=i)}{i\pi_t+(1-i)(1-\pi_t)}M_{T}X_t\xi_t}_{=:V_{Ti}} \\ &\quad +  \underbrace{\frac{1}{T}\sum_{t=1}^{T} \left(I_d-\frac{\indicator(a_t=i)}{i\pi_t+(1-i)(1-\pi_t)}M_{T}X_tX_t^{\top}\right)(\widehat{\beta}_{i,T} - \beta_i)}_{=:B_{Ti}},\quad \forall i=0,1.
\end{align*}
The term $V_{Ti}$ characterizes the variance of $\widehat{\beta}_{i}^{\uipw} - \beta_i$. It can be further decomposed by expressing $M_T = \Sigma^{-1} + (M_T - \Sigma^{-1})$, where the first term constitutes the primary component forming a zero-mean martingale and the second term can be negligible if $M_T$ is sufficiently close to $\Sigma^{-1}$. On the other hand, the term $B_{Ti}$ represents a vanishing bias term stochastically dominated by the variance of $V_{Ti}$. 

The incorporation of IPW substantially complicates the variance analysis compared to the offline LASSO. Consider $V_{T1}$, corresponding to arm-1, as an illustrative example. Define the arm-optimal sets by $\mathcal{X}_1:=\{X: \inp{\beta_1-\beta_0}{X}\geq 0\}$ and $\mathcal{X}_0:=\{X:\inp{\beta_0-\beta_1}{X}>0\}$. If a sample $X_t\in \mathcal{X}_1$, it is utilized in constructing $\widehat{\beta}_{1}^{\uipw}$ with a propensity score $\pi_t$ approaching one as the estimators $\widehat{\beta}_{1,t}$ and $\widehat{\beta}_{0,t}$ become more accurate when $t\rightarrow \infty$. Conversely, if $X_t\in \mathcal{X}_0$, the propensity score approaches zero as $t$ tends to infinity. Due to IPW, the effective sample size is predominantly determined by samples from $\mathcal{X}_1$, whereas samples from $\mathcal{X}_0$ contribute to variance inflation.  Specifically, if $\varepsilon_t\asymp t^{-\gamma}$, the variance contributed by $X_t\in \mathcal{X}_1$ is $O(T^{-2})$, while the variance contributed by $X_t\in \mathcal{X}_0$ is $O(t^{\gamma}/T^2)$. Define the arm-optimality-specific population matrices as  
\begin{align}\label{eq:arm-cov}
\Lambda_1:=\EE[XX^{\top}\indicator(\inp{\beta_1-\beta_0}{X}\geq 0)]\quad  {\rm and}\quad \Lambda_0:=\EE[XX^{\top}\indicator(\inp{\beta_1-\beta_0}{X}< 0)].
\end{align}
The asymptotic variance of $V_{T1}$ contributed by samples in $\mathcal{X}_1$ is approximately $\sigma_1^2\Sigma^{-1}\Lambda_1 \Sigma^{-1}/T$, while the asymptotic variance contributed by samples in $\mathcal{X}_0$ is approximately $\sigma_0^2\Sigma^{-1}\Lambda_0 \Sigma^{-1}/T^{1-\gamma}$, with the latter being the dominant term.  

It remains to bound the bias term $B_{Ti}$. Observe that as long as the de-correlation matrix $M_T$ can approximate the inverse of $\widehat{\Sigma}_{i,T}=\frac{1}{T}\sum_{t=1}^{T} \frac{\indicator(a_t=i)}{i\pi_t+(1-i)(1-\pi_t)}X_tX_t^{\top}$, then both $I_d-M_T\widehat{\Sigma}_{i,T}$ and $\widehat{\beta}_{i,T}-\beta$ would be small in an appropriate sense, thereby controlling $B_{Ti}$. Due to the incorporation of IPW, both $\widehat{\Sigma}_{1,T}$ and $\widehat{\Sigma}_{0,T}$ converge to the population covariance $\Sigma$, facilitating the construction of the de-correlation matrix more conveniently than the sample covariance matrices $\widehat{\Lambda}_{1,T}$ and $\widehat{\Lambda}_{0,T}$ employed in offline debiased LASSO \eqref{eq:offline-debias}. Additionally, since the total sample covariance $\widehat{\Sigma}_{T}=\frac{1}{T}\sum_{t=1}^{T} X_tX_t^{\top}$ also converges to $\Sigma$, we can simply exploit $\widehat{\Sigma}_{T}$ to construct $M_T$, making $M_T$ an approximate inverse of $\widehat{\Sigma}_{T}$ for both arms.

Let $m_{l,T}$ denote the solution to the following optimization problem for each $l\in[d]$:
\begin{align}\label{eq:optimization1}
    \min_{m\in \mathbb{R}^{d}} \frac{1}{2}m^{\top}\widehat{\Sigma}_{T}m - \inp{m}{e_l} + \mu_{T_1}\|m\| _{\ell_1},
\end{align}
where $e_l$ is the $l$-th canonical basis vector in $\RR^d$, and $\mu_{T_1}$ is a suitably chosen regularization parameter. We then construct $M_T=(m_{1,T},\cdots,m_{d,T})^{\top}$. 
%By KKT condition, we can derive
%\begin{align}\label{eq:kkt1}
%    \|\widehat{\Sigma}_{T}m_{l,T} - e_l\|_{\max}\leq \mu_{T_1} .
%\end{align}
Lemma \ref{lemma:feasible1} in Appendix \ref{sec: teclemma} demonstrates that this optimization problem is feasible with high probability if  $\mu_{T_1}=C_{\mu_1}\big(\log(d)/T\big)^{1/2}$ for a sufficiently large constant $C_{\mu_1}>0$. 

We remark that our debiasing procedure does not impose an additional storage burden compared to the bandit Algorithm~\ref{alg:onlineHT}. Specifically, it suffices to retain the summations $\sum_{t=1}^T \frac{\indicator(a_t=i)}{i\pi_t+(1-i)(1-\pi_t)}X_ty_t$ for $i=0,1$ as well as the sample covariance matrix $\widehat{\Sigma}_{T}=\frac{1}{T}\sum_{t=1}^{T} X_tX_t^{\top}$, which requires $O(d^2)$ space complexity. These summations can be efficiently updated in an online fashion. 

We denote $\Omega:=\Sigma^{-1}$ as the precision matrix and define $s_{\Omega}:=\max_{l\in [d]} \left|j\in [d]: \Omega_{l,j}\neq 0\right|$ as its row sparsity. Recall that the parameter $\gamma \in [0, 1)$ governs the rate at which the exploration probabilities decay. We can now state the central limit theorem result for IPW debiased estimators.   
\begin{Theorem}
    \label{thm:maininference-nodc}
    Suppose Assumptions \ref{assump:basic}-\ref{assump:margincondition} and conditions in Theorem~\ref{thm:estimation-nodc} hold, with $\nu>0$ in Assumption \ref{assump:margincondition}. The sparsity parameters $s_0$ and $s_{\Omega}$ satisfy $s_0\vee s_{\Omega}=o\big((T^{1-\gamma}\wedge \sqrt{T})/\log d\big)$ and $s_{\Omega}s_0=o(T^{\nu(1-\gamma)/2}/\log d)$. Let $M_{T}$ be constructed according to (\ref{eq:optimization1}) with $\mu_{T_1}:=C_{\mu_1}\big(\log (d)/t\big)^{1/2}$ for a large numerical constant $C_{\mu_1}>0$. Then, for any $l\in [d]$ and $i=0,1$, we have
    \begin{align*}
        \frac{\widehat{\beta}_{i(l)}^{\uipw} -\beta_{i(l)}}{\sqrt{\sigma_i^2S_{i(l)}^2/T^{1-\gamma}}}\stackrel{{\rm d.}}{\longrightarrow} N(0,1),\quad \textrm{ as  } d, T\to\infty,
    \end{align*}
    where $S_{i(l)}^2:=\frac{2}{c_{\varepsilon}(1+\gamma)}e_l^{\top}\Omega\Lambda_{1-i}\Omega e_l + T^{-\gamma}e_l^{\top}\Omega\Lambda_{i}\Omega e_l$.
\end{Theorem}

Theorems \ref{thm:regret-nodc} and \ref{thm:maininference-nodc} illustrate the trade-off between regret performance and inference efficiency in Algorithm~\ref{alg:onlineHT}. Specifically, when $\gamma$ is chosen appropriately to balance the two sources of regret thereby achieving the best regret upper bound $\widetilde{O}(T^{1-\gamma})$, the variance of statistical inference is $O(T^{-(1-\gamma)})$. This indicates that optimizing regret leads to a decrease in inference efficiency. This finding is consistent with the minimax lower bound established in \cite{simchi2023multi}, which asserts that the product of the estimation error and the square root of the regret remains of constant order \emph{in the worst case}.  Consequently, if the sole objective is statistical inference, maintaining a constant exploration probability $\varepsilon_t\equiv c_\varepsilon\in(0,1)$, i.e., $\gamma=0$,  yields the most efficient estimator with a variance of $O(T^{-1})$, corresponding to a $\sqrt{T}$-consistent estimator. However, this approach results in a trivial $O(T)$ regret.  Alternatively, choosing $\gamma=(1+\nu)/(3+\nu)$ achieves the best regret upper bound of $\widetilde{O}(T^{2/(3+\nu)})$, while the variance for parameter inference is $O(T^{-2/(3+\nu)})$.

The asymptotic normality established in Theorem \ref{thm:maininference-nodc} relies on the unknown parameters $\sigma_0$, $\sigma_1$, $S_1$, and $S_0$. To construct confidence intervals and perform hypothesis testing, it is essential to estimate these quantities accurately. We propose the following estimators for each $i \in \{0,1\}$: 
\begin{align*}
    &\widehat{\sigma}_i^2: = \frac{1}{T}\sum_{t=1}^{T}\frac{\indicator(a_t=i)}{i\pi_t+(1-i)(1-\pi_t)} (y_t-X_t^{\top}\widehat{\beta}_{i,t-1})^2;  \\
    &\widehat{S}_{i(l)}^2:= \frac{2}{c_{\varepsilon}(1+\gamma)}m_{l,T}^{\top}\widehat{\Lambda}_{1-i,T}m_{l,T} + T^{-\gamma}m_{l,T}^{\top}\widehat{\Lambda}_{i,T}m_{l,T},
\end{align*}
where $c_{\varepsilon}>0$ is the constant in Theorem~\ref{thm:estimation-nodc} defining exploration proabilities, and  $\widehat{\Lambda}_{i,T}:=T^{-1}\sum_{t=1}^{T}\indicator(a_t=i) X_tX_t^{\top}$ for $i\in \{0,1\}$.

\begin{Theorem}
    \label{thm:stuinference-nodc}
    Suppose the conditions in Theorem \ref{thm:maininference-nodc} hold. For any $l\in [d]$, we have
    \begin{align*}
        \frac{\widehat{\beta}_{i(l)}^{\uipw} -\beta_{i(l)}}{\sqrt{\widehat{\sigma}_i^2\widehat{S}_{i(l)}^2/T^{1-\gamma}}} \stackrel{{\rm d.}}{\longrightarrow} N(0,1)\quad \textrm{ as } d, T\to\infty .
    \end{align*}
\end{Theorem}

%Based on Theorem \ref{thm:stuinference-nodc}, an asymptotically valid confidence interval at coverage level of $100(1-\theta)\%$ for any $\theta\in (0,1)$ can be constructed as
%\begin{align*}
    %\widehat{\text{CI}}_{l,T}^{(i)}:=\left(\widehat{\beta}_{i(l)}^{\uipw} - z_{\theta/2}\frac{\widehat{\sigma}_i\widehat{S}_{i(l)}}{\sqrt{T^{1-\gamma}}},\quad \widehat{\beta}_{i(l)}^{\uipw} + z_{\theta/2}\frac{\widehat{\sigma}_i\widehat{S}_{i(l)}}{\sqrt{T^{1-\gamma}}}\right),\quad \forall l\in[d], 
%\end{align*}
%where $z_{\theta}:=\Phi^{-1}(1-\theta)$ is the upper $\theta$ quantile of the standard normal. Theorem~\ref{thm:stuinference-nodc} guarantees that $\lim_{T\rightarrow \infty} \PP(\beta_{i,l}\in \widehat{\text{CI}}_{l,T}^{(i)})=1-\theta$. 

The difference between the two estimators, $\widehat{\beta}_1^{\uipw}-\widehat{\beta}_0^{\uipw}$, also exhibits asymptotic normality in a manner similar to the individual estimators. We demonstrate that the asymptotic covariance between  $\widehat{\beta}_1^{\uipw}$ and $\widehat{\beta}_0^{\uipw}$ is zero. Consequently, the variance of $\widehat{\beta}_1^{\uipw}-\widehat{\beta}_0^{\uipw}$ can be estimated as the direct sum of the variance estimates of $\widehat{\beta}_1^{\uipw}$ and $\widehat{\beta}_0^{\uipw}$. The following corollary immediately enables testing the hypothesis (\ref{eq:test-H3}) for comparing different arm parameters. 

\begin{Corollary}
    \label{cor:difference-nodc}
    Suppose the conditions in Theorem \ref{thm:maininference-nodc} hold. For any $l\in [d]$, we have
    \begin{align*}
        \frac{(\widehat{\beta}_{1(l)}^{\uipw}-\widehat{\beta}_{0(l)}^{\uipw}) - (\beta_{1(l)}-\beta_{0(l)})}{\sqrt{(\widehat{\sigma}_1^2\widehat{S}_{1(l)}^2 + \widehat{\sigma}_0^2\widehat{S}_{0(l)}^2)/T^{1-\gamma}}} \stackrel{{\rm d.}}{\longrightarrow} N(0,1),\quad \textrm{ as } d, T\to\infty. 
    \end{align*}
\end{Corollary}

\section{Simultaneous Optimal Inference and Regret Performance}\label{sec:para-dc}
Section~\ref{sec:para-nodc} demonstrates that a moderately vanishing sequence of exploration probabilities, $\varepsilon_t$, is essential for achieving non-trivial regret. However, this requirement causes the inverse propensities, $\pi_t^{-1}$ or $(1-\pi_t)^{-1}$, to diverge to infinity. While such divergent inverse propensities facilitate bias correction, they concurrently result in a significant inflation of the variance of the resulting debiased estimator. This inflated variance restricts the IPW debiased estimator (\ref{eq:debias-ipw}) from attaining optimal statistical efficiency, leading to excessively wide confidence intervals and diminishing the sensitivity of parameter evaluations.

In this section, we demonstrate that, under an additional covariate diversity condition, an exploration-free algorithm achieved by setting $\varepsilon_t\equiv 0$ in Algorithm~\ref{alg:onlineHT}, can simultaneously achieve the optimal regret bound of $O(\log T)$ and optimal inference efficiency. The covariate diversity condition ensures that the covariate $X_t$ is diversely distributed across all directions, thereby ensuring the exploration-free algorithm to inherently explore each arm adequately. Consequently, an alternative debiasing method based on average weighting (AW) can effectively correct the bias. Most importantly, AW avoids variance inflation in the absence of IPW,  resulting in an asymptotically unbiased estimator with a variance of $O(T^{-1})$. 
 
\subsection{Exploration-Free Algorithm and $O(\log T)$ Regret}
Intuitively, the IPW terms $\mathbb{I}(a_t=1)/\pi_t$ and $\mathbb{I}(a_t=0)/(1-\pi_t)$ in equation (\ref{eq:debias-ipw}) become unnecessary when the algorithm operates in an exploration-free manner. The following \emph{covariate diversity} assumption is essential and is commonly imposed in high-dimensional bandit literature, as highlighted by \cite{bastani2021mostly} and \cite{ren2024dynamic}. Recall the arm-specific population matrices $\Lambda_0$ and $\Lambda_1$ defined in equation (\ref{eq:arm-cov}).

\begin{Assumption}[Covariate Diversity]
    \label{covariate diversity}
    There exists positive constant $\lambda_0$ such that 
    $$
    \min\{\lambda_{\min}(\Lambda_1), \lambda_{\min}(\Lambda_0)\}\geq \lambda_0.
    $$ 
    % and, for any unit vector $v\in \mathbb{R}^d$ and $i=0,1$, conditional on the history $\mathcal{F}_{t-1}$, the probability
    % \begin{align*}
    %     \PP(v^{\top}X_tX_t^{\top}v\indicator(a_t=i)\geq h|\mathcal{F}_{t-1})\geq \zeta.
    % \end{align*}
\end{Assumption}

%The exploration-free algorithm was initially proposed and investigated by \cite{ma2024high}, demonstrating that the algorithm can deliver $\sqrt{t}$-consistent estimators of arm parameters for all $t\in[T]$. Their analysis relies on a stronger covariate diversity assumption, which requires that for any unit vector $v \in \mathbb{R}^d$ and $i \in {0, 1}$, conditional on the history $\mathcal{F}_{t-1}$, the probability satisfies

This assumption is satisfied by various covariate distributions, such as the uniform distribution and the truncated multivariate Gaussian distribution. \cite{bastani2021mostly} also provides sufficient, easier-to-verify conditions that ensure Assumption \ref{covariate diversity} holds.

\begin{Theorem}
    \label{thm:estimation-dc}
    If we take $\rho:=\frac{s_0}{s}=\frac{\lambda_0^4}{150\phi_{\max}^4(s)}$ and $\eta=\frac{\lambda_0}{8\phi_{\max}^2(s)}$. Suppose Assumptions \ref{assump:basic} - \ref{covariate diversity} hold. When setting $\varepsilon_t\equiv 0$ in Algorithm \ref{alg:onlineHT}, then after a burn-in period with length $T_0=C_1s^{(2+\nu)/\nu}\log^{(2+\nu)/\nu} d$ for some constant $C_1>0$, the following bound holds with probability at least $1-d^{-10}$ for all $t\geq T_0$,
    \begin{align*}
        \|\widehat{\beta}_{i,t} - \beta_i\|^2\lesssim \frac{\sigma_i^2s_0}{\lambda_0^2}\frac{\log d}{t},\quad \forall i=0,1.
    \end{align*}
\end{Theorem}

The $O(t^{-1/2})$ convergence rates of $\widehat{\beta}_{i,t}$, together with the margin condition stated in Assumption~\ref{assump:margincondition}, yield a regret bound of $\widetilde{O}(\log T)$.

\begin{Theorem}
    \label{thm:regret-dc}
    Suppose Assumptions \ref{assump:basic} - \ref{covariate diversity} hold.  By setting $\varepsilon_t\equiv 0$, Algorithm \ref{alg:onlineHT} achieves a regret with an upper bound:
    \begin{align*}
        R_{T}\lesssim \lambda_0^{-(1+\nu)}s_0^{(1+\nu)/2}T^{(1-\nu)/2}\log^2 d \quad \text{if }\ 0\leq \nu<1;\quad\quad  R_{T}\lesssim \lambda_0^{-2}s_0\log T\log^2 d \quad\text{if }\  \nu=1.
    \end{align*}
\end{Theorem}

Theorem \ref{thm:regret-dc} improves upon the regret performance established in \cite{bastani2021mostly} for low-dimensional LCB, which relies on both the margin and covariate diversity conditions. Here, we demonstrate that the same optimal $O(\log T)$ regret bound can be achieved even in high-dimensional LCB, albeit with an $s_0^{(1+\nu)/2}$ factor  that depends on the sparsity level rather than a $\operatorname{poly}(d)$ factor that depends on the ambient dimension \citep{bastani2021mostly}. 

\begin{comment}
    \begin{Remark}
\red{(delete?)} The margin condition and the covariate diversity condition contribute to regret reduction from distinct perspectives. The margin condition mitigates regret by ensuring that only a small proportion of samples lie near the decision boundary. In contrast, the covariate diversity condition both reduces estimation error, thereby decreasing the regret associated with parameter estimation, and enables an exploration-free algorithm, eliminating regret arising from exploration. Consequently, this allows us to optimize regret from $\widetilde{O}(\sqrt{T})$ to $\widetilde{O}(\log T)$. 
\end{Remark}
\end{comment}

\subsection{$\sqrt{T}$-Consistent Inference}
While an $O(\log T)$ regret bound has been established in \citep{bastani2020online} using their novel forced-sampling algorithm design, it becomes extremely difficult to characterize the uncertainty of their estimators due to the complex and uninterpretable dependencies induced by the forced sampling scheme. In contrast, we show that our exploration-free algorithm not only maintains the $O(\log T)$ regret bound but also enables accurate statistical inference.

Under Assumption~\ref{covariate diversity}, the arm-specific population matrices  $\Lambda_1$ and $\Lambda_0$ are invertible, indicating the possibility of finding an approximate inverse of the arm-specific sample covariances $\widehat{\Lambda}_{1,T}:=T^{-1}\sum_{t=1}^{T}\indicator(a_t=1) X_tX_t^{\top}$ and $\widehat{\Lambda}_{0,T}:=T^{-1}\sum_{t=1}^{T}\indicator(a_t=0) X_tX_t^{\top}$, respectively. Consequently, IPW becomes unnecessary and  we introduce a debiased estimator based on average weighting (AW) defined as 
\begin{align*}
    \widehat{\beta}_{i}^{\uaw}:&=\widehat{\beta}_{i,T} + \frac{1}{T}\sum_{t=1}^{T}\indicator(a_t=i) M_T^{(i)}X_t(y_t-X_t^{\top}\widehat{\beta}_{i,T}) \\
    &= \beta_i + \underbrace{\frac{1}{T}\sum_{t=1}^{T} \indicator(a_t=i)M_T^{(i)}X_t\xi_t}_{V_{Ti}} +  \underbrace{\frac{1}{T}\sum_{t=1}^{T} \left(I_d-\indicator(a_t=i)M_T^{(i)}X_tX_t^{\top}\right)(\widehat{\beta}_{i,T} - \beta)}_{B_{Ti}}.
\end{align*}
The bias term $B_T$ is controlled provided that the de-correlation matrices $M_T^{(1)}$ and $M_T^{(0)}$ can accurately approximate the inverses of $\widehat{\Lambda}_{1,T}$ and $\widehat{\Lambda}_{0,T}$, respectively. These matrices are constructed similarly to those described in Section~\ref{sec:para-nodc}. 
Specifically, the $l$-th row of $M_T^{(i)}$, denoted by $m_{l,T}^{(i)}$, is the solution to
\begin{equation}
    \min_{m\in \mathbb{R}^{d}}\ \frac{1}{2}m^{\top}\widehat{\Lambda}_{i,T}m - \inp{m}{e_l} + \mu_{T_2}\|m\| _{\ell_1} \label{eq:optimization2}.
\end{equation}
%By KKT condition, we can derive
%\begin{equation}
%    \|\widehat{\Lambda}_{i,T}m_{l}^{(i)} - e_l\|_{\max}\leq \mu_{T_2} \label{kkt2},
%\end{equation}
Lemma \ref{lemma:feasible2} in Appendix \ref{sec: teclemma} shows that the optimization problem is feasible with high probability if  $\mu_{T_2}:=C_{\mu_2} \big(s_0\log(d)/T\big)^{\nu/2}$ with a large numerical constant $C_{\mu_2}>0$.

We briefly explain why the term $V_{Ti}$ exhibits a variance of $O(T^{-1})$ by considering arm 1 as an example. The de-correlation matrix $M_T^{(1)}$ is designed to approximate $\Lambda_1^{-1}$. The term $V_{T1}$ can be further decomposed by expressing $M_T^{(1)}=\Lambda_1^{-1}+\big(M_T^{(1)}-\Lambda_1^{-1}\big)$, where the first term, defining a zero-mean martingale, is the dominating component. The variance contribution of each observation to this martingale is  $O(T^{-2})$. Since the margin condition ensures that the proportion of samples near the decision boundary remains relatively small, nearly all the variance is contributed by observations from the arm-1 optimality set $\mathcal{X}_1:=\big\{X: \langle \beta_1-\beta_0, X\rangle \geq 0\big\}$, as $T$ approaches infinity. Consequently, the overall asymptotic covariance of $V_{T1}$ converges to $\Lambda_1^{-1}/T$. This result is comparable to findings in classical offline high-dimensional inference applied to i.i.d. data (\cite{10.1214/17-AOS1630}).

Define the row sparsity of the inverses of arm-specific covariance matrices as 
\begin{align*}
         s_{\Lambda^{-1}}:=\max\left\{\max_{l\in [d]} \left|j\in [d]: \Lambda_{1(lj)}^{-1}\neq 0\right|, \max_{l\in [d]} \left|j\in [d]: \Lambda_{0(lj)}^{-1}\neq 0\right|\right\}.
\end{align*}
We establish the asymptotic normality of the AW estimators as follows.
\begin{Theorem}
    \label{thm:maininference-dc}
Suppose Assumptions \ref{assump:basic}-\ref{covariate diversity} hold and assume that $s_{\Lambda^{-1}}=o(T/\log d)$, $s_0=o(T^{\nu/(2+\nu)}/\log d)$, and $s_{\Lambda^{-1}}^2s_0^{\nu}=o(T^{\nu}/\log^2 d)$. Let $M_{T}^{(1)}$ and $M_T^{(0)}$ be constructed according to (\ref{eq:optimization2}) with $\mu_{T_2}:=C_{\mu_2}\big(s_0\log(d)/T\big)^{1/2}$ for some constant $C_{\mu_2}$.  Then, for any $l\in [d]$ and $i=0,1$, the following holds:
    \begin{align*}
        \frac{\widehat{\beta}_{i(l)}^{\uaw} -\beta_{i(l)}}{\sigma_i\sqrt{\Lambda_{i(ll)}^{-1}/T}}\stackrel{{\rm d.}}{\longrightarrow} N(0,1)\quad \textrm{ as  } d, T\to\infty.
    \end{align*}
\end{Theorem}

Theorems~\ref{thm:regret-dc} and~\ref{thm:maininference-dc} demonstrate that the covariate diversity condition enables the simultaneous achievements of optimal inference efficiency, characterized by a variance of $O(T^{-1})$, and optimal regret, quantified as $\widetilde{O}(\log T)$.

We propose data-driven estimates of the variances. For $l\in [d]$ and $i=0,1$, define
\begin{align*}
    \widehat{\sigma}_i^2:= \frac{1}{|\{t\in [T]: a_t=i\}|}\sum_{t=1}^{T}\indicator(a_t=i) (y_t-X_t^{\top}\widehat{\beta}_{i,t-1})^2 \quad {\rm and}\quad \widehat{\Lambda}_{i(ll)}^{-1}:= m_{l,T}^{(i)\top}\widehat{\Lambda}_{i,T}m_{l,T}^{(i)}.
\end{align*}

\begin{Theorem}
    \label{thm:stuinference-dc}
    Suppose the conditions in Theorem \ref{thm:maininference-dc} hold. Then, for any $l\in [d]$ and $i=0,1$, we have
    \begin{align*}
        \frac{\widehat{\beta}_{i(l)}^{\uaw} -\beta_{i(l)}}{\widehat{\sigma}_i\sqrt{\widehat{\Lambda}_{i(ll)}^{-1}/T}} \stackrel{{\rm d.}}{\longrightarrow} N(0,1),\quad \textrm{ as } d, T\to\infty. 
    \end{align*}
\end{Theorem}

Similarly, we can establish the asymptotic distribution for the difference $\widehat{\beta}_1^{\uaw}-\widehat{\beta}_0^{\uaw}$. 

\begin{Corollary}
    \label{cor:difference-dc}
    Suppose the conditions in Theorem \ref{thm:maininference-dc} hold. For any $l\in [d]$, we have
    \begin{align*}
        \frac{(\widehat{\beta}_{1(l)}^{\uaw}-\widehat{\beta}_{0(l)}^{\uaw}) - (\beta_{1(l)}-\beta_{0(l)})}{\sqrt{(\widehat{\sigma}_1^2\widehat{\Lambda}_{1(ll)}^{-1} + \widehat{\sigma}_0^2\widehat{\Lambda}_{0(ll)}^{-1})/T}} \stackrel{{\rm d.}}{\longrightarrow} N(0,1),\quad \textrm{ as } d,T\to\infty.
    \end{align*}
\end{Corollary}

%Theorem~\ref{thm:stuinference-dc} and Corollary~\ref{cor:difference-dc} immediately allow us to construct confidence intervals and test hypotheses (\ref{eq:test-H1}) - (\ref{eq:test-H3}).

\section{Inference on the Optimal Policy's Value}
We now examine the statistical inference for the policy's value under the optimal policy, which represents the maximum expected reward and is commonly referred to as the $Q$-function in existing literature (\cite{murphy2003optimal, sutton2018reinforcement}). Given the context information $X$, we define the optimal action as $a^{*}(X):=\indicator(\inp{\beta_1-\beta_0}{X}>0)=\indicator(X\in \calX_1)$. The value associated with the optimal policy is defined by $V^{*}:=\EE\left[\inp{\beta_{a^{*}(X)}}{X}\right]$.
%\begin{align*}
    %V^{*}:=\EE\left[\inp{\beta_{a^{*}(X)}}{X}\right].
%\end{align*}

Under the covariate diversity condition, we find that debiasing is unnecessary when inferring the optimal policy. This is because, in the absence of exploration, the policy selects the optimal action with probability approaching one over time. Consequently, this results in a remarkably concise form of the estimator of $V^{\ast}$.

Intuitively, when the context $X_t$ is situated near the decision boundary—such that the gap between the two arms $|\inp{\beta_1-\beta_0}{X_t}|$ is very small, it becomes challenging to identify the optimal action at time $t$. However, even if a suboptimal action is chosen in these instances, the rewards for both arms are nearly identical, resulting in only negligible bias in the estimation of the optimal value. Furthermore, the margin condition ensures that the proportion of such samples remains relatively small. Consequently, all available data can be directly utilized to estimate the optimal value. We propose the following estimator:
\begin{align*}
    \widehat{V}_{T}: = \frac{1}{T}\sum_{t=1}^{T} y_t.
\end{align*}

Theorem~\ref{thm:valueinf} establishes the asymptotical normality of $\widehat V_T$.  Notably, when the objective is solely to infer the optimal policy value, the sparsity conditions on $\Omega$ or $\Lambda_0, \Lambda_1$ discussed in the preceding sections become unnecessary.

\begin{Theorem}
    \label{thm:valueinf}
    Suppose Assumption \ref{assump:basic}-\ref{covariate diversity} hold, with $\nu>0$ in Assumption \ref{assump:margincondition}. Assume $s_0=o(T^{\nu/(1+\nu)}/\log d)$, and there exists $b_{\max}>0$ such that $\|\beta_0\|+\|\beta_1\|\leq b_{\max}$. Then, by setting $\varepsilon_t\equiv 0$ in Algorithm~\ref{alg:onlineHT}, we have 
    \begin{align*}
        \frac{\widehat{V}_T - V^{*}}{\sqrt{S_V^2/T}} \stackrel{{\rm d.}}{\longrightarrow} N(0,1)\quad \textrm{ as } d, T\to\infty,
    \end{align*}
    where $S_V^2:=\sigma_1^2\int \indicator(X\in\calX_1)d\mathcal{P}_{\mathcal{X}} + \sigma_0^2\int \indicator(X\in\calX_0)d\mathcal{P}_{\mathcal{X}} + \textrm{Var}[\inp{\beta_{a^{*}(X)}}{X}]$.
\end{Theorem}

Theorem \ref{thm:valueinf} demonstrates that the estimator $\widehat V_T$ is asymptotically normal and achieves $\sqrt{T}$-consistency. The asymptotic variance of the proposed estimator arises from two sources. The first two terms, involving $\sigma_1^2$ and $\sigma_0^2$, are associated with the noise in the observed rewards. These terms are proportional to the fraction of optimal samples in each arm, specifically the size of $\mathcal{X}_1$ and $\mathcal{X}_0$. Consequently, the arm with a higher proportion of optimal samples is selected more frequently, thereby contributing more significantly to the overall variance. On the other hand, $\text{Var}[\inp{\beta_{a^{*}(X)}}{X}]$ represents the variance inherent in the contextual information, even if the optimal policy is known.

We propose the following data-driven estimate of the variance: 
\begin{align*}
    \widehat{S}_V^2:&=\underbrace{\frac{1}{T}\sum_{t=1}^T \widehat{\sigma}_{1}^2\indicator(a_t=1) + \widehat{\sigma}_{0}^2\indicator(a_t=0)}_{G_T}  + \underbrace{\frac{1}{T} \sum_{t=1}^T \inp{\widehat{\beta}_{a_t,t-1}}{X_t}^2 - \left(\frac{1}{T} \sum_{t=1}^T \inp{\widehat{\beta}_{a_t,t-1}}{X_t}\right)^2}_{W_T},
\end{align*}
where $G_T$ and $W_T$ can consistently estimate $\sigma_1^2\int \indicator(X\in\calX_1)d\mathcal{P}_{\mathcal{X}} + \sigma_0^2\int \indicator(X\in\calX_0)d\mathcal{P}_{\mathcal{X}}$ and $\text{Var}[\inp{\beta_{a^{*}(X)}}{X}]$, respectively. The estimators $\widehat\sigma_1^2$ and $\widehat\sigma_0^2$ are defined as in Section~\ref{sec:para-dc}.

\begin{Theorem}
    \label{thm:stuvalueinf}
    Suppose the conditions in Theorem \ref{thm:valueinf} hold. We have
    \begin{align*}
        \frac{\widehat{V}_T - V^{*}}{\sqrt{\widehat{S}_V^2/T}} \stackrel{{\rm d.}}{\longrightarrow} N(0,1)\quad \textrm{ as } d, T\to\infty.
    \end{align*}
\end{Theorem}

\section{Numerical Results}

\subsection{Simulation Studies}\label{sec:simulation}
We demonstrate the empirical performance of our method for both parameter and optimal value inference. We consider two scenarios with different covariate dimensions, decision time horizons, and sparsity levels for the two-armed bandit: (1) $d=600$, $T=300$ and $s_0=3$; (2) $d=600$, $T=300$ and $s_0=5$. For each scenario, we generate the first $s_0$ entries of $\beta_1$ and $\beta_0$ from $U[0.5, 1]$, while keeping the other entries at zero. Additionally, we consider three different exploration parameter choices: $\varepsilon_t\asymp t^{-\gamma}$ for $\gamma=1/3, 1/2$, and the exploration free algorithm with $\varepsilon_t=0$. We conduct 1,000 independent trials for all scenarios.

\begin{figure} [t!]
	\centering
		\includegraphics[scale=0.18]{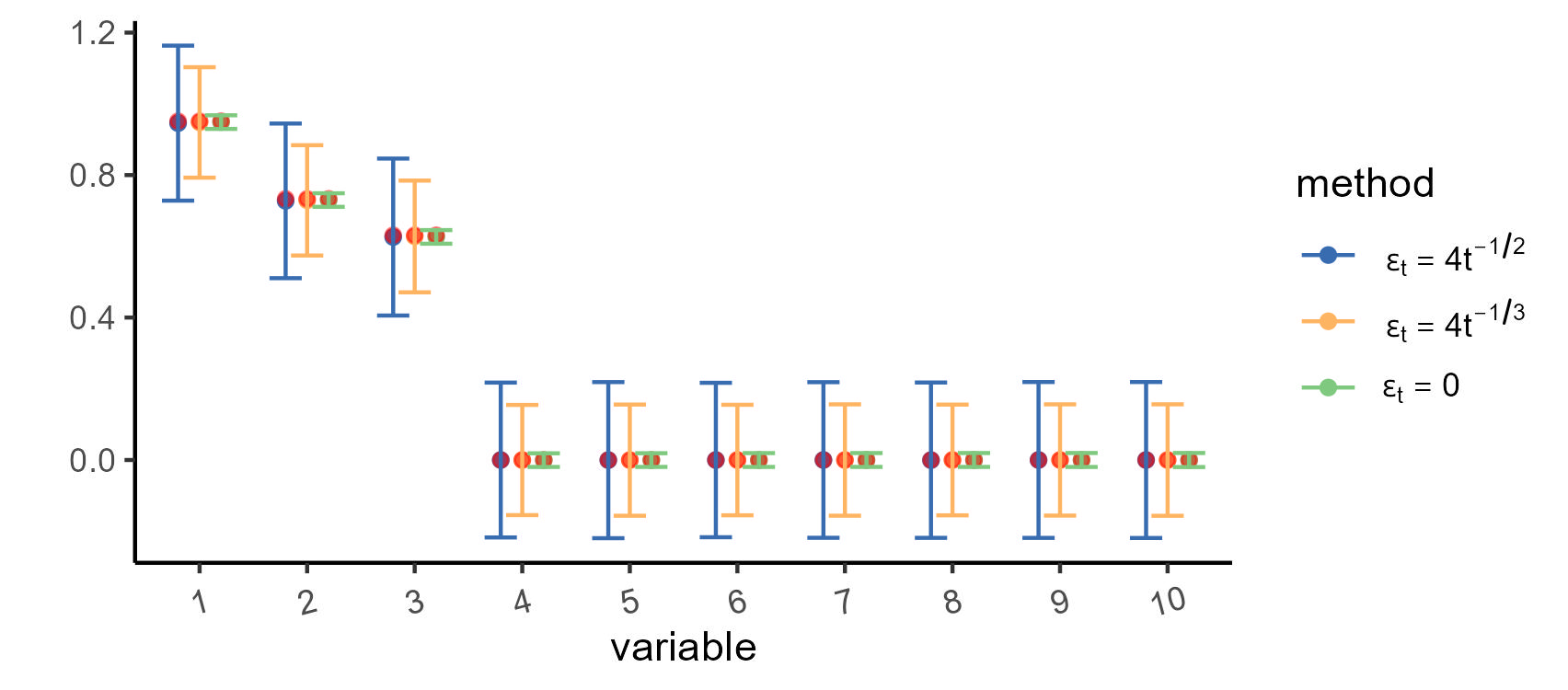}
        %}
	
    %\setcounter {subfigure} {0} {
		\includegraphics[scale=0.18]{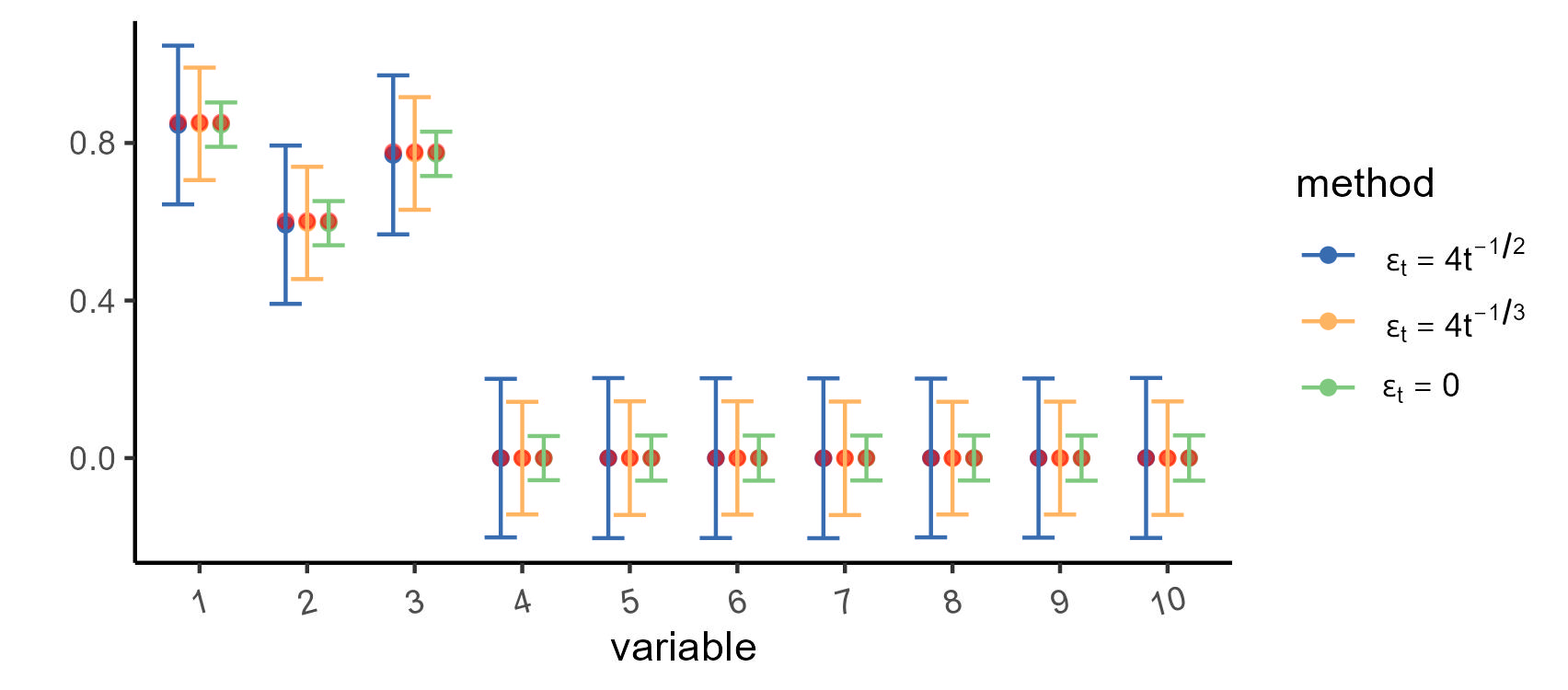} 
    %}
    {\small\linespread{0.9}\selectfont
	\caption{Point and interval estimators of the first ten entries in $\beta_1$(top) and $\beta_0$(bottom) under scenario (1). The red points indicate the true value. }
	\label{fig:parameter1}} 
\end{figure}

\begin{figure} [t!]
	\centering
		\includegraphics[scale=0.18]{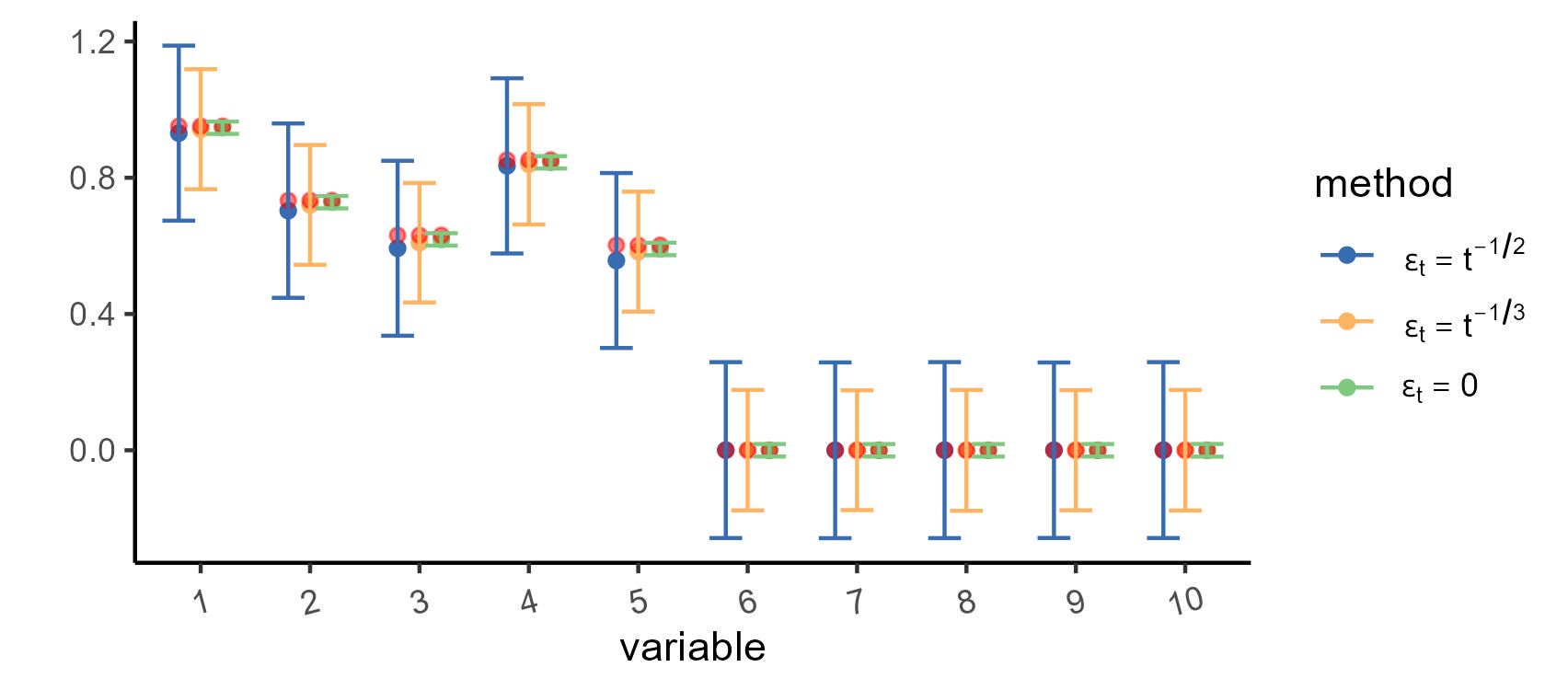}
    %}
	%\setcounter {subfigure} {0} {
		\includegraphics[scale=0.18]{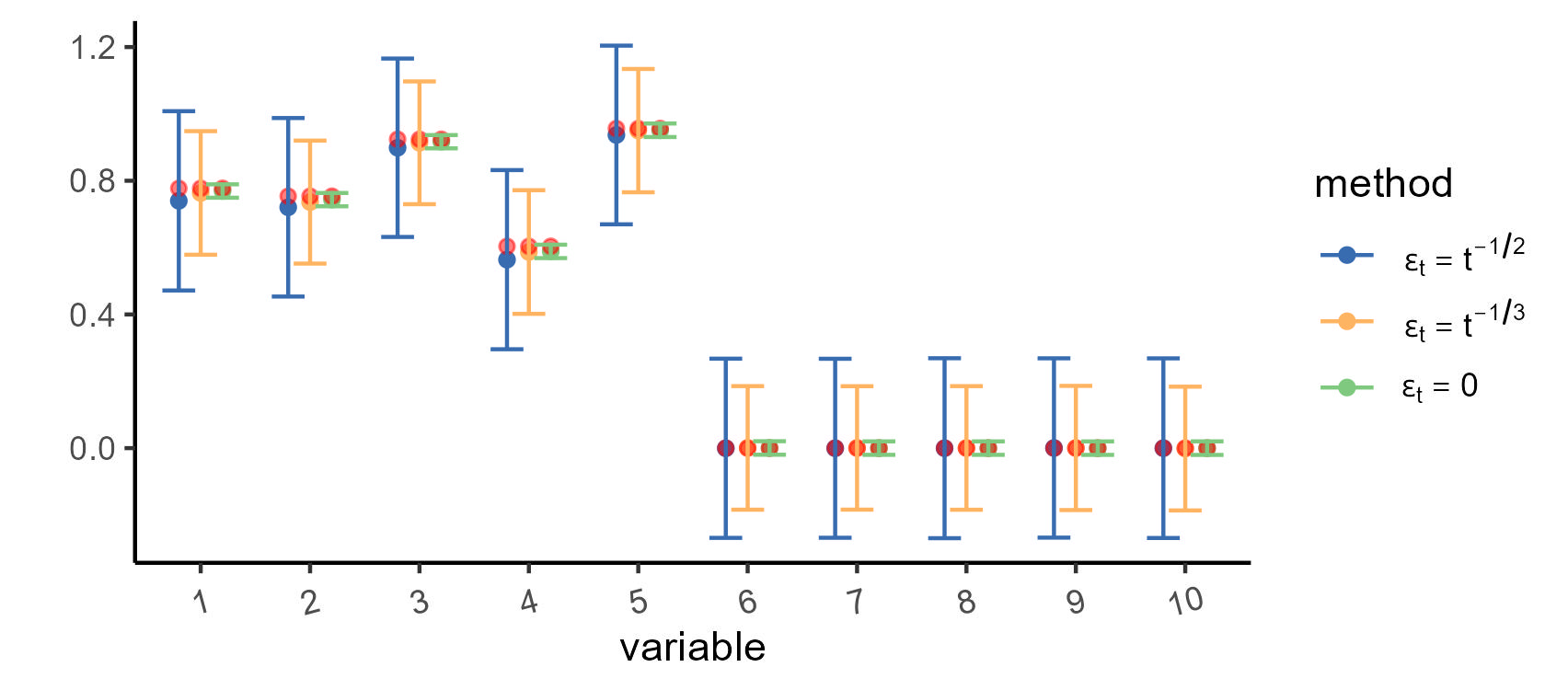}
    %}
    {\small\linespread{0.9}\selectfont
    \caption{Point and interval estimators of the first ten entries in $\beta_1$(top) and $\beta_0$(bottom) under scenario (2). The red points indicate the true value. }
    \label{fig:parameter2}}
\end{figure}

Figure \ref{fig:parameter1} and \ref{fig:parameter2} displays the point estimation and 95\% confidence interval of the first ten entries of $\beta_1$ and $\beta_0$ under the two different scenarios. It is evident that all three methods yield unbiased estimators, confirming the validity of our debiasing procedure. Furthermore, $\varepsilon_t\asymp t^{-1/2}$ results in the widest confidence intervals, while $\varepsilon_t=0$ produces the narrowest, aligning with our theoretical findings that the exploration free method leads to more efficient statistical inference. 

\begin{figure} [t!]
	\centering
		\includegraphics[scale=0.18]{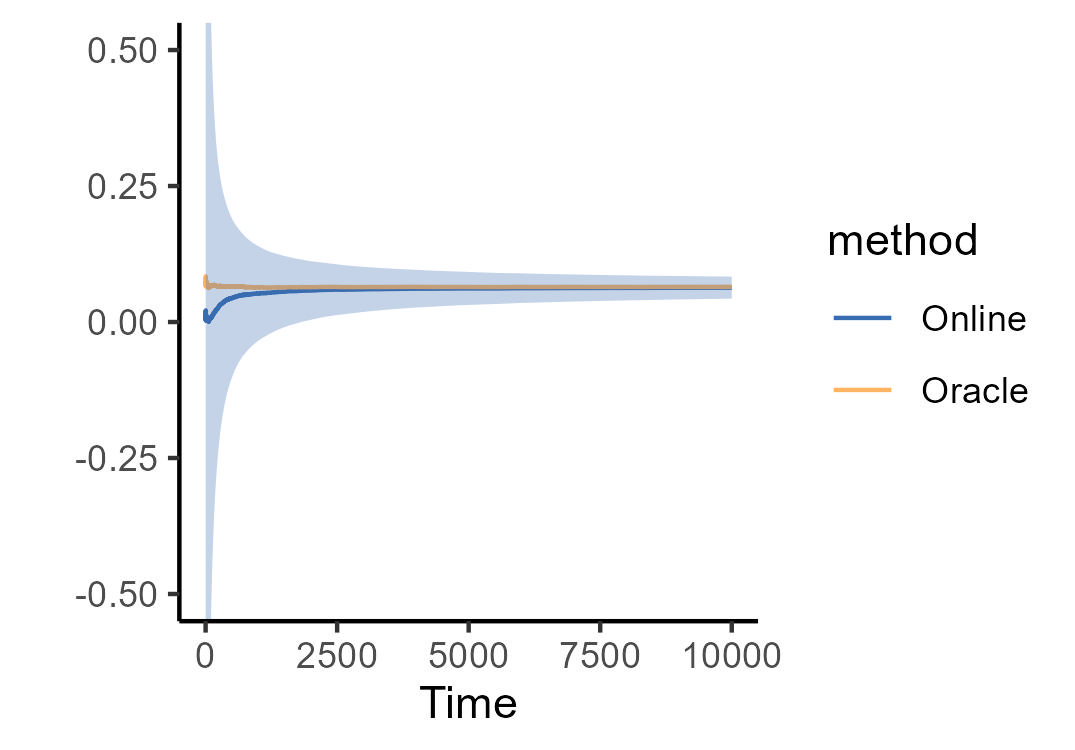}
    %}
	%\setcounter {subfigure} {0} {
		\includegraphics[scale=0.18]{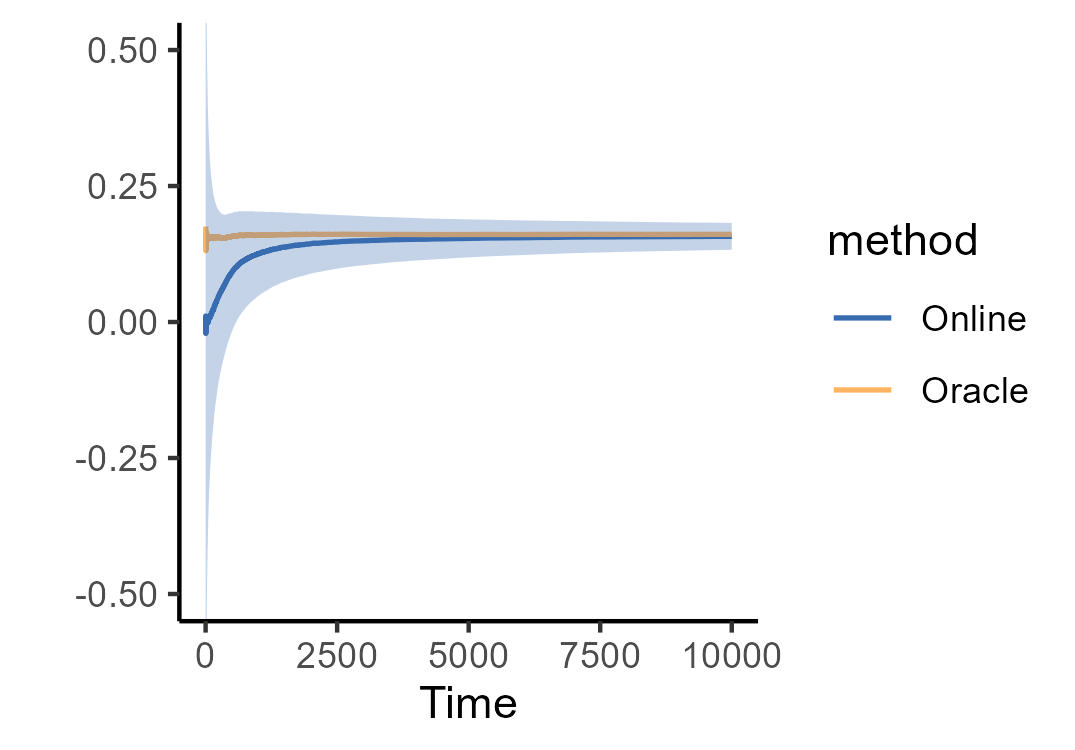}
    %}
    {\small\linespread{0.9}\selectfont
    \caption{Optimal value estimation by our online estimator(``Online") under scenario (1)(left panel) and (2)(right panel). The shaded region is the constructed 95\% confidence interval. ``Oracle" denotes the average of the true optimal value of each collected data.}
    \label{fig: simu-value}}
\end{figure}

Figure \ref{fig: simu-value} illustrates the point estimation and the 95\% confidence interval of the optimal value, compared with the average of the true optimal value of each collected data. It is obvious that our estimator converges to the oracle values in a short time, and the 95\% confidence intervals can always cover the oracle values.

\subsection{Real Data Analysis: Warfarin Dosing}\label{sec:real data}

%We apply our inference procedure for parameter and optimal value to a publicly available clinical trial dataset concerned Warfarin dosing. This dataset was originally collected by staffs at the Pharmacogenetics and Pharmacogenomics Knowledge Base (\url{https://www.pharmgkb.org/downloads}) and studied in \cite{international2009estimation}. Later, the data was used by many works including \cite{liu2015comparison,  bastani2020online, steiner2021machine}, to study the predictive pharmacogenetic dosing algorithms for Warfarin under several different machine learning or bandit models.  

\subsubsection{Data Set Description}

We apply our inference procedure for parameter and optimal value to a publicly available clinical trial dataset concerned Warfarin dosing. This dataset was originally collected by staffs at the Pharmacogenetics and Pharmacogenomics Knowledge Base (\url{https://www.pharmgkb.org/downloads}) and studied in \cite{international2009estimation}. Warfarin is the most widely used oral anticoagulant agent worldwide. Despite its widespread use, there are notable drawbacks, including a narrow therapeutic range and the potential for significant side effects. More importantly, plasma levels of anticoagulants are affected by numerous factors, which result in difficulties in dose determination. Given the high rate of adverse effects associated with incorrect dosing, there is interest in developing enhanced strategies to determine the optimal dosage.

In previous works, \cite{international2009estimation} first introduced a pharmacogenetic algorithm and demonstrated its superiority over the clinical algorithm. Subsequent studies by \cite{liu2015comparison, steiner2021machine} and \cite{bastani2020online} explored machine learning algorithms such as Support Vector Machine, boosted regression tree, artificial neural network, and high dimensional bandit algorithm. These studies collectively demonstrated that the proposed algorithms yielded more accurate dose predictions compared to the conventional doctor's policy. However, most of these studies primarily focused on dose prediction rather than identifying the factors that have the most significant impact on dosing. 

We begin with a brief description of the dataset. This data includes the true patient-specific optimal Warfarin doses, which are initially unknown but are eventually determined through a physician-guided dose adjustment process over the course of a few weeks for 5,528 patients. It also includes patient-level covariates that have been found to be predictive of the optimal Warfarin dosage \cite{international2009estimation}. To be specific, these covariates include demographics like gender, race, ethnicity, age, height, and weight; the reason for treatment, such as deep vein thrombosis or pulmonary embolism; pre-existing diagnoses, including indicators for conditions like diabetes, congestive heart failure or cardiomyopathy, valve replacement, and smoking status; medications, including indicators for potentially interacting drugs like aspirin, Tylenol, and Zocor; and most importantly, the presence of genotype variants of CYP2C9 and VKORC1. In total, there are $d=93$ patient-specific covariates, including indicators for missing values. 

\cite{international2009estimation} suggests the optimal dosage can be grouped into (1) Low: under 3mg/day (33\% of cases), (2) Medium: 3-7mg/day , and (3) High: over 7mg/day. Therefore, we extend our framework and formulate this problem into a three-armed bandit, and define arm $0, 1, 2$ to be Low dosage, Medium dosage and High dosage, respectively. For each patient, the reward is set to be 1 if the algorithm selects the arm corresponding to the patient’s true optimal dose; otherwise, the reward is 0. 

\subsubsection{Parameter and Value Inference}
Previous studies have reported that non-genetic factors such as age, height, weight, race, and drug interactions can account for approximately 20\% of the inter-individual variability Warfarin dosing (\cite{international2009estimation, liu2015comparison}). On the other hand, genetic factors, particularly CYP2C9 and VKORC1 genotypes are widely recognized as crucial predictors of Warfarin dose requirements across diverse populations worldwide. These genetic factors can explain 50-60\% of the variability in Warfarin dosage (\cite{international2009estimation, cosgun2011high}). Our primary objective is to investigate which variables among the 93 covariates significantly influence the inter-individual variability in Warfarin dosing.

We employ both IPW and AW estimators to derive the point and interval estimators for all the variables, and compare the results with the \textbf{Oracle} estimator, which is obtained through OLS linear regression. To achieve the environment with covariate diversity, we draw 700 optimal samples from each of the three arms, yielding a total sample size of $T=2100$. 

\begin{figure*} [t!]
	\centering
	\setcounter {subfigure} {0} {
		\includegraphics[scale=0.15]{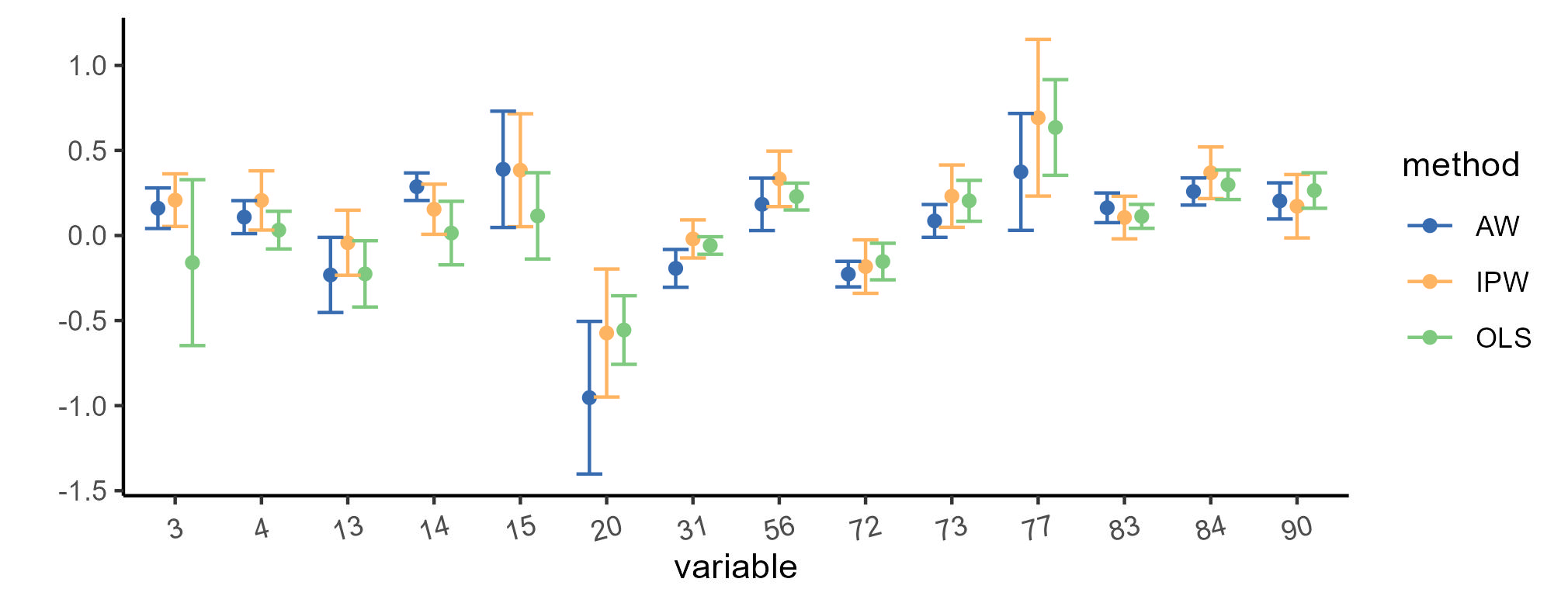}}
	\\
\setcounter {subfigure} {0} {
		\includegraphics[scale=0.15]{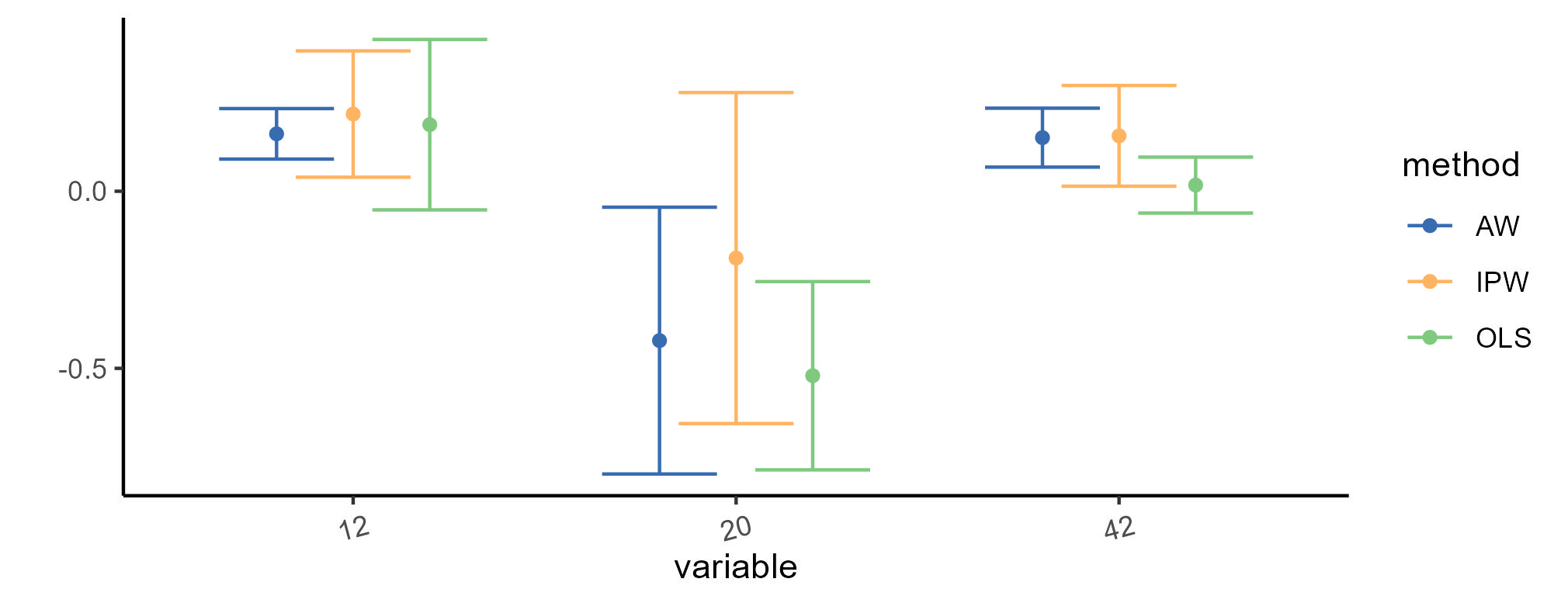} }
    \\    
\setcounter {subfigure} {0} {
		\includegraphics[scale=0.15]{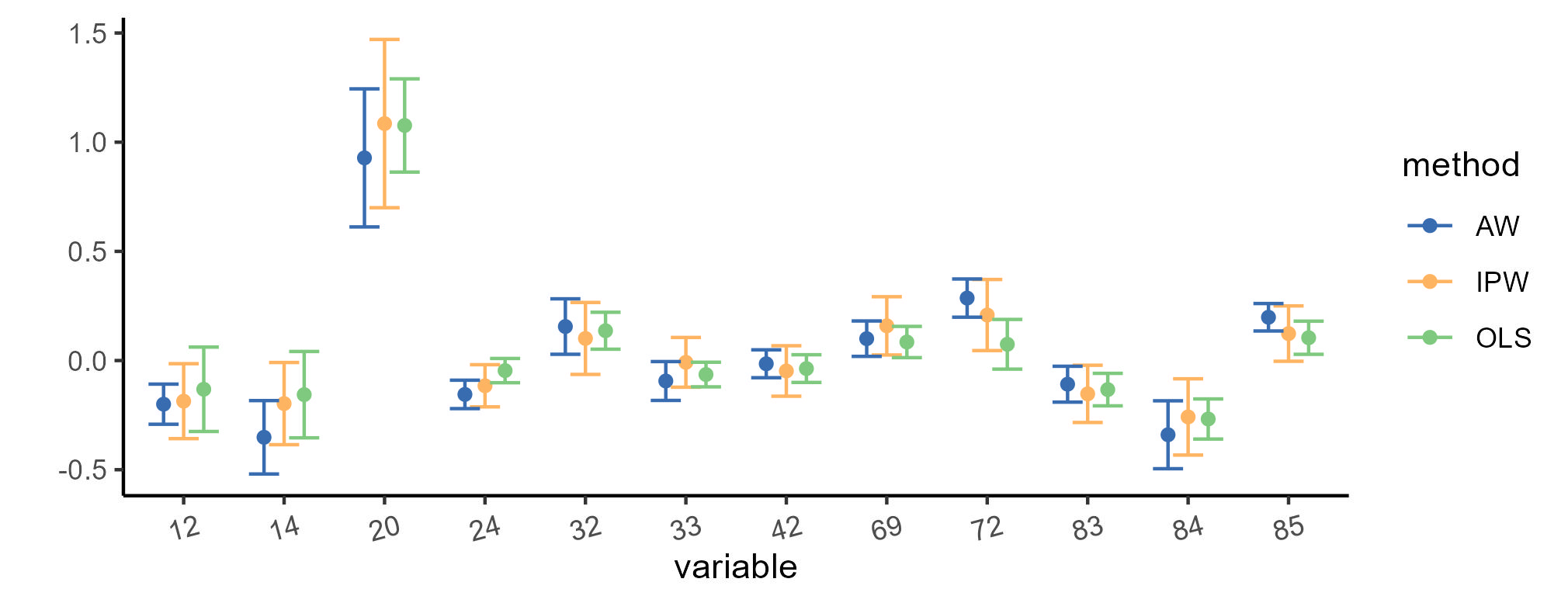}}
    {\small\linespread{0.9}\selectfont    
	\caption{Comparison of the point and interval estimators of the significant variables in $\beta_0$(top), $\beta_1$(middle) and $\beta_2$(bottom).}
	\label{fig:singlepara}} 
\end{figure*}

Figure \ref{fig:singlepara} shows the significant variables identified by at least two of the aforementioned methods. The regression coefficients associated with these factors have confidence intervals that do not cover zero, indicating their statistical significance. In general, both IPW and AW methods can correctly recognize the significance of these variables, producing point and interval estimators closed to those from the Oracle method. Moreover, we observe that, within the same sample size setting, the AW estimator generated more precise confidence intervals compared to the IPW estimator, which aligns with our theoretical findings. The variables identified significant for low and high dosages outnumber those for medium dosage, suggesting that most individuals are suitable for and safe with a medium dosage. Only those with particular characteristics are recommended for low or high doses, which is consistent with the doctor's policy in \cite{international2009estimation}, where medium dosages were prescribed for all patients.

\begin{figure*} [t!]
	\centering
	\setcounter {subfigure} {0} {
		\includegraphics[scale=0.18]{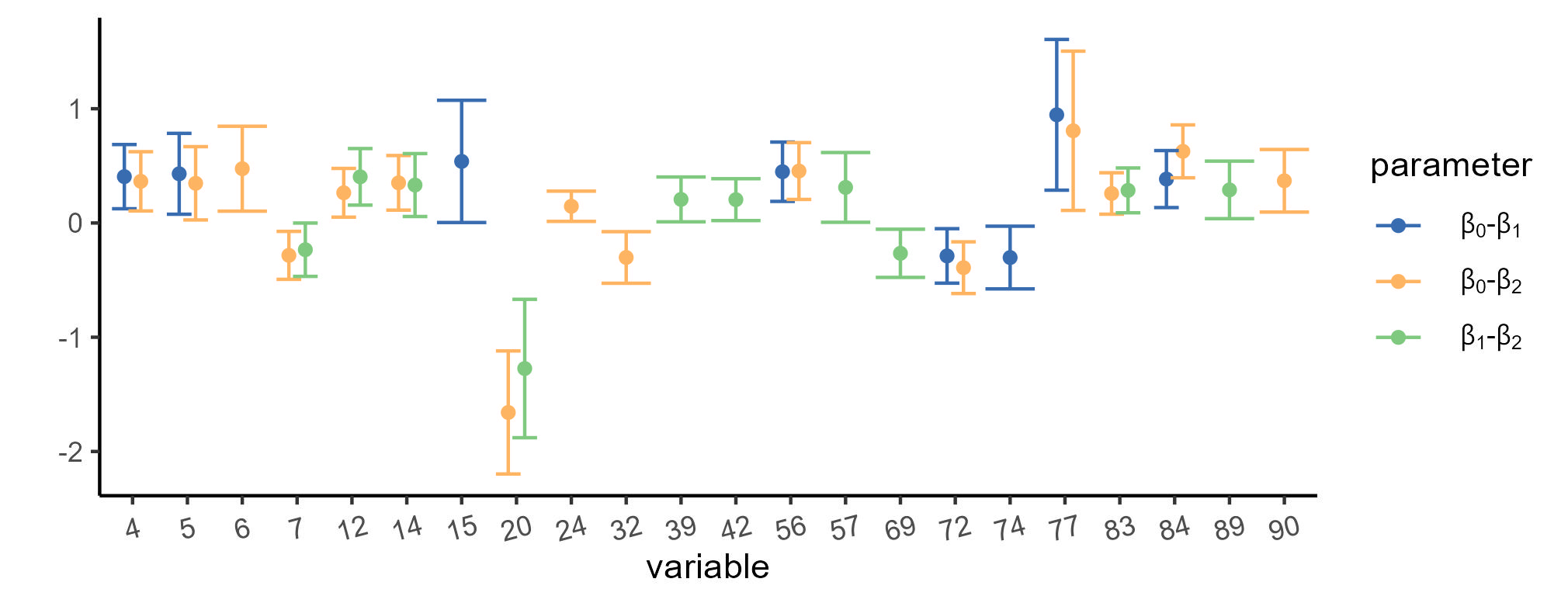}}
	\\ 
\setcounter {subfigure} {0} {
		\includegraphics[scale=0.18]{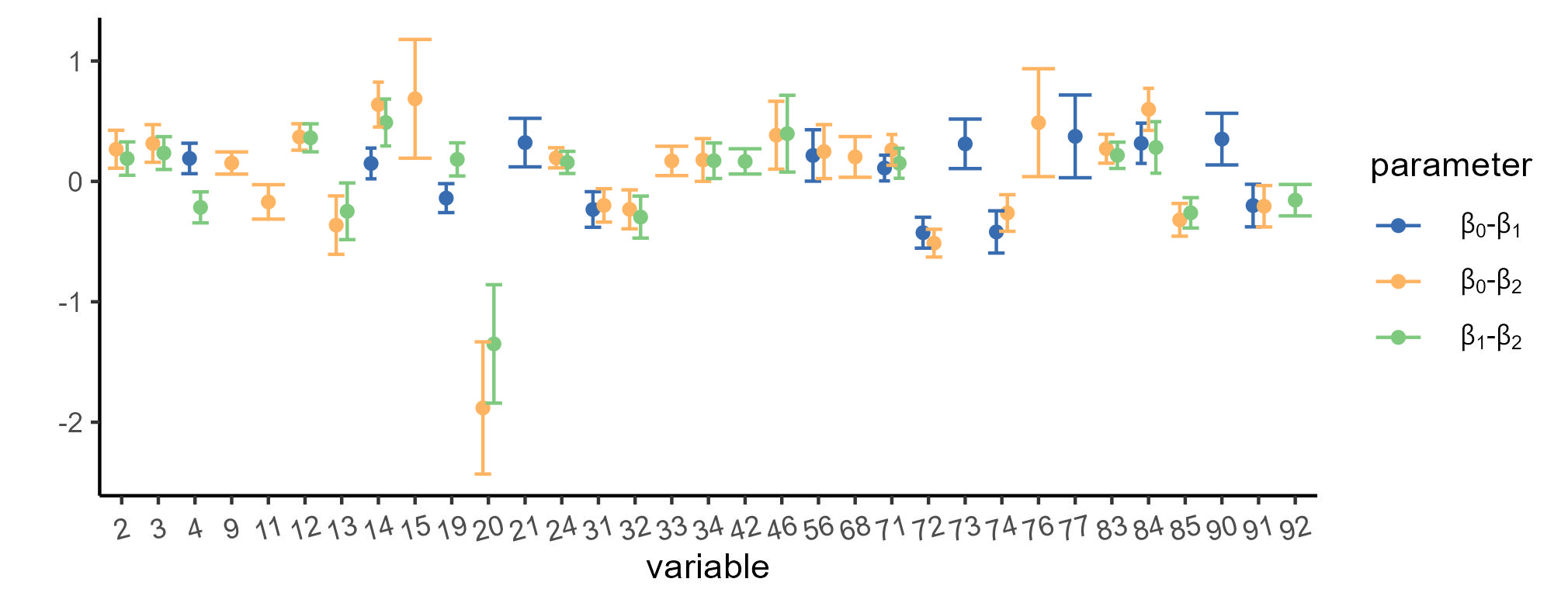}}
    {\small\linespread{0.9}\selectfont    
	\caption{Comparison of the point and interval estimators of the significant variables in $\beta_0-\beta_1$, $\beta_0-\beta_2$ and $\beta_1-\beta_2$ under $\varepsilon_t=5t^{-1/3}$(top) and $\varepsilon_t=0$(bottom).}
	\label{fig:differencepara} }
\end{figure*}

In comparison to individual parameters, we are more interested in the difference of the parameters, because it shows which variables influence and determine the varying dosages. Figure \ref{fig:differencepara} highlights all the significant variables in $\beta_0-\beta_1$, $\beta_0-\beta_2$ and $\beta_1-\beta_2$ identified by IPW and AW estimators. In general, the variables found significant by both methods exhibit similarities. Due to its lower variance, the AW estimator tends to detect more significant variables, as illustrated in Figure \ref{fig:differencepara}.

Among the variables identified as significant identified by both estimators, we first observe variable 12 and 14, denoting Age 70-79 and 80+ respectively, exhibit positive significance for $\beta_0-\beta_2$ and $\beta_1-\beta_2$. This suggests that older patients aged 70 and above are more suitable for lower doses. Similar conclusions have been presented in studies such as \cite{miura2009relationship}, that increased age increases the sensitivity to Warfarin, leading to a decreased required daily dosage. Variable 20, representing weight, displays negative significance for both $\beta_0-\beta_2$ and $\beta_1-\beta_2$, indicating that patients with higher weight are more likely to be assigned higher dosages. Studies like \cite{tellor2018evaluation} and \cite{alshammari2020warfarin} have also noted that obese patients require a higher dose of Warfarin compared to those with a normal BMI. Our analysis also highlights the iterations between other drugs and Warfarin. For instance, variable 56, corresponding to the use of Amiodarone (Cordarone), exhibits a positive significant effect for both $\beta_0-\beta_1$ and $\beta_0-\beta_2$, suggesting that the addition of Amiodarone necessitates a reduction in the Warfarin dosage. This correlation has been documented in studies like \cite{mcdonald2012warfarin}, that the interaction between Warfarin and Amiodarone results in an increased Warfarin effect, so a decrease in Warfarin is needed to further avoid severe unnecessary bleeding. In addition to the non-genetic factors and drug interactions, variable 71-92 represent the various genotypes of CYP2C9 and VKORC1, with many showing significant effects. For example, variable 72 displays negative significance for $\beta_0-\beta_1$ and $\beta_0-\beta_2$ via both methods, indicating that patients with Cyp2C9 genotype *1/*1 are more likely to receive medium and high dosages over low ones. In \cite{biss2012vkorc1}, it also stated that Warfarin daily dose requirement in patients with homozygous wild-type CYP2C9 genotype (*1/*1) is significantly higher than in those with *1/*3 , *1/*2, *2/*2, or *2/*3 genotype. Variable 83 pertains to VKORC1 -1639 consensus A/G, showing positive significance for $\beta_0-\beta_2$ and $\beta_1-\beta_2$, implying that patients with this genotype are more likely to receive medium and low dosages. \cite{henderson2019vkorc} also supports this finding, linking this genotype to dose reductions. Besides these variables, other factors identified significant by IPW or AW estimators can also be observed similar findings in the existing medical literature. 

\begin{figure*} [t!]
	\centering
	\includegraphics[scale=0.16]{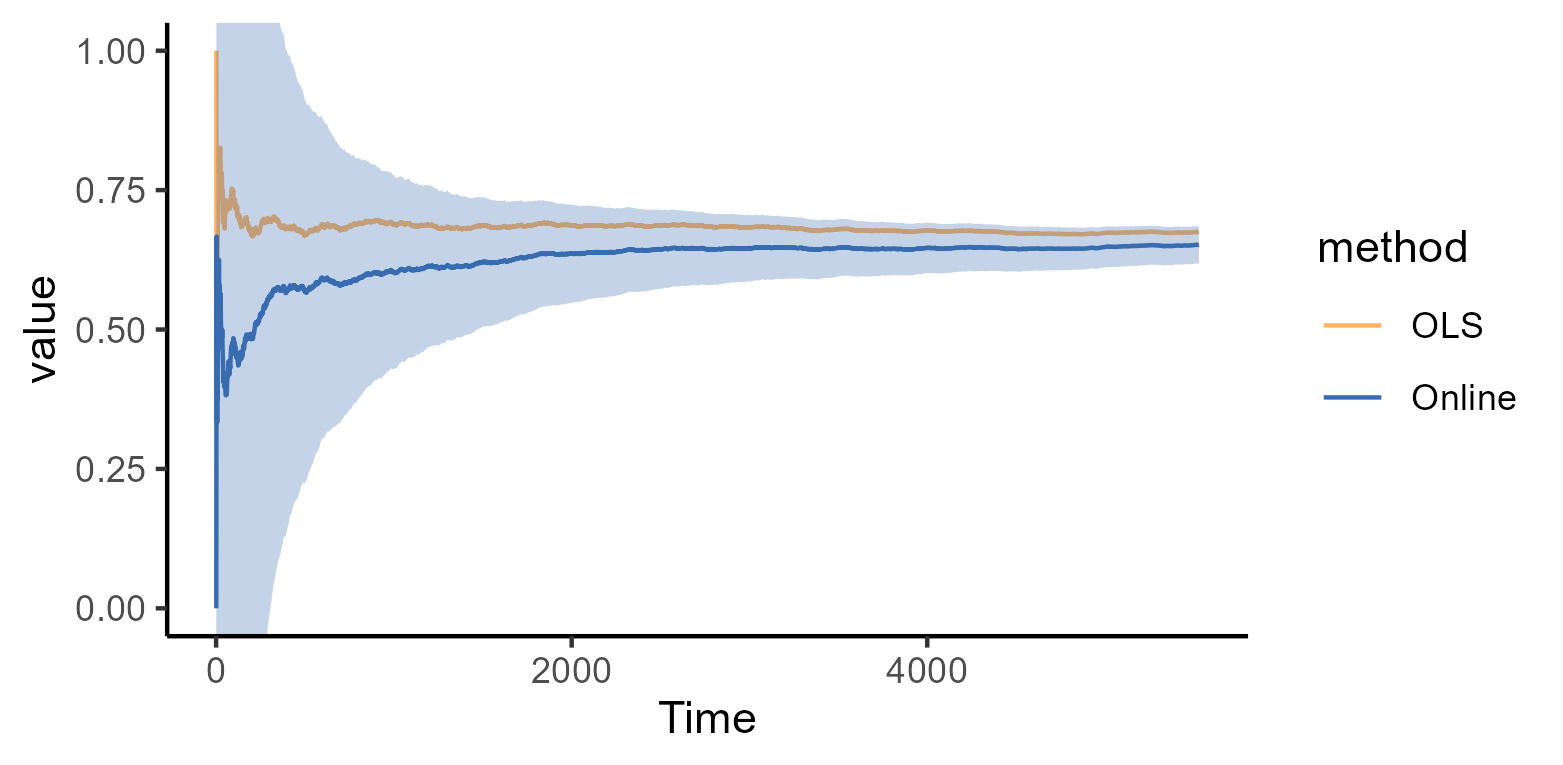}
    {\small\linespread{0.9}\selectfont
	\caption{Comparison of optimal value estiamtion under our online estimator and OLS. The shaded region is 95\% confidence interval constructed by our framework.}
	\label{fig:valueconverge} }
\end{figure*}

Finally, we evaluate our optimal value inference framework. This time, we utilize all $T=5525$ data to demonstrate the convergence of our optimal value estimator. Although the true optimal value is $V^*=1$, the data may not fully satisfy the assumptions required by our algorithm, particularly the covariate diversity. As a result, even the OLS estimator derived from the complete dataset does not achieve the nearly optimal value, with an estimated value around 0.67. We therefore plot our online estimator alongside the OLS estimator in Figure \ref{fig:valueconverge}, using the OLS estimator as a benchmark. It is apparent that our estimated value closely aligns with the OLS value, and the constructed 95\% confidence interval consistently covers the OLS estimate.

{
%\small 
\linespread{1.0}\selectfont 

\bibliographystyle{plainnat}

\bibliography{reference}

}

\newpage
\bigskip
\begin{center}
{\large\bf SUPPLEMENTARY MATERIAL to ``Regret Minimization and Statistical Inference in Online Decision Making with High-dimensional Covariates"}
\end{center}

\renewcommand\thesection{\Alph{section}} 
\setcounter{section}{0} 
%We provide additional simulation results and another real data example in Appendix \ref{app:numerical}. The extension to $k$-armed bandits, algorithmic acceleration using efficient SVD implementation, adaptation to non-uniform sampling distributions, and discussions of the upper confidence bound (UCB) and Thompson sampling (TS) algorithms are presented in Appendix \ref{app:extension}. All proofs and technical lemmas are included in Appendices \ref{app:proofs_thms} and \ref{app:prooflemma}.

\section{Additional Technical Lemmas} \label{sec: teclemma}

\begin{Lemma}
    \label{lemma:feasible1}
    Under Assumption \ref{assump:basic}, with probability at least $1-d^{-10}$, 
    \begin{align*}
        \|I_d-\Omega\widehat{\Sigma}_T\|_{\max}\lesssim \sqrt{\frac{\log d}{T}}.
    \end{align*}
\end{Lemma}
\begin{Lemma}
    \label{lemma:mlconvergence1}
    Consider the $M_t$ constructed by (\ref{eq:optimization1}) with $\mu_{T_1}=C_{\mu_1}\sqrt{\frac{\log d}{T}}$ for some constant $C_{\mu_1} >0$. Under the result of Lemma \ref{lemma:feasible1}, for any $l\in [d]$, with probability at least $1-d^{-10}$,
    \begin{align*}
        &\|m_{l,T} - \Omega e_l\| \lesssim \frac{\sqrt{s_{\Omega}}}{\phi_{\min}(s_{\Omega})}\mu_{T_1} \\
        &\|m_{l,T} - \Omega e_l\|_{\ell 1}\lesssim \frac{s_{\Omega}}{\phi_{\min}(s_{\Omega})}\mu_{T_1}.
    \end{align*}
\end{Lemma}
\begin{Lemma}
    \label{lemma:feasible2}
    Under Assumption \ref{assump:basic}, with probability at least $1-d^{-10}$, for $i=0,1$
    \begin{align*}
        \|I_d-\Lambda_i^{-1}\widehat{\Lambda}_{i,T}\|_{\max}\lesssim \left(\frac{s_0\log^2 d}{T}\right)^{\nu/2}.
    \end{align*}
\end{Lemma}
\begin{Lemma}
    \label{lemma:mlconvergence2}
    Consider the $M_t^{(i)}$ constructed by (\ref{eq:optimization2}) with $\mu_{T_2}=C_{\mu_2}\left(s_0\frac{\log d}{T}\right)^{\nu/2}$ for some constant $C_{\mu_2} >0$ and $i=0,1$. Under the result of Lemma \ref{lemma:feasible2}, for any $l\in [d]$, with probability at least $1-d^{-10}$,
    \begin{align*}
        &\|m_{l,T}^{(i)} - \Lambda_i^{-1}e_l\| \lesssim \frac{\sqrt{s_{\Lambda^{-1}}}}{\lambda_0}\mu_{T_2} \\
        &\|m_{l,T}^{(i)} - \Lambda_i^{-1}e_l\|_{\ell 1}\lesssim \frac{s_{\Lambda^{-1}}}{\lambda_0}\mu_{T_2}.
    \end{align*}
\end{Lemma}

\section{Proofs of Main Results}

\subsection{Proof of Theorem \ref{thm:estimation-nodc}}
\begin{proof}
    The proof follows the same arguments as the proof of Theorem 1 in \cite{ma2024high}. Without loss of generality, we only prove $i=1$ case and omit all the subscript. We first denote $\widetilde{\beta}=\widehat{\beta}_{t-1}-\eta g_t$, and the support $B=B_{t+1}\cup B_t\cup B_{*}$ as the union of the support set $\widehat{\beta}_{t+1}$, $\widehat{\beta}_{t}$ and $\beta$. For iterative method, we have
    \begin{align*}
        \|\widehat{\beta}_t\|^2=\|\mathcal{H}_s(B(\widetilde{\beta}_t)) - \beta\|^2\leq \left(1+\frac{\rho+\sqrt{\rho(4+\rho)}}{2}\right)\|B(\widetilde{\beta}_t) - \beta\|^2,
    \end{align*}
    by the tight bound of hard thresholding operator. Here $\rho=s_0/s$ is the relative sparsity level. By selecting a small $\rho$ (e.g. $\rho\leq 1/4$), it is clear that
    \begin{align*}
        \|\widehat{\beta}_t - \beta\|^2 &\leq \left(1+\frac{3}{2}\sqrt{\rho} \right)\|B(\widetilde{\beta}_t) - \beta-\beta\|^2 \\
        &= \left(1+\frac{3}{2}\sqrt{\rho} \right)\left(\|\widehat{\beta}_{t-1} - \beta\|^2 -2\eta\inp{B(g_t)}{\widehat{\beta}_{t-1} - \beta}+ \eta^2\|B(g_t)\|^2\right) \\
        &\leq \left(1+\frac{3}{2}\sqrt{\rho} \right) \left(\|\widehat{\beta}_{t-1} - \beta\|^2 - 2\eta\inp{\nabla f(\widehat{\beta}_{t-1})}{\widehat{\beta}_{t-1} - \beta} + 2\eta^2\|B(g_t-\nabla f(\widehat{\beta}_{t-1}))\|^2 \right.\\ &\left.\quad + 2\eta^2\|B(\nabla f(\widehat{\beta}_{t-1}))\|^2 + 2\eta\|B(g_t-\nabla f(\widehat{\beta}_{t-1}))\|\|\widehat{\beta}_{t-1} - \beta\| \right),
    \end{align*}
    where we use the fact that $\inp{\nabla f(\widehat{\beta}_{t-1})}{\widehat{\beta}_{t-1} - \beta}=\inp{B(\nabla f(\widehat{\beta}_{t-1}))}{\widehat{\beta}_{t-1} - \beta}$ by the definition of $B(\cdot)$. Then by Assumption \ref{assump:basic},
    \begin{align*}
        &\inp{\nabla f(\widehat{\beta}_{t-1})}{\widehat{\beta}_{t-1} - \beta}\geq 2\phi_{\min}(s)\|\widehat{\beta}_{t-1} - \beta\|^2 \\
        & \|B(\nabla f(\widehat{\beta}_{t-1}))\|\leq 2\phi_{\max}(s)\|\widehat{\beta}_{t-1} - \beta\|.
    \end{align*}
    We can show that
    \begin{align*}
        \|\widehat{\beta}_t - \beta\|^2&\leq \left(1+\frac{3}{2}\sqrt{\rho} \right)(1-4\eta\phi_{\min}(s) + 8\eta^2\phi_{\max}^2(s))\|\widehat{\beta}_{t-1} - \beta\|^2 \\
        &\quad + 6\eta^2\|B(g_t-\nabla f(\widehat{\beta}_{t-1}))\|^2 + 6\eta\|B(g_t-\nabla f(\widehat{\beta}_{t-1}))\|\|\widehat{\beta}_{t-1} - \beta\| \\
        &\leq \left(1+\frac{3}{2}\sqrt{\rho} \right)(1-4\eta\phi_{\min}(s) + 8\eta^2\phi_{\max}^2(s))\|\widehat{\beta}_{t-1} - \beta\|^2 \\ 
        &\quad + 18s\eta^2\max_{l\in [d]}|\inp{g_t-\nabla f(\widehat{\beta}_{t-1})}{e_l}|^2 + 18\eta\sqrt{s}\max_{l\in [d]}|\inp{g_t-\nabla f(\widehat{\beta}_{t-1})}{e_l}|\|\widehat{\beta}_{t-1} - \beta\|.
    \end{align*}
    The following Lemma quantifies the variation of the stochastic gradient:
    \begin{Lemma}
        \label{lemma: sgd}
        Define $\{e_l\}_1^d$ as the canonical basis of $\RR^d$. When choosing $\varepsilon_t\asymp t^{-\gamma}$ for $\gamma\in [0,1)$ in Algorithm \ref{alg:onlineHT}, then with probability at least $1-d^{-10}$,
        \begin{align*}
            \max_{l\in [d]}|\inp{g_t-\nabla f(\widehat{\beta}_{t-1})}{e_l}|^2\leq CsD^2\frac{\log(d)}{t^{1-\gamma}}\|\widehat{\beta}_{t-1} - \beta\|^2 + C\frac{\sigma^2D^2\log (d)}{t^{1-\gamma}}.
        \end{align*}
    \end{Lemma}
    With Lemma \ref{lemma: sgd}, 
    \begin{align*}
        \|\widehat{\beta}_t - \beta\|^2&\leq \left(1-\frac{1}{4\kappa_{*}^4}+ C\frac{s_0D\sqrt{\log (d)}}{\phi_{\min}(s)t^{(1-\gamma)/2}}\right)\|\widehat{\beta}_{t-1} - \beta\|^2 \\
        &\quad + C\frac{s_0\sigma^2D^2\log (d)}{\phi_{\min}^2(s)t^{1-\gamma}} + C\sqrt{\frac{s_0\sigma^2D^2\log (d)}{\phi_{\min}^2(s)t^{1-\gamma}}}\|\widehat{\beta}_{t-1} - \beta\|,
    \end{align*}
    for all $t\in [T]$. When $t$ is large, we have
    \begin{align*}
        \|\widehat{\beta}_t - \beta\|^2\leq \left(1-\frac{1}{5\kappa_{*}^4}\right)\|\widehat{\beta}_{t-1} - \beta\|^2 + C\frac{s_0\sigma^2D^2\log (d)}{\phi_{\min}^2(s)t^{1-\gamma}} + C\sqrt{\frac{s_0\sigma^2D^2\log (d)}{\phi_{\min}^2(s)t^{1-\gamma}}}\|\widehat{\beta}_{t-1} - \beta\|
    \end{align*}
    then we can derive the desired result by induction. 
\end{proof}

\subsection{Proof of Theorem \ref{thm:regret-nodc}}
\begin{proof}
    Without loss of generality, assume at time $t$, arm 1 is optimal, i.e., $\inp{\beta_1-\beta_0}{X_t}>0$, then the expected regret is
\begin{align*}
    r_t&=\EE\left[\inp{\beta_1-\beta_0}{X_t}\indicator(\text{choose arm 0}) \right] \\
    &= \EE\left[\inp{\beta_1-\beta_0}{X_t}\indicator(\text{choose arm 0}, \inp{\widehat{\beta}_{1,t-1}-\widehat{\beta}_{0,t-1}}{X_t}\geq 0)\right] \\ &\quad + \EE\left[\inp{\beta_1-\beta_0}{X_t}\indicator(\text{choose arm 0}, \inp{\widehat{\beta}_{0,t-1}-\widehat{\beta}_{1,t-1}}{X_t}> 0)\right] \\
    &\leq \frac{\varepsilon_t}{2}R_{\max} + \underbrace{\EE\left[\inp{\beta_1-\beta_0}{X_t}\indicator(\inp{\widehat{\beta}_{0,t-1}-\widehat{\beta}_{1,t-1}}{X_t}> 0)\right]}_{(2)}.
\end{align*}
By Assumption \ref{assump:basic}(c) and Theorem \ref{thm:estimation-nodc}, with probability at least $1-d^{-100}$, $\max_{X\in \mathcal{X}}\|\inp{\beta_1-\widehat{\beta}_{1,t-1}}{X}\|_{\max}$ is bounded by $C_1\frac{\sigma_1\sqrt{s_0}\log d}{\phi_{\min}(s)(t-1)^{(1-\gamma)/2}}$ for some constant $C_1$, and same for $\max_{X\in \mathcal{X}}\|\inp{\beta_0-\widehat{\beta}_{0,t-1}}{X}\|_{\max}$. Denote $\delta_t=C_1\frac{\sigma_1\sqrt{s_0}\log d}{\phi_{\min}(s)(t-1)^{(1-\gamma)/2}}$, and event $B_t=\{\inp{\beta_1-\beta_0}{X_t}>2\delta_t\}$, then the latter expectation can be written as
\begin{align*}
    (2)&\leq \EE\left[\inp{\beta_1-\beta_0}{X_t}\indicator\left(\inp{\widehat{\beta}_{0,t-1}-\widehat{\beta}_{1,t-1}}{X_t}> 0\cap B_t\right)\right] \\ &\quad + \EE\left[\inp{\beta_1-\beta_0}{X_t}\indicator\left(\inp{\widehat{\beta}_{0,t-1}-\widehat{\beta}_{1,t-1}}{X_t}> 0\cap B_t^c\right)\right] \\
    &\leq R_{\max} \EE\left[\indicator\left(\inp{\widehat{\beta}_{0,t-1}-\widehat{\beta}_{1,t-1}}{X_t}> 0\cap B_t\right)\right] + 2\delta_t\EE\left[\indicator(B_t^c)\right].
\end{align*}
Under event $B_t$ and $\inp{\widehat{\beta}_{0,t-1}-\widehat{\beta}_{1,t-1}}{X_t}> 0$, 
\begin{align*}
    0>\inp{\widehat{\beta}_{1,t-1}-\widehat{\beta}_{0,t-1}}{X_t} &= \inp{\widehat{\beta}_{1,t-1}-\beta_1}{X_t} + \inp{\beta_0-\widehat{\beta}_{0,t-1}}{X_t} + \inp{\beta_1-\beta_0}{X_t}\\
    &> \inp{\widehat{\beta}_{1,t-1}-\beta_1}{X_t} + \inp{\beta_0-\widehat{\beta}_{0,t-1}}{X_t} + 2\delta_t.
\end{align*}
This means either $\inp{\widehat{\beta}_{1,t-1}-\beta_1}{X_t}<-\delta_t$ or $\inp{\beta_0-\widehat{\beta}_{0,t-1}}{X_t}<-\delta_t$, which further implies either $\max |\inp{\beta_0-\widehat{\beta}_{0,t-1}}{X}|$ or $\max |\inp{\beta_0-\widehat{\beta}_{0,t-1}}{X}|$ should be larger than $\delta_t$. Therefore, 
\begin{align*}
    \EE\left[\indicator(\inp{\widehat{\beta}_{0,t-1}-\widehat{\beta}_{1,t-1}}{X_t}> 0)\cap B_t\right]&= \PP\left(\inp{\widehat{\beta}_{0,t-1}-\widehat{\beta}_{1,t-1}}{X_t}> 0 \cap B_t\right) \\
    &\leq \PP\left(\max |\inp{\beta_1-\widehat{\beta}_{1,t-1}}{X}|>\delta_t\right) + \PP\left(\max |\inp{\beta_0-\widehat{\beta}_{0,t-1}}{X}|>\delta_t\right) \\
    &\leq \frac{2}{d^{100}},
\end{align*}
For the other term, by Assumption \ref{assump:margincondition}, $\EE\left[\indicator(B_t^c)\right]=\PP(\inp{\beta_1-\beta_0}{X_t}<2\delta_t)\leq 2C_0\delta_t^{\nu}$. Then
\begin{align*}
    r_t&\lesssim t^{-\gamma}\sqrt{s_0}D + \frac{2}{d^{100}}R_{\max}  + 4C_0\delta_t^{1+\nu}. 
\end{align*}
The total regret up to time $T$ is
\begin{align*}
    R_T&=\sum_{t=1}^{T} r_t \lesssim \sum_{t=1}^T t^{-\gamma}R_{\max}  + \frac{T}{d^{100}}R_{\max} + \sum_{t=1}^{T} \delta_t^{1+\nu} \\
    &\lesssim T^{1-\gamma}R_{\max} + \phi_{\min}(s)^{-(1+\nu)}s_0^{(1+\nu)/2}T^{(\gamma-1)(1+\nu)/2+1}\log^2d.
\end{align*}
\end{proof}

\subsection{Proof of Theorem \ref{thm:maininference-nodc}}
\begin{proof}
    Without loss of generality, we only prove $i=1$ case and omit all the subscript. $i=0$ can be proved following the same arguments. Recall that for any $l\in [d]$,
    \begin{align*}
        \sqrt{T^{1-\gamma}}(\widehat{\beta}_{(l)}^{\uipw} - \beta_{(l)})&= \underbrace{\frac{1}{\sqrt{T^{1+\gamma}}}e_l^{\top}\sum_{t=1}^{T} \frac{\indicator(a_t=1)}{\pi_t}\Omega X_t\xi_t}_{V_T} + \underbrace{\frac{1}{\sqrt{T^{1+\gamma}}}e_l^{\top}\sum_{t=1}^{T} \frac{\indicator(a_t=1)}{\pi_t}(M_T-\Omega) X_t\xi_t}_{R_1} \\ &\quad \underbrace{\frac{1}{\sqrt{T^{1+\gamma}}}e_l^{\top}\sum_{t=1}^{T} \left(I-\frac{\indicator(a_t=1)}{\pi_t}M_{t-1}X_tX_t^{\top}\right)(\widehat{\beta}_{T} - \beta)}_{R_2}.
    \end{align*}
    The next Lemma shows $R_1+R_2$ are negligible bias terms,
    \begin{Lemma}
        \label{lemma:bias1}
        Under the conditions in Theorem \ref{thm:maininference-nodc},
        \begin{align*}
            \|R_1+R_2\|_{\max}=o_p(1).
        \end{align*}
    \end{Lemma}

    Next, we apply Theorem 3.2 and Corollary 3.1 in \cite{hall2014martingale}, the Martingale Central Limit Theorem to show the asymptotic normality of $V_T$. \\
    \emph{Step 1: checking Lindeberg Condition} \\
    For any $\delta>0$,
    \begin{align*}
        &\sum_{t=1}^{T} \EE\left[\frac{1}{T^{1+\gamma}}\left(e_{l}^{\top}\frac{\indicator(a_t=1)}{\pi_t^2}\Omega X_t\xi_t\right)^2 \indicator\left(\bigg|\frac{1}{\sqrt{T^{1+\gamma}}}e_{l}^{\top}\frac{\indicator(a_t=1)}{\pi_t}\Omega X_t\xi_t \bigg|>\delta\right)\bigg| \mathcal{F}_{t-1}\right] \\
        &\leq \frac{1}{T^{1+\gamma}}\sum_{t=1}^{T}\EE\left[\frac{\indicator(a_t=1)}{\pi_t^2} e_{l}^{\top}\Omega X_tX_t^{\top}\Omega^{\top}e_l\xi_t^2 \indicator\left(\big|e_{l}^{\top}\Omega X_t\xi_t \big|>\delta\sqrt{T^{1+\gamma}}\frac{\varepsilon_t}{2} \right)\bigg| \mathcal{F}_{t-1}\right] \\
        &\leq \frac{\sigma^2}{T^{1+\gamma}}\sum_{t=1}^{T} \frac{2}{\varepsilon_t}\max_{X\in \mathcal{X}} \|e_{l}^{\top}\Omega X\|^2 \times \EE\left[\indicator\left(\big|e_{l}^{\top}\Omega X_t\xi_t \big|>\delta\sqrt{T^{1+\gamma}}\frac{\varepsilon_t}{2} \right)\right].
    \end{align*}
    For simplicity, we denote $m_l=\Omega e_l$. Since $\xi_t$ is a subGaussian random variable, we denote event $\mathcal{E}_{1,t}=\{|\xi_t|\leq c_1\sigma\log^{1/2}d\}$ for some constant $c_1$ with $\PP(\mathcal{E}_{1,t})\geq 1 - d^{-20}$. Also notice that $|m_l^{\top}X_t|\leq \|m_l\|_{\ell 1}\|X_t\|_{\max}\leq \sqrt{s_{\Omega}}C_{\min}D$. Then as long as $s_{\Omega} \leq \frac{c_2T^{1-\gamma}}{\log d}$ for small enough constant $c_2$, we have $|m_l^{\top}X_t|\leq \frac{\delta}{c_1}\sqrt{\frac{T^{1+\gamma}}{\log d\sigma^2}}\frac{\varepsilon_t}{2} $ for any $t\geq 2$.
    
     As a result,
    \begin{align*}
        \PP\left(\big|e_{l}^{\top}\Omega X_t\xi_t \big|>\delta\sqrt{T^{1+\gamma}}\frac{\varepsilon_t}{2}\right)&\leq \PP\left(|\xi_t|\geq c_1\sigma\log^{1/2}d\right) = \PP(\mathcal{E}_{1,t}^c)\leq d^{-20}.
    \end{align*}
    Therefore, by the fact that $\sum_{t=1}^{T}\frac{2}{\varepsilon_t}\lesssim T^{1+\gamma}$ and $\sum_{t=2}^{T}\frac{2}{\varepsilon_t}\sqrt{\frac{1}{t-1}}\lesssim T^{\gamma+\frac{1}{2}}$,
    \begin{align*}
        &\frac{\sigma^2}{T^{1+\gamma}}\sum_{t=1}^{T}\frac{2}{\varepsilon_t} \max_{X\in \mathcal{X}} \|e_{l}^{\top}\Omega X\|^2 \times \EE\left[\indicator\left(\big|e_{l}^{\top}\Omega X_t\xi_t \big|>\delta\sqrt{T}\frac{\varepsilon_t}{2} \right)\right]\\ &\quad \lesssim \max_{X\in \mathcal{X}} \|e_{l}^{\top}\Omega X\|^2 d^{-20}\rightarrow 0,
    \end{align*}
    and the Lindeberg condition is satisfied. \\
    \emph{Step 2: calculating the variance}
    \begin{align*}
        &\frac{1}{T^{1+\gamma}}\sum_{t=1}^{T} \EE\left[(e_{l}^{\top}\frac{\indicator(a_t=1)}{\pi_t^2}\Omega X_t\xi_t)^2|\mathcal{F}_{t-1}\right] \\
        &\quad = \frac{\sigma^2}{T^{1+\gamma}}\sum_{t=1}^{T} \EE\left[\frac{1}{\pi_t}e_l^{\top}\Omega X_tX_t^{\top}\Omega^{\top}e_l|\mathcal{F}_{t-1}\right] \\
        &\quad = \frac{\sigma^2}{T^{1+\gamma}}\sum_{t=1}^{T} m_{l}^{\top}\EE\left[\frac{1}{\pi_t}X_tX_t^{\top}|\mathcal{F}_{t-1}\right] m_{l}.
    \end{align*}
    $\EE\left[\frac{1}{\pi_t}X_tX_t^{\top}|\mathcal{F}_{t-1}\right]$ can be decomposed into
    \begin{align*}
        \EE\left[\frac{1}{\pi_t}X_tX_t^{\top}|\mathcal{F}_{t-1}\right] &= \frac{1}{1-\frac{\varepsilon_t}{2}}\EE\left[X_tX_t^{\top}\indicator(\inp{\widehat{\beta}_{1,t-1} - \widehat{\beta}_{0,t-1}}{X_t}>0)|\mathcal{F}_{t-1}\right] \\ &\quad + \frac{2}{\varepsilon_t}\EE\left[X_tX_t^{\top}\indicator(\inp{\widehat{\beta}_{1,t-1} - \widehat{\beta}_{0,t-1}}{X_t}<0)|\mathcal{F}_{t-1}\right] \\
        &:= \frac{1}{1-\frac{\varepsilon_t}{2}}\Lambda_{1,t} + \frac{2}{\varepsilon_t}\Lambda_{0,t}.
    \end{align*}
    The next Lemma shows the upper bound of $\|\Lambda_{i,t} - \Lambda_i\|_{\max}$ and $|m_l^{\top}(\Lambda_{i,t} - \Lambda_i)m_l|$.
    \begin{Lemma}
        \label{lambdaconvergence1}
        Under the conditions in Theorem \ref{thm:maininference-nodc}, for $i=1,0$, with probability at least $1-d^{-10}$, we have
        \begin{align*}
            \|\Lambda_{i,t} - \Lambda_i\|_{\max}\lesssim \left(\frac{s_0\log d}{(t-1)^{1-\gamma}}\right)^{\nu/2} \quad \text{and} \quad |m_l^{\top}(\Lambda_{i,t} - \Lambda_i)m_l|\lesssim \left(\frac{s_0\log^3 d}{t-1}\right)^{\nu/2}.
        \end{align*}
    \end{Lemma}
    Therefore, 
    \begin{align*}
        &\quad \left| \frac{1}{T^{1+\gamma}}\sum_{t=1}^{T} m_{l}^{\top}\EE\left[\frac{1}{\pi_t}X_tX_t^{\top}|\mathcal{F}_{t-1}\right] m_{l} - \frac{1}{T^{1+\gamma}}\sum_{t=1}^{T} \left(\frac{1}{1-\frac{\varepsilon_t}{2}}m_l^{\top}\Lambda_1 m_l + \frac{2}{\varepsilon_t}m_l^{\top}\Lambda_0 m_l\right)\right| \\
        &\lesssim \frac{1}{T^{1+\gamma}} \sum_{t=2}^{T} \left(\frac{s_0\log^3 d}{(t-1)^{1-\gamma}}\right)^{\nu/2}+ \frac{1}{T^{1+\gamma}} \sum_{t=2}^{T}  \frac{2}{\varepsilon_t}\left(\frac{s_0\log^3 d}{(t-1)^{1-\gamma}}\right)^{\nu/2} \lesssim \sqrt{\frac{s_0^{\nu}\log^{3\nu} d}{T^{\nu(1-\gamma)}}}.
    \end{align*}
    Then the above equation indicates \\ $\frac{1}{T^{1+\gamma}}\sum_{t=1}^{T} m_{l}^{\top}\EE\left[\frac{1}{\pi_t}X_tX_t^{\top}|\mathcal{F}_{t-1}\right] m_{l}\overset{p}{\rightarrow} \frac{1}{T^{1+\gamma}}\sum_{t=1}^{T} \left(\frac{1}{1-\frac{\varepsilon_t}{2}}m_l^{\top}\Lambda_1 m_l + \frac{2}{\varepsilon_t}m_l^{\top}\Lambda_0 m_l\right)$ when $T,d_1,d_2\rightarrow \infty$ as long as $s_0=o(\frac{T^{1-\gamma}}{\log^3 d})$.  

    Since $\frac{1}{T^{1+\gamma}}\sum_{t=1}^{T} \left(\frac{1}{1-\frac{\varepsilon_t}{2}}m_l^{\top}\Lambda_1 m_l + \frac{2}{\varepsilon_t}m_l^{\top}\Lambda_0 m_l\right)\rightarrow \frac{1}{T^{\gamma}}m_l^{\top}\Lambda_1 m_l + \frac{2}{c_2(1+\gamma)}m_l^{\top}\Lambda_0 m_l$ when $T\rightarrow \infty$, we can conclude that $\frac{1}{T^{1+\gamma}}\sum_{t=1}^{T} m_{l}^{\top}\EE\left[\frac{1}{\pi_t}X_tX_t^{\top}|\mathcal{F}_{t-1}\right] m_{l} \overset{p}{\rightarrow} \frac{1}{T^{\gamma}}m_l^{\top}\Lambda_1 m_l + \frac{2}{c_2(1+\gamma)}m_l^{\top}\Lambda_0 m_l$. 

\end{proof}

\subsection{Proof of Theorem \ref{thm:stuinference-nodc}}
\begin{proof}
    Denote $m_l=\Omega e_l$. We need to prove for any $l\in [d]$, $i=0,1$, $\widehat{\sigma}_i^2$ and $m_{l,T}^{\top}\widehat{\Lambda}_{i,T}m_{l,T}$ are consistent estimators for $\sigma_i^2$ and $m_{l}^{\top}\Lambda_i m_l$, respectively. Without loss of generality, we only prove $i=1$ case and omit all the subscript $i$. \\
    {\it Step 1: showing $\widehat{\sigma}^2\overset{p}{\rightarrow} \sigma^2$.} \\
    Notice that, $\widehat{\sigma}^2$ can be written as
    \begin{align*}
        \widehat{\sigma}^2&= \underbrace{\frac{1}{T}\sum_{t=1}^{T} \frac{\indicator(a_t=1)}{\pi_t}\inp{\beta-\widehat{\beta}_{t-1}}{X_t}^2}_{\uppercase\expandafter{\romannumeral1}} + \underbrace{\frac{2}{T}\sum_{t=1}^{T} \frac{\indicator(a_t=1)}{\pi_t} \inp{\beta-\widehat{\beta}_{t-1}}{X_t}\xi_t}_{\uppercase\expandafter{\romannumeral2}} \\ &\quad + \underbrace{\frac{1}{T}\sum_{t=1}^{T} \frac{\indicator(a_t=1)}{\pi_t} \xi_t^2}_{\uppercase\expandafter{\romannumeral3}}.
    \end{align*}
    For term \uppercase\expandafter{\romannumeral1}, by Assumption \ref{assump:basic} (c), $|\inp{\beta-\widehat{\beta}_{t-1}}{X_t}|$ is bounded by $\|\beta-\widehat{\beta}_{t-1}\|\phi_{\max}(s)\log^{1/2} d=O(\sqrt{\frac{s_0\log d}{(t-1)^{1-\gamma}}})\log^{1/2} d$ with high probability, then the uniform bound is
    \begin{align*}
        \big|\frac{1}{T} \frac{\indicator(a_t=1)}{\pi_t}\inp{\beta-\widehat{\beta}_{t-1}}{X_t}^2\big|&\lesssim \frac{t^\gamma}{T}\|\beta - \widehat{\beta}_{t-1}\|^2\phi_{\max}^2(s)\log d \lesssim \frac{t^\gamma}{T}\frac{s_0\log^2 d\sigma^2}{(t-1)^{1-\gamma}}\lesssim \frac{s_0\log^2 d\sigma^2}{T}, 
    \end{align*}
    and similarly,
    \begin{align*}
        \EE\left[\frac{1}{T^2} \frac{\indicator(a_t=1)}{\pi_t^2}\inp{\beta - \widehat{\beta}_{t-1}}{X_t}^4\big|\mathcal{F}_{t-1}\right]&\leq \frac{1}{T^2}\frac{2}{\varepsilon_t}\|\beta - \widehat{\beta}_{t-1}\|^4\phi_{\max}^4(s) \\
        &\lesssim \frac{t^\gamma}{T^2}\frac{s_0^2\log^4 d\sigma^4}{(t-1)^{2-2\gamma}}\lesssim \frac{t^{3\gamma - 2}s_0^2\log^4 d\sigma^4}{T^2}.
    \end{align*}
    By martingale Bernstein inequality, with probability at least $1-d_1^{-10}$,
    \begin{align*}
        \uppercase\expandafter{\romannumeral1}&\lesssim \bigg|\frac{1}{T}\sum_{t=1}^{T} \EE\left[\frac{\indicator(a_t=1)}{\pi_t}\inp{\beta - \widehat{\beta}_{t-1}}{X_t}^2\big|\mathcal{F}_{t-1}\right]\bigg| + \frac{s_0\log^3 d\sigma^2}{T} + \sqrt{\frac{s_0^2\log^5 d}{T^{3-3\gamma}}}\sigma^2 \\
        &\lesssim \bigg|\frac{1}{T}\sum_{t=1}^{T} \inp{\beta - \widehat{\beta}_{t-1}}{X_t}^2\bigg| + \frac{s_0\log^3 d\sigma^2}{T} + \sqrt{\frac{s_0^2\log^5 d}{T^{3-3\gamma}}}\sigma^2 \\
        &\lesssim \frac{s_0\log d\sigma^2}{T^{1-\gamma}} + \frac{s_0\log^3 d\sigma^2}{T} + \sqrt{\frac{s_0^2\log^5 d}{T^{3-3\gamma}}}\sigma^2,
    \end{align*}
    which indicates $\uppercase\expandafter{\romannumeral1}\overset{p}{\rightarrow} 0$ when $s_0=o(\frac{T^{1-\gamma}}{\log d})$. 
    
    Similarly, for term $\uppercase\expandafter{\romannumeral2}$, note that $|\xi_t|$ is bounded by $\sigma\log^{1/2} d$ with high probability, then for any $2\leq t\leq T$, 
    \begin{align*}
        \bigg|\frac{2}{T} \frac{\indicator(a_t=1)}{\pi_t} \inp{\beta-\widehat{\beta}_{t-1}}{X_t}\xi_t\bigg|\lesssim \frac{t^\gamma}{T}\sqrt{\frac{s_0\log^2 d\sigma^4}{(t-1)^{1-\gamma}}}\lesssim \frac{\sqrt{s_0\log^2 d}}{T}\sigma^2,
    \end{align*}
    and
    \begin{align*}
        \EE\left[\frac{4}{T^2} \frac{\indicator(a_t=1)}{\pi_t^2} \inp{\beta-\widehat{\beta}_{t-1}}{X_t}^2\xi_t^2|\mathcal{F}_{t-1}\right] \lesssim \frac{t^\gamma\sigma^2}{T^2}|\inp{\beta-\widehat{\beta}_{t-1}}{X_t}|^2 \lesssim  \frac{t^{2\gamma-1}s_0\log^2d \sigma^4}{T^2}.
    \end{align*}
    By martingale Bernstein inequality, with probability at least $1-d_1^{-10}$, 
    \begin{align*}
        \bigg|\uppercase\expandafter{\romannumeral2} - \frac{2}{T} \sum_{t=1}^{T} \EE\left[\frac{\indicator(a_t=1)}{\pi_t}\inp{\beta-\widehat{\beta}_{t-1}}{X_t} \xi_t|\mathcal{F}_{t-1}\right]\bigg|=\left|\uppercase\expandafter{\romannumeral2} \right| \lesssim \frac{\sqrt{s_0\log^4 d}}{T} + \frac{s_0^{1/2}\log^{3/2} d\sigma^2}{T^{1-\gamma}},
    \end{align*}
    which indicates $\uppercase\expandafter{\romannumeral2}\overset{p}{\rightarrow} 0$ as long as $s_0=o(\frac{T}{\log^3 d})$. 
    
    Lastly, for term $\uppercase\expandafter{\romannumeral3}$,
    \begin{align*}
        \|\frac{1}{T} \frac{\indicator(a_t=1)}{\pi_t} \xi_t^2\|_{\psi_1}\lesssim \frac{t^\gamma\sigma^2}{T},
    \end{align*}
    and
    \begin{align*}
        \EE\left[\frac{1}{T^2} \frac{\indicator(a_t=1)}{\pi_t^2} \xi_t^4|\mathcal{F}_{t-1}\right]  \lesssim  \frac{t^\gamma\sigma^4}{T^2}.
    \end{align*}
    By martingale Bernstein inequality, with probability at least $1-d_1^{-10}$, 
    \begin{align*}
        \bigg|\uppercase\expandafter{\romannumeral3} - \frac{1}{T} \sum_{t=1}^{T} \EE\left[\frac{\indicator(a_t=1)}{\pi_t} \xi_t^2|\mathcal{F}_{t-1}\right]\bigg| \lesssim \sqrt{\frac{\log d}{T^{1-\gamma}}}\sigma^2,
    \end{align*}
    which indicates $\uppercase\expandafter{\romannumeral3}\overset{p}{\rightarrow} \sigma^2$ as long as $\frac{\log d}{T^{1-\gamma}} \rightarrow 0$. \\
    {\it Step 2: showing $m_{l,T}^{\top}\widehat{\Lambda}_{T}m_{l,T}\overset{p}{\rightarrow} m_{l}^{\top}\Lambda m_l$.} \\
    Similar as the proof of Theorem \ref{thm:maininference-nodc}, we have
    \begin{align*}
        \left|m_{l,T}^{\top}\widehat{\Lambda}_T m_{l,T}-m_l^{\top}\Lambda m_l\right|&\leq \underbrace{\left|\inp{\widehat{\Lambda}_T m_{l,T}}{m_{l,T}}-\inp{\Lambda m_{l,T}}{m_{l,T}}\right|}_{(a)} \\ &\quad + \underbrace{\left|\inp{\Lambda m_{l,T}}{m_{l,T}} - \inp{\Lambda m_{l}}{m_{l}}\right|}_{(b)}.
    \end{align*}
    Notice that
    \begin{align*}
        (a)&\leq \|(\widehat{\Lambda}_T - \Lambda)m_{l,T}\|_{\max}\|m_{l,T}\|_{\ell 1} \\
         &\lesssim \|(\widehat{\Lambda}_T - \Lambda)m_{l,T}\|_{\max}\left(\|m_l\|_{\ell 1} + \frac{s_{\Omega}}{\phi_{\min}(8s_{\Omega})}\sqrt{\frac{\log d}{T}}\right).
    \end{align*}
    Next we show the upper bound of $\|(\widehat{\Lambda}_T - \Lambda)m_{l,T}\|_{\max}$. Since $\|(\widehat{\Lambda}_T - \Lambda)m_{l,T}\|_{\max}\leq \|(\widehat{\Lambda}_T - \Lambda)(m_{l,T} - m_l)\|_{\max} + \|(\widehat{\Lambda}_T - \Lambda)m_l\|_{\max}$, we show the upper bound of each term separately. 

    Note that $\|\widehat{\Lambda}_{T} -\Lambda\|_{\max}\leq \|\widehat{\Lambda}_{T} - \frac{1}{T}\sum_{k=1}^{T}\Lambda_k \|_{\max} + \|\frac{1}{T}\sum_{k=1}^{T}\Lambda_k -\Lambda\|_{\max}$, and by Lemma \ref{lambdaconvergence1}, $\|\frac{1}{T}\sum_{k=1}^{T}\Lambda_k -\Lambda\|_{\max}\lesssim \left(\frac{s_0\log d}{T^{1-\gamma}}\right)^{\nu/2}$. To prove the upper bound for $\|\widehat{\Lambda}_{T} - \frac{1}{T}\sum_{k=1}^{T}\Lambda_k \|_{\max}$, note that for any $l,j\in [d]$, $|\indicator(a_k=1)e_l^{\top}X_kX_k^{\top}e_j|\leq D^2$ and $\EE[\indicator(a_k=1)(e_l^{\top} X_kX_k^{\top}e_j)^2|\mathcal{F}_{k-1}]\leq D^4$. Then by Bernstein inequality, $\|\widehat{\Lambda}_{T} - \frac{1}{T}\sum_{k=1}^{T}\Lambda_k \|_{\max}\lesssim \sqrt{\frac{\log d}{T}}$ with probability at least $1-d^{-10}$. As a result,
    \begin{align*}
        \|(\widehat{\Lambda}_T - \Lambda)(m_{l,T} - m_l)\|_{\max}\leq \|\widehat{\Lambda}_T - \Lambda\|_{\max}\|m_{l,T} - m_l\|_{\ell 1}\lesssim \frac{s_{\Omega}}{\phi_{\min}(8s_{\Omega})}\sqrt{\frac{s_0^{\nu}\log^{1+\nu} d}{T^{1+\nu-\nu\gamma}}} + \frac{s_{\Omega}}{\phi_{\min}(8s_{\Omega})}\frac{\log d}{T}.
    \end{align*}
    Then we look at the second term. $\|(\widehat{\Lambda}_T - \Lambda)m_l\|_{\max}\leq \|(\widehat{\Lambda}_T - \frac{1}{T}\sum_{k=1}^{T}\Lambda_k)m_l\|_{\max} + \|(\frac{1}{T}\sum_{k=1}^{T}\Lambda_k - \Lambda)m_l\|_{\max}$. The latter term is smaller than $\|\frac{1}{T}\sum_{k=1}^{T}\Lambda_k - \Lambda\|_{\max}\|m_l\|_{\ell 1}\lesssim s_{\Omega}^{1/2}\left(\frac{s_0\log d}{T^{1-\gamma}}\right)^{\nu/2}$ from the above analysis. For any $j\in [d]$, $|\indicator(a_k=1)e_j^{\top}X_kX_k^{\top}m_l|\lesssim \|m_l\|_{\ell 1}$. Moreover, 
    \begin{align*}
        \EE[\indicator(a_k=1)(e_j^{\top}X_kX_k^{\top}m_l)^2|\mathcal{F}_{k-1}]\leq \EE[(e_j^{\top}X_kX_k^{\top}m_l)^2|\mathcal{F}_{k-1}] &\leq D^2 \EE[m_l^{\top}X_kX_k^{\top}m_l|\mathcal{F}_{k-1}] \lesssim \Omega_{ll}.
    \end{align*}
    Then by martingale Bernstein inequality and combine with a union bound, with probability at least $1-d^{-10}$,
    \begin{align*}
        \|(\widehat{\Lambda}_T - \frac{1}{T}\sum_{k=1}^{T}\Lambda_k)m_l\|_{\max}\lesssim \frac{\sqrt{s_{\Omega}}\log d}{T} + \sqrt{\Omega_{ll}\frac{\log d}{T}}.
    \end{align*}
    Then as long as $s_{\Omega}s_0=o(\frac{T^{\nu(1-\gamma)/2}}{\log^2 d})$ and $s_{\Omega}=o(\frac{T^{2/3}}{\log d})$, $(a)\overset{p}{\rightarrow} 0$.

    For term $(b)$, 
    \begin{align*}
        (b)&\leq \left|\inp{\Lambda m_{l,T}}{m_{l,T} - m_l}\right| + \left|\inp{\Lambda (m_{l,T}-m_l)}{m_l}\right| \\
        &\leq \|\Lambda\|\|m_{l,T}\|\|m_{l,T} - m_l\|+ \|\Lambda\|\|m_l\|\|m_{l,T} - m_l\| \\
        &\lesssim \|\Lambda\|\|m_{l,T} - m_l\|\left(\|m_l\| + \frac{\sqrt{s_{\Omega}}}{\phi_{\min}(8s_{\Omega})}\sqrt{\frac{\log d}{T}}\right) \\
        &\lesssim \|\Lambda\|\frac{\sqrt{s_{\Omega}}}{\phi_{\min}(8s_{\Omega})}\sqrt{\frac{\log d}{T}}\left(\|m_l\| + \frac{\sqrt{s_{\Omega}}}{\phi_{\min}(8s_{\Omega})}\sqrt{\frac{\log d}{T}}\right).
    \end{align*}
    By Assumption \ref{assump:basic}, $\|m_l\|\leq C_{\min}$ is bounded. As long as $s_{\Omega}=o(\frac{T}{\log d})$, $(b)\overset{p}{\rightarrow} 0$. 
\end{proof}

\subsection{Proof of Corollary \ref{cor:difference-nodc}}
\begin{proof}
    For any $l\in [d]$, we have
    \begin{align*}
        (\widehat{\beta}_{1(l)}^{\uipw} - \widehat{\beta}_{0(l)}^{\uipw}) - (\beta_{1(l)} - \beta_{1(l)})&= \underbrace{\frac{1}{T}e_l^{\top}\sum_{t=1}^{T} \indicator(a_t=1)\Omega X_t\xi_t}_{V_{1T}} + \underbrace{\frac{1}{T}e_l^{\top}\sum_{t=1}^{T} \indicator(a_t=1)(M_{T} - \Omega)X_t\xi_t}_{R_{11}} \\ &\quad \underbrace{\frac{1}{T}e_l^{\top}\sum_{t=1}^{T} \left(I-\indicator(a_t=1)M_T^{(1)}X_tX_t^{\top}\right)(\widehat{\beta}_{1,T} - \beta_1)}_{R_{12}} \\ &\quad - \underbrace{\frac{1}{T}e_l^{\top}\sum_{t=1}^{T} \indicator(a_t=0)\Omega X_t\xi_t}_{V_{0T}} + \underbrace{\frac{1}{T}e_l^{\top}\sum_{t=1}^{T} \indicator(a_t=0)(M_{T} - \Omega)X_t\xi_t}_{R_{01}} \\ &\quad \underbrace{\frac{1}{T}e_l^{\top}\sum_{t=1}^{T} \left(I-\indicator(a_t=0)M_TX_tX_t^{\top}\right)(\widehat{\beta}_{0,T} - \beta_0)}_{R_{02}}.
    \end{align*}
    In the proof of Theorem \ref{thm:maininference-nodc}, we have shown that $R_{11}$, $R_{12}$, $R_{01}$ and $R_{02}$ are negligible compared to the variance of $V_{1T}$ and $V_{0T}$. We can rewrite $V_{1T}-V_{0T}$ as
    \begin{align*}
        \underbrace{\frac{1}{T}\sum_{t_1=1}^{T} \indicator(a_{t_1}=1)m_{l}^{\top}X_{t_1}\xi_{t_1}}_{V_{1T}} - \underbrace{\frac{1}{T}\sum_{t_2=1}^{T} \indicator(a_{t_2}=0)m_{l}^{\top}X_{t_2}\xi_{t_2}}_{V_{0T}},
    \end{align*}
    where $m_l=\Omega e_l$ for $i=0,1$. As long as we show $V_{1T}$ is uncorrelated with $V_{0T}$, then the asymptotic variance of $\sqrt{T}(V_{1T}-V_{0T})$is the sum of their individual variance, i.e., $\sigma_1^2S_1^2+\sigma_0^2S_0^2$ with $S_1$ and $S_0$ defined in Theorem \ref{thm:maininference-nodc}. Notice that,
    \begin{align*}
        &\quad \frac{1}{T^2}\sum_{t_1=1}^{T} \indicator(a_{t_1}=1)m_{l}^{\top}X_{t_1}\xi_{t_1} \sum_{t_2=1}^{T} \indicator(a_{t_2}=0)m_{l}^{\top}X_{t_2}\xi_{t_2} \\
        &=\frac{1}{T^2}\sum_{t_1=1}^{T}\sum_{t_2=1}^{T} \indicator(a_{t_1}=1)\indicator(a_{t_2}=0) m_{l}^{\top}X_{t_1}m_{l}^{\top}X_{t_2}\xi_{t_1}\xi_{t_2} .
    \end{align*}
    When $t_1=t_2$, $\indicator(a_{t_1}=1)\indicator(a_{t_2}=0)=0$. When $t_1\neq t_2$, $\EE[\indicator(a_{t_1}=1)\indicator(a_{t_2}=0) m_{l}^{\top}X_{t_1}m_{l}^{\top}X_{t_2}\xi_{t_1}\xi_{t_2}]=0$ due to the i.i.d. distributed $\xi_t$. As a result, the two terms are uncorrelated. \\
    Then similar as Theorem \ref{thm:stuinference-nodc}, we can replace $\sigma_1^2S_1^2+\sigma_0^2S_0^2$ with $\widehat{\sigma}_1^2\widehat{S}_1^2+\widehat{\sigma}_0^2\widehat{S}_0^2$ and conclude the proof.
\end{proof}

\subsection{Proof of Theorem \ref{thm:estimation-dc}}
\begin{proof}
    The proof shares a similar fashion with the proof of Theorem \ref{thm:estimation-nodc}. We first define the oracle gradient at time $t$ as $\nabla f^*(\widehat\beta_{i, t})=\frac{2}{t}\sum_{\tau=1}^{t} \EE[\indicator(a^*(X_{\tau})=i)X_{\tau}X_{\tau}^{\top}](\widehat{\beta}_{i,t-1}-\beta_i)=2\Lambda_i(\widehat{\beta}_{i,t-1}-\beta_i)$ for $i=0,1$ where $a^*(X_{\tau})= \indicator(\inp{\beta_1-\beta_0}{X_{\tau}}>0)$ is the oracle optimal arm. This is different from the population gradient $\nabla f(\widehat\beta_{i,t})=\frac{2}{t}\sum_{\tau=1}^{t} \EE[\indicator(a_{\tau}=i)X_{\tau}X_{\tau}^{\top}|\mathcal{F}_{t-1}](\widehat{\beta}_{i,t-1}-\beta_i)=\frac{2}{t}\sum_{\tau=1}^t \Lambda_{i,t}(\widehat{\beta}_{i,t-1}-\beta_i)$.

%\commentWT{check page 27 of \cite{ma2024high}, only the term $\inp{\nabla f_t(\widehat{\beta}_{i,t-1})}{\widehat{\beta}_{i,t-1} - \beta_i}$ needs to be replaced. For the induction process, suppose we have
%\begin{equation*}
%    \begin{aligned}
%        x_{t+1}\le (1-\frac{1}{\kappa^4} +c_2\frac{\sum_{i=1}^t\sqrt{x_i}}{t})x_t +\frac{C_0}{t+1},
%    \end{aligned}
%\end{equation*}
%we can assume a  bound $x_t\le \frac{C_1}{t}$ for $t\ge T_0$, but we only have a trivial bound, e.g., $x_t\le1$ for the burn-in phase $t\le T_0$. Then, the induction needs the guarantee:
%\begin{equation*}
%        \begin{aligned}
%         &(1-\frac{1}{\kappa^4} + C c_2\sqrt{\frac{C_1}{t}} )\frac{C_1}{t} +\frac{C_0}{t+1}\le \frac{C_1}{t+1} 
%         \\
%         &  (1-\frac{1}{\kappa^4} + C c_2\sqrt{\frac{C_1}{t}} ) (1+\frac{1}{t})\le 1-\frac{C_0}{C_1}. 
%    \end{aligned}
%\end{equation*}
%Only when $t\ge C c_2^2C_1 \kappa^8$, we have
%\begin{equation*}
%    \begin{gathered}
%        (1-\frac{1}{2\kappa^4}) \le 1 -\frac{C_0}{C_1} \\
%        2 C_0\kappa^4 \le C_1
%    \end{gathered}
%\end{equation*}
%However, the burn-in phase $T_0$ is related to $C_1$, making the choice of any $C_1$ impossible. For example, suppose we choose $C_1 = 2 C_0\kappa^4 $ and $T_0=C c_2^2C_1 \kappa^8=Cc_2^2C_0\kappa^{12}$, then we have
%\begin{equation*}
%    x_{T_0}\le \frac{C_1}{T_0}\le \frac{1}{C c_2^2\kappa^8},
%\end{equation*}
%which is far smaller than the trivial bound $x_{T_0}\le 1$ before the induction begins.
%}

    An argument analog to the proof of Theorem \ref{thm:estimation-nodc} gives that
     \begin{align*}
        &\quad \max\left\{\|\widehat{\beta}_{1,t} - \beta_1\|^2, \|\widehat{\beta}_{0,t} - \beta_0\|^2\right\} \\
        &\leq \max_{i\in\{0,1\}} \left(1+\frac{3}{2}\sqrt{\rho} \right)\left(\|\widehat{\beta}_{i,t-1} - \beta_i\|^2 -2\eta\inp{B(g_{i,t})}{\widehat{\beta}_{i,t-1} - \beta}+ \eta^2\|B(g_{i,t})\|^2\right) \\
        &\leq \max_{i\in\{0,1\}} \left(1+\frac{3}{2}\sqrt{\rho} \right) \left(\|\widehat{\beta}_{i,t-1} - \beta_i\|^2 - 2\eta\inp{\nabla f^*(\widehat{\beta}_{i,t-1})}{\widehat{\beta}_{i,t-1} - \beta_i} + 2\eta^2\|B(g_{i,t}-\nabla f^*(\widehat{\beta}_{i,t-1}))\|^2 \right.\\ &\left.\quad + 2\eta^2\|B(\nabla f^*(\widehat{\beta}_{i,t-1}))\|^2 + 2\eta\|B(g_{i,t}-\nabla f^*(\widehat{\beta}_{i,t-1}))\|\|\widehat{\beta}_{i,t-1} - \beta_i\| \right),
    \end{align*}
    where we use the fact that $\inp{\nabla f(\widehat{\beta}_{i,t-1})}{\widehat{\beta}_{i,t-1} - \beta_i}=\inp{B(\nabla f(\widehat{\beta}_{i,t-1}))}{\widehat{\beta}_{i,t-1} - \beta_i}$ by the definition of $B(\cdot)$.

    For any $v$ such that $\|v\|_0\leq 2s$, we have $v^{\top}\Lambda_i v\leq v^{\top}\Sigma v\leq \phi_{\max}(s)\|v\|^2$, then it is clear that the $2s$-sparse maximal eigenvalue of $\Lambda_i$ is bounded by $\phi_{\max}(s)$. As a result, $\|B(\nabla f(\widehat{\beta}_{i,t-1}))\|\leq 2\phi_{\max}(s)\|\widehat{\beta}_{i,t-1} - \beta_i\|$. Also by Assumption \ref{covariate diversity}, $\inp{B(g_{i,t})}{\widehat{\beta}_{i,t-1} - \beta_i}\geq \lambda_0\|\widehat{\beta}_{i,t-1} - \beta_i\|^2$. 

    This directly implies
    \begin{equation}
        \begin{aligned}
        &\max\left\{\|\widehat{\beta}_{1,t} - \beta_1\|^2, \|\widehat{\beta}_{0,t} - \beta_0\|^2\right\} \leq \max_{i\in\{0,1\}}\left(1+\frac{3}{2}\sqrt{\rho} \right)\left(1-2\lambda_0\eta+8\eta^2\phi_{\max}^2(s)\right)\|\widehat{\beta}_{i,t-1} - \beta_i\|^2 \\
        &\quad + 18s\eta^2\|g_t-\nabla f^*(\widehat{\beta}_{i,t-1})\|_{\max}^2 + 18\eta\sqrt{s}\|g_{i,t}-\nabla f^*(\widehat{\beta}_{i,t-1})\|_{\max}\|\widehat{\beta}_{i,t-1} - \beta_i\| \\
        &\leq \max_{i\in\{0,1\}}\left(1+\frac{3}{2}\sqrt{\rho} \right)\left(1-2\lambda_0\eta+8\eta^2\phi_{\max}^2(s)\right)\|\widehat{\beta}_{i,t-1} - \beta_i\|^2 \\
        &\quad + 18s\eta^2\|g_{i,t}-\nabla f(\widehat{\beta}_{i,t-1})\|_{\max}^2 + 18\eta\sqrt{s}\|g_{i,t}-\nabla f(\widehat{\beta}_{i,t-1})\|_{\max}\|\widehat{\beta}_{i,t-1} - \beta_i\| \\
        &\quad + 18s\eta^2\|\nabla f(\widehat{\beta}_{i,t-1})-\nabla f^*(\widehat{\beta}_{i,t-1})\|_{\max}^2 + 18\eta\sqrt{s}\|\nabla f(\widehat{\beta}_{i,t-1})-\nabla f^*(\widehat{\beta}_{i,t-1})\|_{\max}\|\widehat{\beta}_{i,t-1} - \beta_i\|.
        \label{eq:recursive}
        \end{aligned}
    \end{equation}
    Lemma 7 in \cite{ma2024high} has already shown the upper bound of $\|g_{i,t}-\nabla f(\widehat{\beta}_{i,t-1})\|_{\max}$ for both $i=0,1$ as 
    \begin{align*}
        \|g_{i,t}-\nabla f(\widehat{\beta}_{i,t-1})\|_{\max}^2\lesssim sD^2\frac{\log d}{t}\|\widehat{\beta}_{i,t-1} - \beta_i\|^2 + \frac{\sigma_i^2D^2\log d}{t}
    \end{align*}
    with probability at least $1-d^{-10}$. Then we only need to prove the upper bound for $\|\nabla f(\widehat{\beta}_{i,t-1})-\nabla f^*(\widehat{\beta}_{i,t-1})\|_{\max}$. It can be decomposed into 
    \begin{align*}
        \|\nabla f(\widehat{\beta}_{i,t-1})-\nabla f^*(\widehat{\beta}_{i,t-1})\|_{\max}&\leq \frac{2}{t}\sum_{\tau=1}^t\left\|(\Lambda_{i,\tau}-\Lambda_i)(\widehat{\beta}_{i,t-1}-\beta_i)\right\|_{\max} \\
        &\leq \frac{4}{t}\sqrt{s}\sum_{\tau=1}^t \left\|\Lambda_{i,\tau}-\Lambda_i\right\|_{\max}\|\widehat{\beta}_{i,t-1}-\beta_i\|.
    \end{align*}
   From the proof of Lemma \ref{lambdaconvergence1}, we have that
   \begin{align*}
       \left\|\Lambda_{i,\tau}-\Lambda_i\right\|_{\max}&\leq 4D^2\PP\left(a_\tau\neq a^*(X_\tau)|\mathcal{F}_{\tau-1}\right)=4D^2\PP\left(|\Delta_{X_\tau}|\leq |\Delta_{X_\tau} - \widehat{\Delta}_{X_\tau}||\mathcal{F}_{t-1}\right) \\
       &\leq 4D^2C_0\left(\max_{X\in \mathcal{X}} |\inp{\widehat{\beta}_{1,\tau-1}-\beta_1}{X}| + \max_{X\in \mathcal{X}} |\inp{\widehat{\beta}_{0,\tau-1}-\beta_0}{X}|\right)^{\nu} \\
       &\lesssim D^2C_0\left(\max_i \|\widehat{\beta}_{i,\tau-1}-\beta_i\|\right)^{\nu}\phi_{\max}(s)\log^{1/2}d,
   \end{align*}
   where the second last inequality comes from Assumption \ref{assump:margincondition}. Overall, we have
   \begin{align*}
       \max_i\|\nabla f(\widehat{\beta}_{i,t-1})-\nabla f^*(\widehat{\beta}_{i,t-1})\|_{\max}^2\lesssim sD^4C_0^2\left(\sum_{\tau=1}^{t} \max_i \|\widehat{\beta}_{i,\tau-1}-\beta_i\|^{\nu} \right)^2\phi_{\max}^2(s)\frac{\log d}{t^2}\max_i\|\widehat{\beta}_{i,t-1}-\beta_i\|^2.
   \end{align*}
   
   Apply all the above to the recursive relationship in (\ref{eq:recursive}) leads to 
    \begin{align*}
        &\max\left\{\|\widehat{\beta}_{1,t} - \beta_1\|^2, \|\widehat{\beta}_{0,t} - \beta_0\|^2\right\}\lesssim \max_{i\in\{0,1\}}\left(1+\frac{3}{2}\sqrt{\rho} \right)\left(1-2\lambda_0\eta+8\eta^2\phi_{\max}^2(s)\right)\|\widehat{\beta}_{i,t-1} - \beta_i\|^2 \\
        &\quad + \eta s\left(D\sqrt{\frac{\log d}{t}}+ \frac{D^2C_0\phi_{\max}(s)\log^{1/2}d}{t}\sum_{\tau=1}^{t} \max_i \|\widehat{\beta}_{i,\tau-1}-\beta_i\|\right)\|\widehat{\beta}_{i,t-1} - \beta_i\|^2\\
        &\quad + \eta^2(\sigma_0^2\vee \sigma_1^2)\frac{sD^2\log d}{t} + \eta\sqrt{s}(\sigma_0\vee \sigma_1) D\sqrt{\frac{\log d}{t}}\|\widehat{\beta}_{i,t-1} - \beta_i\| \\
        &\quad + \eta^2s^2D^4C_0^2\left(\sum_{\tau=1}^{t} \max_i \|\widehat{\beta}_{i,\tau-1}-\beta_i\|^{\nu} \right)^2\phi_{\max}^2(s)\frac{\log d}{t^2}\max_i\|\widehat{\beta}_{i,t-1}-\beta_i\|^2.
    \end{align*}
    We set $\rho=\frac{\lambda_0^4}{150\phi_{\max}^4(s)}$ and $\eta=\frac{\lambda_0}{8\phi_{\max}^2(s)}$. Next we show the upper bound of $\max\{\|\widehat{\beta}_{1,t} - \beta_1\|^2, \|\widehat{\beta}_{0,t} - \beta_0\|^2\}$ by induction. After the burn in period with length $T_0=C_1s_0\log d$ for some large constant $C_1$, the estimation error becomes $\max\left\{\|\widehat{\beta}_{1,t} - \beta_1\|^2, \|\widehat{\beta}_{0,t} - \beta_0\|^2\right\}\leq \frac{1}{C_2}$ for some large constant $C_2$. Suppose $\max\left\{\|\widehat{\beta}_{1,\tau} - \beta_1\|^2, \|\widehat{\beta}_{0,\tau} - \beta_0\|^2\right\}\leq C_0^2C_3(\sigma_0^2\vee \sigma_1^2)\frac{s_0\log d}{\phi_{\max}^2(s)\tau}$ at $T_0<\tau\leq t-1$, we have
    \begin{align*}
        &\max\left\{\|\widehat{\beta}_{1,t} - \beta_1\|^2, \|\widehat{\beta}_{0,t} - \beta_0\|^2\right\}\leq \left(1+\frac{3}{2}\sqrt{\rho} \right)\left(1-2\lambda_0\eta+8\eta^2\phi_{\max}^2(s)\right)\max_{i}\|\widehat{\beta}_{i,t-1} - \beta_i\|^2 \\
        &\quad + \eta s_0\left(D\sqrt{\frac{\log d}{t}}+ \frac{D^2C_0\phi_{\max}(s)\log^{1/2}d}{t}(C_1s_0^{3/2}\log^{3/2} d \right.\\ &\left.\quad + \sum_{\tau=T_0+1}^t C_0\sqrt{C_3}\sqrt{(\sigma_0^2\vee \sigma_1^2)\frac{s_0\log d}{\phi_{\max}^2(s)(\tau-1)}})\right)C_0^2C_3(\sigma_0^2\vee \sigma_1^2)\frac{s_0\log d}{\phi_{\max}^2(s)(t-1)}\\
        &\quad + C_4\frac{\sigma_0^2\vee \sigma_1^2}{\phi_{\max}^2(s)}\frac{sD^2\log d}{t} + C_5\frac{\sqrt{s}(\sigma_0\vee \sigma_1)}{\phi_{\max}(s)}\sqrt{\frac{\log d}{t}}\max_{i}\|\hat\beta_{i,t-1}-\beta_i\| \\
        &\quad + C_6\frac{s_0^2C_0^2\log d}{t^2}\left(C_1s_0^{3/2}\log^{3/2} d+ \sum_{\tau=T_0+1}^t C_0\sqrt{C_3}\left((\sigma_0^2\vee \sigma_1^2)\frac{s_0\log d}{\phi_{\max}^2(s)(\tau-1)}\right)^{\nu/2}\right)^2\max_i \|\hat\beta_{i,t-1}-\beta_i\|^2.
    \end{align*}
   When $t\geq T_0\gtrsim C_0^2s^{(2+\nu)/\nu}\log^{(2+\nu)/\nu} d$, by induction, we conclude that
   \begin{align*}
        \max\left\{\|\widehat{\beta}_{1,t} - \beta_1\|^2, \|\widehat{\beta}_{0,t} - \beta_0\|^2\right\}\lesssim \frac{C_0^2(\sigma_0^2\vee \sigma_1^2)D^2s_0}{\phi_{\max}^2(s)}\frac{\log dT}{t},
   \end{align*}
   holding with probability at least $1-d^{-10}$. 
\end{proof}

\subsection{Proof of Theorem \ref{thm:regret-dc}}
\begin{proof}
      Without loss of generality, assume at time $t$, arm 1 is optimal, i.e., $\inp{\beta_1-\beta_0}{X_t}>0$. Under the exploration free algorithm, the expected regret is
\begin{align*}
    r_t&=\EE\left[\inp{\beta_1-\beta_0}{X_t}\indicator(\text{choose arm 0}) \right] \\
    &= \EE\left[\inp{\beta_1-\beta_0}{X_t}\indicator(\text{choose arm 0}, \inp{\widehat{\beta}_{0,t-1}-\widehat{\beta}_{1,t-1}}{X_t}> 0)\right].
\end{align*}
By Assumption \ref{assump:basic}(c) and Theorem \ref{thm:estimation-dc}, when $t\geq T_0$, with probability at least $1-d^{-100}$, $\max_{X\in \mathcal{X}}\|\inp{\beta_1-\widehat{\beta}_{1,t-1}}{X}\|_{\max}$ is bounded by $C_1\frac{\sigma_1\sqrt{s_0}\log d}{\lambda_0\sqrt{t-1}}$ for some constant $C_1$, and same for $\max_{X\in \mathcal{X}}\|\inp{\beta_0-\widehat{\beta}_{0,t-1}}{X}\|_{\max}$, while $t<T_0$ it has the trivial constant estimation error. Denote $\delta_t=C_1\frac{\sigma_1\sqrt{s_0}\log d}{\lambda_0\sqrt{t-1}}$, and event $B_t=\{\inp{\beta_1-\beta_0}{X_t}>2\delta_t\}$, then the above expectation can be written as
\begin{align*}
    &\quad\EE\left[\inp{\beta_1-\beta_0}{X_t}\indicator\left(\inp{\widehat{\beta}_{0,t-1}-\widehat{\beta}_{1,t-1}}{X_t}> 0\cap B_t\right)\right] \\ &\quad\quad + \EE\left[\inp{\beta_1-\beta_0}{X_t}\indicator\left(\inp{\widehat{\beta}_{0,t-1}-\widehat{\beta}_{1,t-1}}{X_t}> 0\cap B_t^c\right)\right] \\
    &\leq R_{\max} \EE\left[\indicator\left(\inp{\widehat{\beta}_{0,t-1}-\widehat{\beta}_{1,t-1}}{X_t}> 0\cap B_t\right)\right] + 2\delta_t\EE\left[\indicator(B_t^c)\right].
\end{align*}
Under event $B_t$ and $\inp{\widehat{\beta}_{0,t-1}-\widehat{\beta}_{1,t-1}}{X_t}> 0$, 
\begin{align*}
    0>\inp{\widehat{\beta}_{1,t-1}-\widehat{\beta}_{0,t-1}}{X_t} &= \inp{\widehat{\beta}_{1,t-1}-\beta_1}{X_t} + \inp{\beta_0-\widehat{\beta}_{0,t-1}}{X_t} + \inp{\beta_1-\beta_0}{X_t}\\
    &> \inp{\widehat{\beta}_{1,t-1}-\beta_1}{X_t} + \inp{\beta_0-\widehat{\beta}_{0,t-1}}{X_t} + 2\delta_t.
\end{align*}
This means either $\inp{\widehat{\beta}_{1,t-1}-\beta_1}{X_t}<-\delta_t$ or $\inp{\beta_0-\widehat{\beta}_{0,t-1}}{X_t}<-\delta_t$, which further implies either $\max |\inp{\beta_0-\widehat{\beta}_{0,t-1}}{X}|$ or $\max |\inp{\beta_0-\widehat{\beta}_{0,t-1}}{X}|$ should be larger than $\delta_t$. Therefore, 
\begin{align*}
    \EE\left[\indicator(\inp{\widehat{\beta}_{0,t-1}-\widehat{\beta}_{1,t-1}}{X_t}> 0)\cap B_t\right]&= \PP\left(\inp{\widehat{\beta}_{0,t-1}-\widehat{\beta}_{1,t-1}}{X_t}> 0 \cap B_t\right) \\
    &\leq \PP\left(\max |\inp{\beta_1-\widehat{\beta}_{1,t-1}}{X}|>\delta_t\right) + \PP\left(\max |\inp{\beta_0-\widehat{\beta}_{0,t-1}}{X}|>\delta_t\right) \\
    &\leq \frac{2}{d^{100}},
\end{align*}
For the other term, by Assumption \ref{assump:margincondition}, $\EE\left[\indicator(B_t^c)\right]=\PP(\inp{\beta_1-\beta_0}{X_t}<2\delta_t)\leq 2C_0\delta_t^{\nu}$. Then
\begin{align*}
    r_t&\lesssim \frac{2}{d^{100}}R_{\max}  + 4C_0\delta_t^{1+\nu}. 
\end{align*}
The total regret up to time $T$ is
\begin{align*}
    R_T&=\sum_{t=1}^{T} r_t \lesssim \frac{T}{d^{100}}R_{\max} + \sum_{t=1}^{T} \delta_t^{1+\nu}.
\end{align*}
As a result, when $\nu=1$, $R_T\lesssim \lambda_0^{-2}s_0\log T\log^2 d$; when $0<\nu<1$, $R_T\lesssim \lambda_0^{-(1+\nu)}s_0^{(1+\nu)/2}T^{(1-\nu)/2}\log^2 d$.
\end{proof}

\subsection{Proof of Theorem \ref{thm:maininference-dc}}
\begin{proof}
    Without loss of generality, we only prove $i=1$ case and omit all the subscript. $i=0$ can be proved in the same way. The proof almost follows the same arguments with Theorem \ref{thm:maininference-nodc}. Recall that for any $l\in [d]$,
    \begin{align*}
        \sqrt{T}(\widehat{\beta}_{(l)}^{\uaw} - \beta_{(l)})&= \underbrace{\frac{1}{\sqrt{T}}e_l^{\top}\sum_{t=1}^{T} \indicator(a_t=1)\Lambda^{-1} X_t\xi_t}_{V_T} + \underbrace{\frac{1}{\sqrt{T}}e_l^{\top}\sum_{t=1}^{T} \indicator(a_t=1)(M_{T} - \Lambda^{-1})X_t\xi_t}_{R_1} \\ &\quad \underbrace{\frac{1}{\sqrt{T}}e_l^{\top}\sum_{t=1}^{T} \left(I_d-\indicator(a_t=1)M_TX_tX_t^{\top}\right)(\widehat{\beta}_{T} - \beta)}_{R_2}.
    \end{align*}
    The next Lemma shows that $R_1+R_2$ are negligible bias terms,
    \begin{Lemma}
        \label{lemma:bias2}
        Under the conditions in Theorem \ref{thm:maininference-dc},
        \begin{align*}
            \|R_1+R_2\|_{\max}=o_p(1).
        \end{align*}
    \end{Lemma}
    Next, we again apply Theorem 3.2 and Corollary 3.1 in \cite{hall2014martingale}, the Martingale Central Limit Theorem to show the asymptotic normality of $V_T$. \\
    \emph{Step 1: checking Lindeberg Condition} \\
    For simplicity, we denote $m_l=\Lambda^{-1}e_l$. For any $\delta>0$,
    \begin{align*}
        &\frac{1}{T}\sum_{t=1}^{T} \EE\left[(e_{l}^{\top}\indicator(a_t=1)\Lambda^{-1} X_t\xi_t)^2 \indicator\left(\bigg|\frac{1}{\sqrt{T}}e_{l}^{\top}\indicator(a_t=1)\Lambda^{-1} X_t\xi_t \bigg|>\delta\right)\bigg| \mathcal{F}_{t-1}\right] \\
        &\leq \frac{1}{T}\sum_{t=1}^{T}\EE\left[\indicator(a_t=1) m_l^{\top} X_tX_t^{\top}m_l\xi_t^2 \indicator\left(\big|m_l^{\top} X_t\xi_t \big|>\delta\sqrt{T} \right)\bigg| \mathcal{F}_{t-1}\right] \\
        &\leq \frac{\sigma^2}{T}\sum_{t=1}^{T} \max_{X\in \mathcal{X}} \|m_l^{\top} X\|^2 \times \EE\left[\indicator\left(\big|m_l^{\top} X_t\xi_t \big|>\delta\sqrt{T} \right)\right],
    \end{align*}
    Since $\xi_t$ is a subGaussian random variable, denote event $\mathcal{E}_{1,t}=\{|\xi_t|\leq c_1\sigma\log^{1/2}d\}$ for some constant $c_1$ with $\PP(\mathcal{E}_{1,t})\geq 1 - d^{-20}$. Also notice that $|m_l^{\top}X_t|\leq \|m_l\|_{\ell 1}\|X_t\|_{\max}\leq \sqrt{s_{\Lambda^{-1}}}\|m_l\|\|X_t\|_{\max}$ and $\|m_l\|\leq \lambda_0$ by Assumption \ref{covariate diversity}. Then as long as $s_{\Lambda^{-1}}\leq \frac{c_2T}{\log d}$ for some small enough $c_2$, we have $|m_l^{\top}X_t|\leq \frac{\delta}{c_1}\sqrt{\frac{T}{\log d\sigma^2}}$ for any $t$.  

    As a result,
    \begin{align*}
        \PP\left(\big|m_l^{\top}X_t\xi_t \big|>\delta\sqrt{T}\right)&\leq \PP\left(|\xi_t|\geq c_1\sigma\log^{1/2}d\right) = \PP(\mathcal{E}_{1,t}^c) \leq 2d^{-20}.
    \end{align*}
    Therefore,
    \begin{align*}
        &\frac{\sigma^2}{T}\sum_{t=1}^{T} \max_{X\in \mathcal{X}} \|m_l^{\top}X\|^2 \times \EE\left[\indicator\left(\big|m_l^{\top}X_t\xi_t \big|>\delta\sqrt{T} \right)\right]\\ &\quad \lesssim \max_{X\in \mathcal{X}} \|m_l^{\top}X\|^2 d^{-20}\rightarrow 0,
    \end{align*}
    and the Lindeberg condition is satisfied. \\
    \emph{Step 2: calculating the variance}
    \begin{align*}
        &\frac{1}{T}\sum_{t=1}^{T} \EE\left[(e_{l}^{\top}\indicator(a_t=1)\Lambda^{-1}X_t\xi_t)^2|\mathcal{F}_{t-1}\right] \\
        &\quad = \frac{\sigma^2}{T}\sum_{t=1}^{T} m_{l}^{\top}\EE\left[\indicator(a_t=1)X_tX_t^{\top}|\mathcal{F}_{t-1}\right] m_{l}.
    \end{align*}
    Notice that $\EE\left[\indicator(a_t=1)X_tX_t^{\top}|\mathcal{F}_{t-1}\right]=\EE\left[X_tX_t^{\top}\indicator(\inp{\widehat{\beta}_{1,t-1} - \widehat{\beta}_{0,t-1}}{X_t}>0)|\mathcal{F}_{t-1}\right]:=\Lambda_{1,t}$. Similar to Lemma \ref{lambdaconvergence1}, we prove a upper bound of $m_{l}^{\top}\Lambda_t m_{l}-m_l^{\top}\Lambda m_l$ and $ \|\Lambda_{i,t} - \Lambda_i\|_{\max}$.
    \begin{Lemma}
        \label{lambdaconvergence2}
        Under the conditions in Theorem \ref{thm:maininference-dc}, when $t\geq T_0$, for $i=1,0$, with probability at least $1-d^{-10}$, we have
        \begin{align*}
            \|\Lambda_{i,t} - \Lambda_i\|_{\max}\lesssim\left(\frac{s_0\log d}{t-1}\right)^{\nu/2} \quad \text{and} \quad |m_l^{\top}(\Lambda_{i,t} - \Lambda_i)m_l|\lesssim \phi_{\max}^2(s_{\Lambda^{-1}})\left(\frac{s_0\log^3 d}{t-1}\right)^{\nu/2}.
        \end{align*}
    \end{Lemma}
    Therefore,
    \begin{align*}
        &\quad \left| \frac{1}{T}\sum_{t=1}^{T} m_{l}^{\top}\EE\left[\indicator(a_t=1)X_tX_t^{\top}|\mathcal{F}_{t-1}\right] m_{l} - \Lambda_{ll}^{-1}\right|  \\
        &\lesssim \frac{1}{T} \sum_{t=T_0}^{T} \left(\frac{s_0\log d}{t-1}\right)^{\nu/2}\lesssim \left(\frac{s_0\log d}{T}\right)^{\nu/2}. 
    \end{align*}
    The above equation indicates $\frac{1}{T}\sum_{t=1}^{T} m_{l}^{\top}\Lambda_t m_{l} \overset{p}{\rightarrow} \Lambda_{ll}^{-1}$ when $T,d_1,d_2\rightarrow \infty$ as long as $s_0=o(\frac{T}{\log d})$. Then we conclude the proof.
\end{proof}

\subsection{Proof of Theorem \ref{thm:stuinference-dc}}
\begin{proof}
    The proof follows almost the same arguments as Theorem \ref{thm:stuinference-nodc}. Denote $m_l^{(i)}=\Lambda_i^{-1} e_l$. We need to prove for any $l\in [d]$, $i=0,1$, $\widehat{\sigma}_i^2$ and $m_{l,T}^{(i)\top}\widehat{\Lambda}_{i,T}m_{l,T}^{(i)}$ are consistent estimators for $\sigma_i^2$ and $\Lambda_{i(ll)}^{-1}=m_{l}^{(i)\top}\Lambda_i m_l^{(i)}$, respectively. Without loss of generality, we only prove $i=1$ case and omit all the subscript and  superscript $i$. \\
    {\it Step 1: showing $\widehat{\sigma}^2\overset{p}{\rightarrow} \sigma^2$.} \\
    Denote $T^{'}=\sum_{t=1}^{T}\indicator(a_t=1)$. It is important to note that, by martingale Weak Law of Large Numbers (Theorem 2.19 in \cite{hall2014martingale}), $T^{'}\overset{p}{\rightarrow}\sum_{t=1}^{T} \EE\left[\indicator(a_t=1)|\mathcal{F}_{t-1}\right]$. We first show that, under Assumption \ref{covariate diversity}, $T'$ is of same order as $T$.     

   Notice that, when $t\geq T_0$,
   \begin{align*}
       \EE\left[\indicator(a_t=1)|\mathcal{F}_{t-1}\right]&=\PP\left(\inp{\widehat{\beta}_{1,t-1}-\widehat{\beta}_{0,t-1}}{X}>0|\mathcal{F}_{t-1}\right) \\
       &\geq \PP\left(\inp{\beta_1-\beta_0}{X}>0\right) - \PP\left((\inp{\beta_1-\beta_0}{X}>0)\cap(\inp{\widehat{\beta}_{1,t-1}-\widehat{\beta}_{0,t-1}}{X}\leq 0)\right) \\
       &\geq \PP\left(\inp{\beta_1-\beta_0}{X}>0\right) - c_1(\frac{s_0\log d}{t-1})^{\nu/2}
   \end{align*}
    for some constant $c_1$, where the last inequality comes from proof of Lemma \ref{lambdaconvergence2} and Lemma \ref{probright}. Moreover, for any unit vector $v$ with sparsity $s_0$, 
    \begin{align*}
        v^{\top}\EE\big[\indicator(\inp{\beta_1-\beta_0}{X}>0)XX'\big]v=\EE\big[\indicator(\inp{\beta_1-\beta_0}{X}>0)(v^{\top}X)^2 \big]\geq \lambda_0.
    \end{align*}
    By Assumption \ref{assump:basic}(c), $|v^{\top}X|\lesssim \phi_{\max}(s_0)\log^{1/2}d$ with high probability. As a result,
    \begin{align*}
        \EE\big[\indicator(\inp{\beta_1-\beta_0}{X}>0)(v^{\top}X)^2 \big]\lesssim \PP\big(\inp{\beta_1-\beta_0}{X}>0\big)\phi_{\max}(s_0)\log^{1/2}d,
    \end{align*}
    which directly implies $\PP\big(\inp{\beta_1-\beta_0}{X}>0\big)\geq C_1\lambda_0/\phi_{\max}(s_0)\log^{1/2}d$ for some constant $C_1$. Then $\sum_{t=1}^{T} \EE\left[\indicator(a_t=1)|\mathcal{F}_{t-1}\right]\geq 1/2C_1\lambda_0/(\phi_{\max}(s_0)\log^{1/2}d)\cdot T$. 
    
    Notice that $\widehat{\sigma}^2$ can be written as
    \begin{align*}
        \widehat{\sigma}^2&= \underbrace{\frac{1}{T^{'}}\sum_{t=1}^{T} \indicator(a_t=1)\inp{\beta-\widehat{\beta}_{t-1}}{X_t}^2}_{\uppercase\expandafter{\romannumeral1}} + \underbrace{\frac{2}{T^{'}}\sum_{t=1}^{T} \indicator(a_t=1) \inp{\beta-\widehat{\beta}_{t-1}}{X_t}\xi_t}_{\uppercase\expandafter{\romannumeral2}} \\ &\quad + \underbrace{\frac{1}{T^{'}}\sum_{t=1}^{T} \indicator(a_t=1) \xi_t^2}_{\uppercase\expandafter{\romannumeral3}}.
    \end{align*}
    For term \uppercase\expandafter{\romannumeral1}, by Assumption \ref{assump:basic}(c), $|\inp{\beta-\widehat{\beta}_{t-1}}{X_t}|$ is bounded by $\|\beta-\widehat{\beta}_{t-1}\|\phi_{\max}(s)\log^{1/2} d=O(\sqrt{\frac{s_0\log^2 d}{t-1}})$ with high probability, then the uniform bound is
    \begin{align*}
        \big|\frac{1}{T^{'}} \indicator(a_t=1)\inp{\beta-\widehat{\beta}_{t-1}}{X_t}^2\big|&\lesssim \frac{1}{T^{'}}\|\beta - \widehat{\beta}_{t-1}\|^2\phi_{\max}^2(s)\log d \lesssim \frac{1}{T^{'}}\frac{s_0\log^2 d\sigma^2}{t-1}\lesssim \frac{s_0\log^2 d\sigma^2}{T^{'}}, 
    \end{align*}
    and similarly,
    \begin{align*}
        \EE\left[\frac{1}{T^{'2}} \indicator(a_t=1)\inp{\beta - \widehat{\beta}_{t-1}}{X_t}^4\big|\mathcal{F}_{t-1}\right]&\leq \frac{1}{T^{'2}}\|\beta - \widehat{\beta}_{t-1}\|^4\phi_{\max}^4(s) \\
        &\lesssim \frac{1}{T^{'2}}\frac{s_0^2\log^4 d\sigma^4}{(t-1)^{2}}.
    \end{align*}
    By martingale Bernstein inequality, with probability at least $1-d_1^{-10}$,
    \begin{align*}
        \uppercase\expandafter{\romannumeral1}&\lesssim \bigg|\frac{1}{T^{'}}\sum_{t=1}^{T} \EE\left[\indicator(a_t=1)\inp{\beta - \widehat{\beta}_{t-1}}{X_t}^2\big|\mathcal{F}_{t-1}\right]\bigg| + \frac{s_0\log^3 d\sigma^2}{T^{'}} + \sqrt{\frac{s_0^2\log^5 d}{T^{'3}}}\sigma^2 \\
        &\lesssim \bigg|\frac{1}{T^{'}}\sum_{t=1}^{T} \inp{\beta - \widehat{\beta}_{t-1}}{X_t}^2\bigg| + \frac{s_0\log^3 d\sigma^2}{T^{'}} + \sqrt{\frac{s_0^2\log^5 d}{T^{'3}}}\sigma^2 \\
        &\lesssim \frac{s_0\log d\log T\sigma^2}{T^{'}} + \frac{s_0\log^3 d\sigma^2}{T^{'}} + \sqrt{\frac{s_0^2\log^5 d}{T^{'3}}}\sigma^2,
    \end{align*}
    which indicates $\uppercase\expandafter{\romannumeral1}\overset{p}{\rightarrow} 0$ when $s_0=o(\frac{T}{\log d})$. 
    
    Similarly, for term $\uppercase\expandafter{\romannumeral2}$, note that $|\xi_t|$ is bounded by $\sigma\log^{1/2} d$ with high probability, then for any $t$, 
    \begin{align*}
        \bigg|\frac{2}{T^{'}} \indicator(a_t=1) \inp{\beta-\widehat{\beta}_{t-1}}{X_t}\xi_t\bigg|\lesssim \frac{1}{T^{'}}\sqrt{\frac{s_0\log^2 d\sigma^4}{(t-1)}}\lesssim \frac{\sqrt{s_0\log^2 d}}{T^{'}}\sigma^2,
    \end{align*}
    and
    \begin{align*}
        \EE\left[\frac{4}{T^{'2}} \frac{\indicator(a_t=1)}{\pi_t^2} \inp{\beta-\widehat{\beta}_{t-1}}{X_t}^2\xi_t^2|\mathcal{F}_{t-1}\right] \lesssim \frac{\sigma^2}{T^{'2}}|\inp{\beta-\widehat{\beta}_{t-1}}{X_t}|^2 \lesssim  \frac{s_0\log^2d \sigma^4}{T^{'2}(t-1)}.
    \end{align*}
    By martingale Bernstein inequality, with probability at least $1-d_1^{-10}$, 
    \begin{align*}
        \bigg|\uppercase\expandafter{\romannumeral2} - \frac{2}{T^{'}} \sum_{t=1}^{T} \EE\left[\frac{\indicator(a_t=1)}{\pi_t}\inp{\beta-\widehat{\beta}_{t-1}}{X_t} \xi_t|\mathcal{F}_{t-1}\right]\bigg|=\left|\uppercase\expandafter{\romannumeral2} \right| \lesssim \frac{\sqrt{s_0\log^4 d}}{T^{'}} + \frac{s_0^{1/2}\log^{3/2} d\log^{1/2} T\sigma^2}{T^{'}},
    \end{align*}
    which indicates $\uppercase\expandafter{\romannumeral2}\overset{p}{\rightarrow} 0$ as long as $s_0=o(\frac{T^2}{\log^3 d})$. 
    
    Lastly, for term $\uppercase\expandafter{\romannumeral3}$,
    \begin{align*}
        \|\frac{1}{T^{'}} \indicator(a_t=1) \xi_t^2\|_{\psi_1}\lesssim \frac{\sigma^2}{T^{'}},
    \end{align*}
    and
    \begin{align*}
        \EE\left[\frac{1}{T^{'2}} \indicator(a_t=1) \xi_t^4|\mathcal{F}_{t-1}\right]  \lesssim  \frac{\sigma^4}{T^{'2}}.
    \end{align*}
    By martingale Bernstein inequality, with probability at least $1-d_1^{-10}$, 
    \begin{align*}
        \bigg|\uppercase\expandafter{\romannumeral3} - \frac{1}{T^{'}} \sum_{t=1}^{T}\EE\left[\indicator(a_t=1) \xi_t^2|\mathcal{F}_{t-1}\right]\bigg| = \bigg|\uppercase\expandafter{\romannumeral3} - \frac{1}{T^{'}} \sum_{t=1}^{T} \EE\left[\indicator(a_t=1)|\mathcal{F}_{t-1}\right]\sigma^2 \bigg| \lesssim \sqrt{\frac{\log d}{T^{'}}}\sigma^2.
    \end{align*}
    By martingale Weak Law of Large Numbers (Theorem 2.19 in \cite{hall2014martingale}), $\frac{\sum_{t=1}^{T} \EE\left[\indicator(a_t=1)|\mathcal{F}_{t-1}\right]}{T'}\overset{p}{\rightarrow} 1$. This indicates $\uppercase\expandafter{\romannumeral3}\overset{p}{\rightarrow} \sigma^2$ as long as $\frac{\log d}{T^{'}} \rightarrow 0$. \\
    {\it Step 2: showing $m_{l,T}^{\top}\widehat{\Lambda}_{T}m_{l,T}\overset{p}{\rightarrow} m_{l}^{\top}\Lambda m_l$.} \\
    Denote $m_l=\Lambda^{-1}e_l$, then $\Lambda_{ll}^{-1}=m_l^{\top}\Lambda m_l$. Similar as the proof of Theorem \ref{thm:maininference-dc}, we have
    \begin{align*}
        \left|m_{l,T}^{\top}\widehat{\Lambda}_T m_{l,T}-m_l^{\top}\Lambda m_l\right|&\leq \underbrace{\left|\inp{\widehat{\Lambda}_T m_{l,T}}{m_{l,T}}-\inp{\Lambda m_{l,T}}{m_{l,T}}\right|}_{(a)} \\ &\quad + \underbrace{\left|\inp{\Lambda m_{l,T}}{m_{l,T}} - \inp{\Lambda m_{l}}{m_{l}}\right|}_{(b)}.
    \end{align*}
    Notice that
    \begin{align*}
        (a)&\leq \|(\widehat{\Lambda}_T - \Lambda)m_{l,T}\|_{\max}\|m_{l,T}\|_{\ell 1} \\
         &\lesssim \|(\widehat{\Lambda}_T - \Lambda)m_{l,T}\|_{\max}\left(\|m_l\|_{\ell 1} + \frac{s_{\Lambda^{-1}}}{\lambda_0}\left(\frac{s_0\log^2 d}{T}\right)^{\nu/2}\right).
    \end{align*}
    Then next Lemma shows the upper bound of $\|M_T\widehat{\Lambda} - M_T\Lambda\|_{\max}$.
    \begin{Lemma}
        \label{lemma:neg1}
        Under the conditions in Theorem \ref{thm:maininference-dc}, for both $i=0,1$,
        \begin{align*}
            \|M_T^{(i)}\widehat{\Lambda}_{i} - M_T^{(i)}\Lambda_i\|_{\max}\lesssim  \frac{s_0^{\nu}s_{\Lambda^{-1}}\log^{2\nu} d}{T^{\nu}} + \left(\frac{s_0\log d}{T}\right)^{\nu/2}
        \end{align*}
    \end{Lemma}
    Also note that $\|m_l\|_{\ell 1}\leq \sqrt{s_{\Lambda^{-1}}}\|m_l\|$ and $\|m_l\|\leq \lambda_0$ is of constant order. Then as long as $s_{\Lambda^{-1}}^{3/2}s_0^{\nu}=o(\frac{T^{\nu}}{\log^2 d})$, $(a)\overset{p}{\rightarrow} 0$ when $T, d\rightarrow \infty$.

    For term $(b)$, 
    \begin{align*}
        (b)&\leq \left|\inp{\Lambda m_{l,T}}{m_{l,T} - m_l}\right| + \left|\inp{\Lambda (m_{l,T}-m_l)}{m_l}\right| \\
        &\leq \|\Lambda\|\|m_{l,T}\|\|m_{l,T} - m_l\| + \|\Lambda\|\|m_l\|\|m_{l,T} - m_l\| \\
        &\lesssim \|\Lambda\|\|m_{l,T} - m_l\|(\|m_l\| + \frac{\sqrt{s_{\Lambda^{-1}}}}{\lambda_0}\left(\frac{s_0\log^2 d}{T}\right)^{\nu/2}) \\
        &\lesssim \|\Lambda\|\frac{\sqrt{s_{\Lambda^{-1}}}}{\lambda_0}\left(\frac{s_0\log^2 d}{T}\right)^{\nu/2}\left(\|m_l\| + \frac{\sqrt{s_{\Lambda^{-1}}}}{\lambda_0}\left(\frac{s_0\log^2 d}{T}\right)^{\nu/2}\right).
    \end{align*}
    Then as long as $s_0^{\nu}s_{\Lambda^{-1}}=o(\frac{T^{\nu}}{\log d})$, $(b)\overset{p}{\rightarrow} 0$ when $T,d\rightarrow \infty$. 
\end{proof}

\subsection{Proof of Corollary \ref{cor:difference-dc}}
\begin{proof}
    The proof follows a similar argument with Corollary \ref{cor:difference-nodc}. For any $l\in [d]$, we have
    \begin{align*}
        (\widehat{\beta}_{1(l)}^{\uaw} - \widehat{\beta}_{0(l)}^{\uaw}) - (\beta_{1(l)} - \beta_{1(l)})&= \underbrace{\frac{1}{T}e_l^{\top}\sum_{t=1}^{T} \indicator(a_t=1)\Lambda_1^{-1} X_t\xi_t}_{V_{1T}} + \underbrace{\frac{1}{T}e_l^{\top}\sum_{t=1}^{T} \indicator(a_t=1)(M_{T}^{(1)} - \Lambda_1^{-1})X_t\xi_t}_{R_{11}} \\ &\quad \underbrace{\frac{1}{T}e_l^{\top}\sum_{t=1}^{T} \left(I-\indicator(a_t=1)M_T^{(1)}X_tX_t^{\top}\right)(\widehat{\beta}_{1,T} - \beta)}_{R_{12}} \\ &\quad - \underbrace{\frac{1}{T}e_l^{\top}\sum_{t=1}^{T} \indicator(a_t=0)\Lambda_0^{-1} X_t\xi_t}_{V_{0T}} + \underbrace{\frac{1}{T}e_l^{\top}\sum_{t=1}^{T} \indicator(a_t=0)(M_{T}^{(0)} - \Lambda_0^{-1})X_t\xi_t}_{R_{01}} \\ &\quad \underbrace{\frac{1}{T}e_l^{\top}\sum_{t=1}^{T} \left(I-\indicator(a_t=0)M_T^{(0)}X_tX_t^{\top}\right)(\widehat{\beta}_{0,T} - \beta)}_{R_{02}}.
    \end{align*}
    In the proof of Theorem \ref{thm:maininference-dc}, we have shown that $R_{11}$, $R_{12}$, $R_{01}$ and $R_{02}$ are negligible compared to the variance of $V_{1T}$ and $V_{0T}$. We can rewrite $V_{1T}-V_{0T}$ as
    \begin{align*}
        \underbrace{\frac{1}{T}\sum_{t_1=1}^{T} \indicator(a_{t_1}=1)m_{l}^{(1)\top}X_{t_1}\xi_{t_1}}_{V_{1T}} - \underbrace{\frac{1}{T}\sum_{t_2=1}^{T} \indicator(a_{t_2}=0)m_{l}^{(0)\top}X_{t_2}\xi_{t_2}}_{V_{0T}},
    \end{align*}
    where $m_l^{(i)}=\Lambda_i^{-1}e_l$ for $i=0,1$. As long as we show $V_{1T}$ is uncorrelated with $V_{0T}$, then the asymptotic variance of $\sqrt{T}(V_{1T}-V_{0T})$is the sum of their individual variance, i.e., $\sigma_1^2S_1^2+\sigma_0^2S_0^2$ with $S_1$ and $S_0$ defined in Theorem \ref{thm:maininference-dc}. Notice that,
    \begin{align*}
        &\quad \frac{1}{T^2}\sum_{t_1=1}^{T} \indicator(a_{t_1}=1)m_{l}^{(1)\top}X_{t_1}\xi_{t_1} \sum_{t_2=1}^{T} \indicator(a_{t_2}=0)m_{l,T}^{(0)\top}X_{t_2}\xi_{t_2} \\
        &=\frac{1}{T^2}\sum_{t_1=1}^{T}\sum_{t_2=1}^{T} \indicator(a_{t_1}=1)\indicator(a_{t_2}=0) m_{l}^{(1)\top}X_{t_1}m_{l}^{(0)\top}X_{t_2}\xi_{t_1}\xi_{t_2} .
    \end{align*}
    When $t_1=t_2$, $\indicator(a_{t_1}=1)\indicator(a_{t_2}=0)=0$. When $t_1\neq t_2$, $\EE[\indicator(a_{t_1}=1)\indicator(a_{t_2}=0) m_{l}^{(1)\top}X_{t_1}m_{l}^{(0)\top}X_{t_2}\xi_{t_1}\xi_{t_2}]=0$ due to the i.i.d. distributed $\xi_t$. As a result, the two terms are uncorrelated. \\
    Then similar as Theorem \ref{thm:stuinference-dc}, we can replace $\sigma_1^2S_1^2+\sigma_0^2S_0^2$ with $\widehat{\sigma}_1^2\widehat{S}_1^2+\widehat{\sigma}_0^2\widehat{S}_0^2$ and conclude the proof.
\end{proof}

\subsection{Proof of Theorem \ref{thm:valueinf}}
\begin{proof}
    Then notice that $\sqrt{T}(\widehat{V}_T - V^{*})$ can be written as
    \begin{align*}
        \sqrt{T}(\widehat{V}_T - V^{*})&=\frac{1}{\sqrt{T}}\sum_{t=1}^{T} \indicator(a_t=a^{*}(X_t))y_t + \indicator(a_t\neq a^{*}(X_t))y_t - \EE[\inp{\beta_{a^{*}(X_t)}}{X_t}] \\
        &= \frac{1}{\sqrt{T}}\sum_{t=1}^{T} \indicator(a_t=a^{*}(X_t))\left(\xi_t + \inp{\beta_{a^{*}(X_t)}}{X_t} \right) + \indicator(a_t\neq a^{*}(X_t))\left(\xi_t + \inp{\beta_{1-a^{*}(X_t)}}{X_t} \right) \\ &\quad - \EE[\inp{\beta_{a^{*}(X_t)}}{X_t}] \\
        &= \underbrace{\frac{1}{\sqrt{T}}\sum_{t=1}^{T}\xi_t + \inp{\beta_{a^{*}(X_t)}}{X_t} -\EE[\inp{\beta_{a^{*}(X_t)}}{X_t}]}_{D_T} \\ &\quad + \underbrace{\frac{1}{\sqrt{T}}\sum_{t=1}^{T} \indicator(a_t\neq a^{*}(X_t))\left(\inp{\beta_{1-a^{*}(X_t)}}{X_t} - \inp{\beta_{a^{*}(X_t)}}{X_t}\right)}_{H_T}. 
    \end{align*}
    Next, we show $D_T$ is the main variance term and $B_T$ is the negligible bias term. \\
    \emph{Step 1: showing $D_T\overset{d.}{\rightarrow} N(0, S_V^2)$} \\
    First, we check the Lindeberg condition. For any $\delta>0$,
    \begin{align*}
        &\frac{1}{T}\sum_{t=1}^{T} \EE\left[\left(\xi_t+\inp{\beta_{a^{*}(X_t)}}{X_t} -\EE[\inp{\beta_{a^{*}(X_t)}}{X_t}]\right)^2 \indicator\left(\frac{1}{\sqrt{T}}\left|\xi_t+\inp{\beta_{a^{*}(X_t)}}{X_t} -\EE[\inp{\beta_{a^{*}(X_t)}}{X_t}] \right|>\delta\right) \big|\mathcal{F}_{t-1}\right] \\
        &\quad \lesssim  \frac{1}{T}\sum_{t=1}^{T} \left((\sigma_1^2\vee\sigma_0^2)+ \phi_{\min}^2(s)\|\beta_1-\beta_0\|^2 \right)\log d \PP\left(\left|\xi_t+\inp{\beta_{a^{*}(X_t)}}{X_t} -\EE[\inp{\beta_{a^{*}(X_t)}}{X_t}] \right|>\sqrt{T}/\delta \right) \\
        &\quad \lesssim \frac{1}{T}\sum_{t=1}^{T} \left((\sigma_1^2\vee\sigma_0^2)+ \phi_{\min}^2(s)\|\beta_1-\beta_0\|^2 \right)\log d\exp\left(-\frac{T}{\delta^2((\sigma_1\vee\sigma_0) + \phi_{\min}(s))}\|\beta_1-\beta_0\|\right),
    \end{align*}
    where the result comes from subGaussian property of $\xi_t$ and $\inp{\beta_{a^{*}(X_t)}}{X_t}$. It is obvious that the above equation goes to 0 when $T\rightarrow \infty$.

    Then we calculate the variance. Since $\xi_t$ and $X_t$ are independent conditioned on the chosen
    action and $a^{\ast}(X_t)$ only depends on $X_t$, we have $\EE[\xi_t\inp{\beta_{a^{\ast}(X_t)}}{X_t}|\mathcal{F}_{t-1}] = 0$. The conditional variance can be expressed as
    \begin{align*}
        \frac{1}{T}\sum_{t=1}^{T}\EE[\xi_t^2|\mathcal{F}_{t-1}]&=\EE[\indicator(a_t=a^{*}(X_t))\sigma_{a^{*}(X_t)}^2|\mathcal{F}_{t-1}] + \EE[\indicator(a_t\neq a^{*}(X_t))\sigma_{1-a^{*}(X_t)}^2|\mathcal{F}_{t-1}] + \text{Var}[\inp{\beta_{a^{*}(X_t)}}{X_t}] \\
        &= \frac{1}{T}\sum_{t=1}^{T} \EE[\indicator(\inp{\beta_1-\beta_0}{X}>0)\indicator(a_t=1)\sigma_1^2|\mathcal{F}_{t-1}]  \\ &\quad + \EE[\indicator(\inp{\beta_1-\beta_0}{X}<0)\indicator(a_t=0)\sigma_0^2|\mathcal{F}_{t-1}] \\ &\quad + \EE[\indicator(a_t\neq a^{*}(X_t))\sigma_{1-a^{*}(X_t)}^2|\mathcal{F}_{t-1}] + \text{Var}[\inp{\beta_{a^{*}(X_t)}}{X_t}].
    \end{align*}
    By Lemma \ref{probright}, when $t$ is large, 
    \begin{align*}
        \EE\left[\indicator(\inp{\beta_1-\beta_0}{X}>0)\indicator(a_t=1)\sigma_1^2|\mathcal{F}_{t-1}\right]&=\EE\left[\indicator((\Delta_{X_t}>0) \cap (|\Delta_{X_t}|> |\Delta_{X_t} - \widehat{\Delta}_{X_t}|))\sigma_1^2\big|\mathcal{F}_{t-1}\right]  \\
        &= \PP(\Delta_{X_t}>0)\PP\left(|\Delta_{X_t}|> |\Delta_{X_t} - \widehat{\Delta}_{X_t}|\big|\mathcal{F}_{t-1}, \Delta_{X_t}>0\right)\sigma_1^2 \\
        &\geq  \PP(\Delta_{X_t}>0)\sigma_1^2\left(1-C_0\left(\frac{s_0\log d}{t-1}\right)^{\nu/2}\right)\rightarrow \PP(\Delta_{X_t}>0)\sigma_1^2.
    \end{align*}
    Similarly, $\EE[\indicator(\inp{\beta_1-\beta_0}{X}<0)\indicator(a_t=0)\sigma_0^2|\mathcal{F}_{t-1}]\rightarrow \PP(\Delta_{X_t}<0)\sigma_0^2$. Moreover, 
    \begin{align*}
        \frac{1}{T} \sum_{t=1}^T \EE[\indicator(a_t\neq a^{*}(X_t))\sigma_{1-a^{*}(X_t)}^2|\mathcal{F}_{t-1}]&\leq \frac{1}{T} \sum_{t=1}^T \PP(a_t\neq a^{*}(X_t)|\mathcal{F}_{t-1})(\sigma_1^2\vee \sigma_0^2) \\
        &\lesssim \frac{1}{T} \sum_{t=T_0}^T \left(\frac{s_0\log d}{t-1}\right)^{\nu/2}(\sigma_1^2\vee \sigma_0^2) \lesssim  \left(\frac{s_0\log d}{T}\right)^{\nu/2}.
    \end{align*}
    When $s_0=o(\frac{T}{\log d})$, the above term is negligible. As a result, the conditional variance converges in probability to $S_V^2=\sigma_1^2\int \indicator(\inp{\beta_1-\beta_0}{X}>0)d\mathcal{P}_{\mathcal{X}} + \sigma_0^2\int \indicator(\inp{\beta_1-\beta_0}{X}<0)d\mathcal{P}_{\mathcal{X}} + \text{Var}[\inp{\beta_{a^{*}(X)}}{X}]$. \\
    \emph{Step 2: showing $H_T=o_p(1)$} \\
    Under Lemma \ref{probright} and Assumption \ref{assump:margincondition}, we have
    \begin{align*}
        &\frac{1}{\sqrt{T}}\sum_{t=1}^{T} \EE\left[\indicator(a_t\neq a^{*}(X_t))\left(\inp{\beta_{1-a^{*}(X_t)}}{X_t} - \inp{\beta_{1-a^{*}(X_t)}}{X_t}\right)|\mathcal{F}_{t-1}\right] \\
        &\quad \leq \frac{1}{\sqrt{T}}\sum_{t=1}^{T} \PP\left(|\Delta_{X_t}|< |\widehat{\Delta}_{X_t} - \Delta_{X_t}| |\mathcal{F}_{t-1}\right)|\widehat{\Delta}_{X_t} - \Delta_{X_t}| \\
        &\quad \lesssim \frac{1}{\sqrt{T}}\sum_{t=T_0}^{T} \PP\left(\indicator(a_t\neq a^{*}(X_t)) |\mathcal{F}_{t-1}\right) \sqrt{\frac{s_0\log d}{t-1}} \\
        &\quad \lesssim \frac{1}{\sqrt{T}}\sum_{t=T_0}^{T} \left(\frac{s_0\log d}{t-1}\right)^{(1+\nu)/2} \lesssim \sqrt{\frac{s_0^{1+\nu}\log^{1+\nu} d}{T^{\nu}}}.
    \end{align*}
    Also notice that, for any $T_0\leq t\leq T$, the uniform bound is
    \begin{align*}
         \frac{1}{\sqrt{T}}\indicator(a_t\neq a^{*}(X_t))|\widehat{\Delta}_{X_t} - \Delta_{X_t}|\lesssim \frac{1}{\sqrt{T}}\sqrt{\frac{s_0\log d}{t-1}}\lesssim \sqrt{\frac{s_0\log d}{T}}
    \end{align*}
    and the conditional variance is
    \begin{align*}
         &\frac{1}{T}\sum_{t=2}^{T} \EE\left[\indicator(a_t\neq a^{*}(X_t))\left(\inp{\beta_{1-a^{*}(X_t)}}{X_t} - \inp{\beta_{1-a^{*}(X_t)}}{X_t}\right)^2|\mathcal{F}_{t-1}\right] \\
         &\quad \lesssim \frac{1}{T}\sum_{t=T_0}^{T} \PP\left(|\Delta_{X_t}|< |\widehat{\Delta}_{X_t} - \Delta_{X_t}| |\mathcal{F}_{t-1}\right)|\widehat{\Delta}_{X_t} - \Delta_{X_t}|^2 \\
         &\quad \lesssim \frac{1}{T} \sum_{t=T_0}^{T} \left(\frac{s_0\log d}{t-1}\right)^{1+\nu/2} \lesssim \sqrt{\frac{s_0^3\log^3 d}{T^{2+\nu}}}.    
    \end{align*}
    By Bernstein inequality, with probability at least $1-d^{-10}$, $H_T\lesssim \sqrt{\frac{s_0^{1+\nu}\log^{1+\nu} d}{T^{\nu}}} + \sqrt{\frac{s_0\log^3 d}{T}} + \sqrt{\frac{s_0^3\log^4 d}{T^{2+\nu}}}$. As long as $s_0=o(\frac{T^{\nu/(1+\nu)}}{\log d})$, $H_T$ converges to 0 in probability. 
\end{proof}

\subsection{Proof of Theorem \ref{thm:stuvalueinf}}
\begin{proof}
    Since
    \begin{align*}
        \widehat{S}_V^2&=\underbrace{\frac{1}{T}\sum_{t=1}^T \widehat{\sigma}_{1,T}^2\indicator(a_t=1) + \widehat{\sigma}_{0,T}^2\indicator(a_t=0)}_{G_T} \\ &\quad + \underbrace{\frac{1}{T} \sum_{t=1}^T \inp{\widehat{\beta}_{a_t,t-1}}{X_t}^2 - \left(\frac{1}{T-T_0} \sum_{t=1}^T \inp{\widehat{\beta}_{a_t,t-1}}{X_t}\right)^2}_{W_T},
    \end{align*}
    we aim to show $G_T$ is consistent for $\sigma_1^2\int \indicator(\inp{\beta_1-\beta_0}{X}>0)d\mathcal{P}_{\mathcal{X}} + \sigma_0^2\int \indicator(\inp{\beta_1-\beta_0}{X}<0) d\mathcal{P}_{\mathcal{X}}$ and $W_t$ is consistent for $\text{Var}[\inp{\beta_{a^{*}(X)}}{X}]$, then $\widehat{S}_V^2$ is consistent for $S_V^2$. By Slutsky's Theorem, we obtain the desired result. \\
    \emph{Step 1: consistency of $G_T$} 
    \begin{align*}
        &G_T - \left(\sigma_1^2\int \indicator(\inp{\beta_1-\beta_0}{X}>0)d\mathcal{P}_{\mathcal{X}} + \sigma_0^2\int \indicator(\inp{\beta_1-\beta_0}{X}<0) d\mathcal{P}_{\mathcal{X}}\right) \\ &\quad = \frac{1}{T}\sum_{t=1}^T \widehat{\sigma}_{1,T}^2\indicator(a_t=1) - \sigma_1^2\int \indicator(\inp{\beta_1-\beta_0}{X}>0)d\mathcal{P}_{\mathcal{X}} \\
        &\quad\quad  + \frac{1}{T}\sum_{t=1}^T \widehat{\sigma}_{0,T}^2\indicator(a_t=0) - \sigma_0^2\int \indicator(\inp{\beta_1-\beta_0}{X}<0)d\mathcal{P}_{\mathcal{X}},
    \end{align*}
    and 
    \begin{align*}
        &\frac{1}{T}\sum_{t=1}^T \widehat{\sigma}_{1,T}^2\indicator(a_t=1) - \sigma_1^2\int \indicator(\inp{\beta_1-\beta_0}{X}>0)d\mathcal{P}_{\mathcal{X}} \\
        &\quad = \underbrace{\frac{1}{T}\sum_{t=1}^T (\widehat{\sigma}_{1,T}^2 - \sigma_1^2)\indicator(a_t=1)}_{(a)} + \underbrace{\frac{1}{T}\sum_{t=1}^T \sigma_1^2\left(\indicator(a_t=1) - \indicator(a^{*}(X_t)=1)\right)}_{(b)} \\
        &\quad\quad + \underbrace{\frac{1}{T}\sum_{t=1}^T \sigma_1^2\left(\indicator(a^{*}(X_t)=1) - \int \indicator(\inp{\beta_1-\beta_0}{X}>0)d\mathcal{P}_{\mathcal{X}}\right)}_{(c)}. 
    \end{align*}
    From the result in Theorem \ref{thm:stuinference-dc}, $\widehat{\sigma}_{1,T}^2$ is consistent for $\sigma_1^2$. Since $\frac{1}{T}\sum_{t=1}^T \indicator(a_t=1)$ is bounded by 1, $(a)$ will converge to 0 in probability.

    For term $(c)$, $\EE[\indicator(a^{*}(X_t)=1)]=\EE[\indicator(\inp{\beta_1-\beta_0}{X_t}>0)]=\int \indicator(\inp{\beta_1-\beta_0}{X}>0)d\mathcal{P}_{\mathcal{X}}$. Then by Weak Law of Large Numbers, $(c)$ converges to 0 in probability. 
    
    For term $(b)$, note that under Lemma \ref{probright},
    \begin{align*}
        \frac{1}{T}\sum_{t=1}^{T} \EE\left[\indicator(a_t=1) - \indicator(a^{*}(X_t)=1)|\mathcal{F}_{t-1}\right]&= \frac{1}{T}\sum_{t=1}^{T} 2\PP\left(a_t\neq a^{*}(X_t)\big|\mathcal{F}_{t-1}\right) \\
        &= \frac{1}{T}\sum_{t=1}^{T} 2\PP\left(|\Delta_{X_t}|<|\Delta_{X_t} - \widehat{\Delta}_{X_t}| \big|\mathcal{F}_{t-1}\right) \\
        &\lesssim \frac{1}{T}\sum_{t=T_0}^{T}\left(\frac{s_0\log d}{t-1}\right)^{\nu/2}\lesssim \left(\frac{s_0\log d}{T}\right)^{\nu/2}.
    \end{align*}
    Moreover, for any $1\leq t\leq T$, the uniform bound is $\frac{2}{T}$, and the conditional variance is
    \begin{align*}
        \frac{1}{T^2}\sum_{t=1}^{T} \EE\left[\left(\indicator(a_t=1) - \indicator(a^{*}(X_t)=1)\right)^2|\mathcal{F}_{t-1}\right]&= \frac{1}{T^2}\sum_{t=1}^{T} 4\PP\left(a_t\neq a^{*}(X_t)\big|\mathcal{F}_{t-1}\right) \\
        &= \frac{1}{T^2}\sum_{t=1}^{T} 4\PP\left(|\Delta_{X_t}|<|\Delta_{X_t} - \widehat{\Delta}_{X_t}| \big|\mathcal{F}_{t-1}\right) \\
        &\lesssim \frac{1}{T^2}\sum_{t=T_0}^{T}\left(\frac{s_0\log d}{t-1}\right)^{\nu/2}\lesssim \sqrt{\frac{s_0^{\nu}\log^{\nu} d}{T^{2+\nu}}}.
    \end{align*}
    By Bernstein inequality, with probability at least $1-d^{-10}$, $(b)\lesssim \left(\frac{s_0\log d}{T}\right)^{\nu/2} + \frac{\log d}{T} + \left(\frac{s_0^{\nu}\log^{2+\nu} d}{T^{2+\nu}}\right)^{1/4}$. As long as $s_0=o(\frac{T^{(2+\nu)/\nu}}{\log^2 d})$, $(b)$ will converge to 0 in probability. Following the same arguments, we can show $\frac{1}{T}\sum_{t=1}^T \widehat{\sigma}_{0,T}^2\indicator(a_t=0)$ is consistent for $\sigma_0^2\int \indicator(\inp{\beta_1-\beta_0}{X}<0)d\mathcal{P}_{\mathcal{X}}$. \\
    \emph{Step 2: consistency of $W_T$} \\
    First, we aim to show $\frac{1}{T} \sum_{t=1}^T \inp{\widehat{\beta}_{a_t,t-1}}{X_t}^2\overset{p}{\rightarrow} \EE[\inp{\beta_{a^{*}(X)}}{X}^2]$. Notice that
    \begin{align*}
        \frac{1}{T} \sum_{t=1}^T \inp{\widehat{\beta}_{a_t,t-1}}{X_t}^2- \EE[\inp{\beta_{a^{*}(X)}}{X}^2]&= \frac{1}{T} \sum_{t=1}^T \inp{\widehat{\beta}_{a_t,t-1}}{X_t}^2 - \frac{1}{T} \sum_{t=1}^T \inp{{\beta}_{a^{*}(X_t)}}{X_t}^2 \\ &\quad + \frac{1}{T} \sum_{t=1}^T \inp{{\beta}_{a^{*}(X_t)}}{X_t}^2 - \EE[\inp{\beta_{a^{*}(X)}}{X}^2].
    \end{align*}
    We apply martingale Bernstein inequality to show $\frac{1}{T} \sum_{t=1}^T \inp{\widehat{\beta}_{a_t,t-1}}{X_t}^2 \overset{p}{\rightarrow} \frac{1}{T} \sum_{t=1}^T \inp{{\beta}_{a^{*}(X_t)}}{X_t}^2$. The conditional expectation is 
    \begin{align*}
        &\quad \frac{1}{T}\sum_{t=1}^{T} \EE\left[\inp{\widehat{\beta}_{a_t,t-1}}{X_t}^2 - \inp{{\beta}_{a^{*}(X_t)}}{X_t}^2|\mathcal{F}_{t-1}\right] \\ &= \frac{1}{T}\sum_{t=1}^{T} \EE\left[\indicator(a_t=a^{*}(X_t))\left(\inp{\widehat{\beta}_{a^{*}(X_t)}}{X_t}^2 - \inp{{\beta}_{a^{*}(X_t)}}{X_t}^2\right)|\mathcal{F}_{t-1}\right] \\ &\quad + \frac{1}{T}\sum_{t=1}^{T} \EE\left[\indicator(a_t\neq a^{*}(X_t))\left(\inp{\widehat{\beta}_{1-a^{*}(X_t)}}{X_t}^2 - \inp{{\beta}_{a^{*}(X_t)}}{X_t}^2\right)|\mathcal{F}_{t-1}\right] \\
        &\leq \frac{1}{T}\sum_{t=1}^{T} \EE\left[\indicator(a_t=a^{*}(X_t))\left(\inp{\widehat{\beta}_{a^{*}(X_t)}}{X_t}^2 - \inp{{\beta}_{a^{*}(X_t)}}{X_t}^2\right)|\mathcal{F}_{t-1}\right] \\ &\quad + \frac{1}{T}\sum_{t=1}^{T} \EE\left[\indicator(a_t\neq a^{*}(X_t))\left(\inp{\widehat{\beta}_{1-a^{*}(X_t)}}{X_t}^2 - \inp{{\beta}_{1 - a^{*}(X_t)}}{X_t}^2 + \inp{{\beta}_{1 - a^{*}(X_t)}}{X_t}^2 - \inp{{\beta}_{a^{*}(X_t)}}{X_t}^2\right)|\mathcal{F}_{t-1}\right] \\
        &\leq \frac{1}{T}\sum_{t=1}^{T} \max_{i\in \{0,1\}} \EE\left[\left(\inp{\widehat{\beta}_{i,t-1}}{X_t} - \inp{{\beta}_{i}}{X_t}\right)\left(\inp{\widehat{\beta}_{i,t-1}}{X_t} + \inp{{\beta}_{i}}{X_t}\right)|\mathcal{F}_{t-1}\right] \\ &\quad + \frac{1}{T}\sum_{t=1}^{T} \EE\left[\indicator(a_t\neq a^{*}(X_t))\left(\inp{{\beta}_{1}}{X_t} - \inp{{\beta}_{0}}{X_t}\right)\left(\inp{{\beta}_{1}}{X_t} + \inp{{\beta}_{0}}{X_t}\right)|\mathcal{F}_{t-1}\right].
    \end{align*}
    For both $i=0,1$ and any $T_0\leq t\leq T$,
    \begin{align*}
        \EE\left[\left(\inp{\widehat{\beta}_{i,t-1}}{X_t} - \inp{{\beta}_{i}}{X_t}\right)\left(\inp{\widehat{\beta}_{i}}{X_t} + \inp{{\beta}_{i}}{X_t}\right)|\mathcal{F}_{t-1}\right]&\leq \left|\inp{\widehat{\beta}_{i,t-1} - \beta_{i}}{X_t}\right|\left|\inp{\widehat{\beta}_{i,t-1} + \beta_{i}}{X_t}\right| \\
        &\lesssim \|\widehat{\beta}_{i,t-1} - \beta_{i}\|\phi_{\max}(s)\log^{1/2}d\|\widehat{\beta}_{i,t-1} + \beta_{i}\|\|X_{t_{S}}\| \\
        &\lesssim \sqrt{\frac{s_0\log^2 d}{t-1}}\left(2\|\beta\|+\sqrt{\frac{s_0\log d}{t-1}}\right)\sqrt{s}.
    \end{align*}
    where the second inequality comes from Assumption \ref{assump:basic}(c) and Cauchy Schwarz inequality, and $S=\text{supp}(\widehat{\beta}_{i,t-1}+\beta)$ with $|S|=s$. Since $\|\beta_i\|$ is bounded, 
    \begin{align*}
        \frac{1}{T}\sum_{t=1}^{T} \EE\left[\left(\inp{\widehat{\beta}_{i,t-1}}{X_t} - \inp{{\beta}_{i}}{X_t}\right)\left(\inp{\widehat{\beta}_{i,t-1}}{X_t} + \inp{{\beta}_{i}}{X_t}\right)|\mathcal{F}_{t-1}\right]\lesssim \frac{s_0\log d}{\sqrt{T}} + \frac{s_0^{3/2}\log^{3/2} d\log T}{T}.
    \end{align*}
    Moreover, 
    \begin{align*}
        &\frac{1}{T}\sum_{t=1}^{T} \EE\left[\indicator(a_t\neq a^{*}(X_t))\left(\inp{{\beta}_{1}}{X_t} - \inp{{\beta}_{0}}{X_t}\right)\left(\inp{{\beta}_{1}}{X_t} + \inp{{\beta}_{0}}{X_t}\right)|\mathcal{F}_{t-1}\right] \\
        &\quad \leq \frac{1}{T}\sum_{t=1}^{T}\left|\inp{\beta_1+\beta_0}{X_t}\right|\EE\left[\indicator(a_t\neq a^{*}(X_t))\inp{{\beta}_{1}-\beta_0}{X_t}|\mathcal{F}_{t-1}\right] \\
        &\quad \lesssim \frac{1}{T}\sum_{t=T_0}^{T}\|\beta_1+\beta_0\|\sqrt{2s_0}\left(\frac{s_0\log d}{t-1}\right)^{(1+\nu)/2}\lesssim \sqrt{\frac{s_0^{2+\nu}\log^{1+\nu} d}{T^{1+\nu}}},
    \end{align*}
    where the second last inequality exactly follows the analysis in the proof of Theorem \ref{thm:valueinf}. 
    In the same way, we can calculate the conditional variance, 
    \begin{align*}
        &\quad \frac{1}{T}\sum_{t=1}^{T} \EE\left[\left(\inp{\widehat{\beta}_{a_t,t-1}}{X_t}^2 - \inp{{\beta}_{a^{*}(X_t)}}{X_t}^2\right)^2|\mathcal{F}_{t-1}\right] \\
        &\leq \frac{1}{T}\sum_{t=1}^{T} \max_{i\in \{0,1\}} \EE\left[\left(\inp{\widehat{\beta}_{i,t-1}}{X_t} - \inp{{\beta}_{i}}{X_t}\right)^2\left(\inp{\widehat{\beta}_{i,t-1}}{X_t} + \inp{{\beta}_{i}}{X_t}\right)^2|\mathcal{F}_{t-1}\right] \\ &\quad + \frac{1}{T}\sum_{t=1}^{T} \EE\left[\indicator(a_t\neq a^{*}(X_t))\left(\inp{{\beta}_{1}}{X_t} - \inp{{\beta}_{0}}{X_t}\right)^2\left(\inp{{\beta}_{1}}{X_t} + \inp{{\beta}_{0}}{X_t}\right)^2|\mathcal{F}_{t-1}\right] \\
        &\lesssim \frac{1}{T}\sum_{t=1}^{T} \max_{i\in \{0,1\}} \|\widehat{\beta}_{i,t-1} - \beta_{i}\|^2\phi_{\max}^2(s)\log d\|\widehat{\beta}_{i,t-1} + \beta_{i}\|^2\|X_{t_{S}}\|^2 \\ &\quad + \|\beta_1+\beta_0\|^2 2s_0 \left(\frac{s_0\log d}{t-1}\right)^{1+\nu/2} \\
        &\lesssim \frac{1}{T}\sum_{t=T_0}^{T} \frac{s_0^2\log^2 d}{t-1} + \frac{s_0^{2+\nu/2}\log^{1+\nu/2}d}{(t-1)^{1+\nu/2}}\lesssim  \frac{s_0^2\log^2 d\log T}{T} + \frac{s_0^{2+\nu/2}\log^{1+\nu/2} d}{T^{1+\nu/2}}.
    \end{align*}
    Then by martingale Bernstein inequality, with probability at least $1-d^{-10}$, $\frac{1}{T}\sum_{t=1}^{T}\inp{\widehat{\beta}_{a_t,t-1}}{X_t}^2 - \inp{{\beta}_{a^{*}(X_t)}}{X_t}^2\lesssim \frac{s_0\log d}{\sqrt{T}}+\frac{s_0^{3/2}\log d\log T}{T} + \sqrt{\frac{s_0^2\log^3 d\log T}{T} + \frac{s_0^{2+\nu/2}\log^{2+\nu/2} d}{T^{1+\nu/2}}}$. Above all, as long as $s_0=o(\sqrt{\frac{T}{\log^2 d}})$, $\frac{1}{T}\sum_{t=1}^{T}\inp{\widehat{\beta}_{a_t,t-1}}{X_t}^2 - \inp{{\beta}_{a^{*}(X_t)}}{X_t}^2$ converges to 0 in probability.

    Also note that $\EE[\inp{\beta_{a^{*}(X)}}{X}^4]$ is finite. Then by Weak Law of Large Numbers, $\frac{1}{T} \sum_{t=1}^T \inp{{\beta}_{a^{*}(X_t)}}{X_t}^2 \overset{p}{\rightarrow} \EE[\inp{\beta_{a^{*}(X)}}{X}^2]$. 

    Following almost the same arguments, we can show $\frac{1}{T} \sum_{t=1}^T \inp{\widehat{\beta}_{a_t,t-1}}{X_t}\overset{p}{\rightarrow} \EE[\inp{\beta_{a^{*}(X)}}{X}]$. The only difference is without all the terms corresponding to $\inp{\widehat{\beta}_{i}}{X_t} + \inp{{\beta}_{i}}{X_t}$. Then by continuous mapping theorem, $\left(\frac{1}{T} \sum_{t=1}^T \inp{\widehat{\beta}_{a_t,t-1}}{X_t}\right)^2\overset{p}{\rightarrow} \EE[\inp{\beta_{a^{*}(X)}}{X}]^2$.
\end{proof}

\section{Proofs of Technical Lemmas}

% \subsection{Proof of Lemma \ref{lemma:identifibility}}
% \begin{proof}
%     Without loss of generality, we only prove $i=1$ case. Notice that for any \red{$X\sim F(X)$},
%     \begin{align*}
%         \inp{\widehat{\beta}_{1,t} - \widehat{\beta}_{0,t}}{X} = \inp{\beta_1-\beta_0}{X} + \inp{\widehat{\beta}_{1,t} - \beta_1}{X} + \inp{\beta_0-\widehat{\beta}_{0,t}}{X}.
%     \end{align*}
%     By Assumption \ref{assump:marginalsubgaussian}, $|\inp{\widehat{\beta}_{i,t} - \beta_i}{X}|\lesssim \|\widehat{\beta}_{i,t} - \beta_i\|_2\log^{1/2}d\lesssim \sqrt{\frac{s_0\log^2d}{t}}$ with probability at least $1-d^{-11}$ for $i=0,1$. Combine with Assumption \ref{assump:arm-opt}, then when $t>T_0$, with probability at least $1-2d^{-11}$,  
%     \begin{align*}
%         \indicator(\inp{\widehat{\beta}_{1,t} - \widehat{\beta}_{0,t}}{X}>0) = \indicator(\inp{\beta_1-\beta_0}{X}>0).
%     \end{align*} 
% \end{proof}

\subsection{Proof of Lemma \ref{lemma:feasible1}}
\begin{proof}
    For any $l,j\in [d]$ and $1\leq k\leq t$, it is obvious that $\EE[e_l^{\top}\Omega X_kX_k^{\top}e_j - I_{(lj)}]=0$. Denote $m_l=\Omega e_l$, notice under Assumption \ref{assump:basic}(c), the uniform bound is
    \begin{align*}
        |e_l^{\top}\Omega X_kX_k^{\top}e_j|\leq |e_l^{\top}\Omega X_k||X_k^{\top}e_j|\lesssim \|m_l\|\phi_{\max}(s_{\Omega})\log^{1/2}d, 
    \end{align*}
    and the variance
    \begin{align*}
        \EE\left[(e_l^{\top}\Omega X_kX_k^{\top}e_j)^2\right]\leq D^2\EE\left[(e_l^{\top}\Omega X_k)^2\right]\leq D^2m_l^{\top}\Sigma m_l\lesssim \Omega_{ll}.
    \end{align*}
    By Assumption \ref{assump:basic}, $\|m_l\|\leq C_{\min}$ is bounded. By Bernstein inequality and combine with a union bound, with probability at least $1-d^{-10}$,
    \begin{align*}
        \|I - \Omega\widehat{\Sigma}_T\|_{\max}\lesssim \sqrt{\frac{\log d}{T}}.
    \end{align*}
\end{proof}

\subsection{Proof of Lemma \ref{lemma:mlconvergence1}}
\begin{proof}
    Let $L(m)=\frac{1}{2}m^{\top}\widehat{\Sigma}_{T}m - \inp{m}{e_l} + \mu_{T_1}\|m\|_{\ell_1}$ be the objective function. Denote $v=m_{l,T}-m_l$, using the fact that $L(m_{l,T})\leq L(m_l)$, we have
    \begin{equation}
        \frac{1}{2}v^{\top}\widehat{\Sigma}_T v\leq \inp{v}{e_l - \widehat{\Sigma}_t m_l} + \mu_{T_1}(\|m_l\|_{\ell 1} - \|m_{l,t}\|_{\ell 1}) \label{optimization11}.
    \end{equation}
    We first derive the upper bound of the right hand side. By Lemma \ref{lemma:feasible1}, with high probability,
    \begin{align*}
        \inp{v}{e_l - \widehat{\Sigma}_t m_l}&\leq \|v\|_{\ell 1}\|e_l - \widehat{\Sigma}_t m_l\|_{\max} \\
        &\leq \|v\|_{\ell 1}\|\frac{1}{2}\mu_{T_1} = (\|v_S\|_{\ell 1} + \|v_{S^c}\|_{\ell 1})\frac{1}{2}\mu_{T_1},
    \end{align*}
    where $S=\text{supp}(\Omega e_{l})$ and $S^c$ is its complement, hence $|S|\leq s_{\Omega}$. For the other term,
    \begin{align*}
        \|m_{l,t}\|_{\ell 1} - \|m_{l}\|_{\ell 1}&= \|m_{l} + v\|_{\ell 1} - \|m_{l}\|_{\ell 1} \\
        &\geq  (\|m_{l,S}\|_{\ell 1} - \|v_S\|_{\ell 1}) + \|v_{S^c}\|_{\ell 1} -\|m_{l}\|_{\ell 1} \\
        &= \|v_{S^c}\|_{\ell 1}  -  \|v_S\|_{\ell 1}.
    \end{align*}
    Then the RHS of (\ref{optimization11}) is bounded by
    \begin{align*}
        (\|v_S\|_{\ell 1} + \|v_{S^c}\|_{\ell 1})\frac{1}{2}\mu_{T_1} + (\|v_S\|_{\ell 1} - \|v_{S^c}\|_{\ell 1} )\mu_{T_1} = \frac{3}{2}\mu_{T_1}\|v_S\|_{\ell 1} - \frac{1}{2}\mu_{T_1}\|v_{S^c}\|_{\ell 1}.
    \end{align*}
    Given that $\widehat{\Sigma}_T$ is positive semi-definite, the LHS of (\ref{optimization11}) is non-negative, leading to $\|v_{S^c}\|_{\ell 1}\leq 3\|v_S\|_{\ell 1}$. As a result,
    \begin{align*}
        \|v\|_{\ell 1}\leq 4\|v_S\|_{\ell 1}\leq 4\sqrt{s_{\Omega}}\|v_S\|\leq 4\sqrt{s_{\Omega}}\|v\|.
    \end{align*}
    Next by Assumption \ref{assump:basic} and following the same arguments in \cite{buhlmann2011statistics}, 
    \begin{align*}
        v^{\top}\widehat{\Sigma}_T v\geq \phi_{\min}(8s_\Omega)\|v\|^2.
    \end{align*}
    Combine the above inequalities, we can obtain
    \begin{align*}
        \frac{1}{2}\phi_{\min}(8s_\Omega)\|v\|^2\leq \frac{3}{2}\mu_{T_1}\|v_S\|_{\ell 1}\leq \frac{3}{2}\mu_{T_1}\sqrt{s_{\Omega}}\|v\|.
    \end{align*}
    Therefore,
    \begin{align*}
        &\|v\|\leq \frac{3\sqrt{s_{\Omega}}}{\phi_{\min}(8s_{\Omega})}\mu_{T_1} \\
        &\|v\|_{\ell 1}\leq \frac{12s_{\Omega}}{\phi_{\min}(8s_{\Omega})}\mu_{T_1}.      
    \end{align*}
\end{proof}

\subsection{Proof of Lemma \ref{lemma:feasible2}}
\begin{proof}
    Without loss of generality, we only prove $i=1$ case and omit all the subscript $i$. For notation simplicity, we denote $m_l=\Lambda_i^{-1} e_l$ for $l\in [d]$. Notice that
    \begin{align*}
        \|I - \Lambda^{-1}\widehat{\Lambda}_T\|_{\max}\leq \left\|I - \Lambda^{-1}\frac{1}{T}\sum_{t=1}^{T}\Lambda_t\right\|_{\max} + \left\|\Lambda^{-1}\frac{1}{T}\sum_{t=1}^{T}\Lambda_t - \Lambda^{-1}\widehat{\Lambda}_T\right\|_{\max},
    \end{align*} 
    where $\Lambda_t=\EE[\indicator(a_t=1)X_tX_t^{\top}|\mathcal{F}_{t-1}]$. The first term satisfies $\|I - \Lambda^{-1}\frac{1}{T}\sum_{t=1}^{T}\Lambda_t\|_{\max}\leq \frac{1}{T}\sum_{t=1}^{T}\|I-\Lambda^{-1}\Lambda_t\|_{\max}$. The next Lemma shows the upper bound of $\|I-\Lambda^{-1}\Lambda_t\|$.
    \begin{Lemma}
        \label{lambdaconvergence3}
        Under the conditions in Theorem \ref{thm:maininference-dc}, when $t\geq T_0$, for $i=1,0$, 
        \begin{align*}
            \|I - \Lambda_i^{-1}\Lambda_{i,t}\|_{\max}\lesssim \left(\frac{s_0\log^2 d}{t-1}\right)^{\nu/2}.
        \end{align*}
    \end{Lemma}
    As a result, $\|I - \Lambda^{-1}\frac{1}{T}\sum_{t=1}^{T}\Lambda_t\|_{\max}\lesssim (\frac{s_0\log d}{T})^{\nu/2}$. 

    Then we analysis the second term. For any $l,j\in [d]$ and $1\leq t\leq T$, by Assumption \ref{assump:basic}(c) and $\|m_l\|\leq \lambda_0$ is bounded,
    \begin{align*}
        |\indicator(a_t=1)e_l^{\top}\Lambda^{-1}X_kX_k^{\top}e_j|\leq |m_l^{\top}X_k||X_k^{\top}e_j|\leq \|m_l\|\phi_{\max}(s_{\Lambda^{-1}})\log^{1/2}d\|X_t\|_{\max}\lesssim \phi_{\max}(s_{\Lambda^{-1}})\log^{1/2}d, 
    \end{align*}
    and 
    \begin{align*}
        \EE\left[\indicator(a_k=1)(e_l^{\top}\Lambda^{-1}X_kX_k^{\top}e_j)^2|\mathcal{F}_{k-1}\right]\leq |\indicator(a_t=1)e_l^{\top}\Lambda^{-1}X_kX_k^{\top}e_j|^2\lesssim \phi_{\max}^2(s_{\Lambda^{-1}})\log d. 
    \end{align*}
    By martingale Bernstein inequality and combine with a union bound, with probability at least $1-d^{-10}$,
    \begin{align*}
        \left\|\Lambda^{-1}\frac{1}{T}\sum_{t=1}^{T}\Lambda_t - \Lambda^{-1}\widehat{\Lambda}_T\right\|_{\max}\lesssim \frac{\phi_{\max}(s_{\Lambda^{-1}})\log^{3/2}d}{T} + \sqrt{\frac{\phi_{\max}(s_{\Lambda^{-1}})\log^2 d}{T}}.
    \end{align*}
   Obviously, this term can be dominated by the former one. Then we conclude the proof. 
\end{proof}

\subsection{Proof of Lemma \ref{lemma:mlconvergence2}}
\begin{proof}
    The proof almost follows the same arguments with Lemma \ref{lemma:mlconvergence1}. Without loss of generality, we only prove $i=1$ case and omit all the subscript. Let $L(m)=\frac{1}{2}m^{\top}\widehat{\Lambda}_{T}m - \inp{m}{e_l} + \mu_{T_2}\|m\|_{\ell_1}$ be the objective function. Denote $v=m_{l,T}-m_l$, where $m_l=\Lambda_i^{-1}e_l$. Using the fact that $L(m_{l,T})\leq L(m_l)$, we have
    \begin{equation}
        \frac{1}{2}v^{\top}\widehat{\Lambda}_T v\leq \inp{v}{e_l - \widehat{\Lambda}_T m_l} + \mu_{T_2}(\|m_l\|_{\ell 1} - \|m_{l,T}\|_{\ell 1}) \label{optimization22}.
    \end{equation}
    We first derive the upper bound of the right hand side. By Lemma \ref{lemma:feasible2}, with high probability,
    \begin{align*}
        \inp{v}{e_l - \widehat{\Lambda}_T m_l}&\leq \|v\|_{\ell 1}\|e_l - \widehat{\Lambda}_T m_l\|_{\max} \\
        &\leq \|v\|_{\ell 1}\|\frac{1}{2}\mu_{T_2} = (\|v_S\|_{\ell 1} + \|v_{S^c}\|_{\ell 1})\frac{1}{2}\mu_{T_2},
    \end{align*}
    where $S=\text{supp}(\Lambda^{-1}e_{l})$ and $S^c$ is its complement, hence $|S|\leq s_{\Lambda^{-1}}$. For the other term,
    \begin{align*}
        \|m_{l,t}\|_{\ell 1} - \|m_{l}\|_{\ell 1}&= \|m_{l} + v\|_{\ell 1} - \|m_{l}\|_{\ell 1} \\
        &\geq  (\|m_{l,S}\|_{\ell 1} - \|v_S\|_{\ell 1}) + \|v_{S^c}\|_{\ell 1} -\|m_{l}\|_{\ell 1} \\
        &= \|v_{S^c}\|_{\ell 1}  -  \|v_S\|_{\ell 1}.
    \end{align*}
    Then the RHS of (\ref{optimization22}) is bounded by
    \begin{align*}
        (\|v_S\|_{\ell 1} + \|v_{S^c}\|_{\ell 1})\frac{1}{2}\mu_{T_2} + (\|v_S\|_{\ell 1} - \|v_{S^c}\|_{\ell 1} )\mu_{T_2} = \frac{3}{2}\mu_{T_2}\|v_S\|_{\ell 1} - \frac{1}{2}\mu_{T_2}\|v_{S^c}\|_{\ell 1}.
    \end{align*}
    Given that $\widehat{\Lambda}_t$ is positive semi-definite, the LHS of (\ref{optimization22}) is non-negative, leading to $\|v_{S^c}\|_{\ell 1}\leq 3\|v_S\|_{\ell 1}$. As a result,
    \begin{align*}
        \|v\|_{\ell 1}\leq 4\|v_S\|_{\ell 1}\leq 4\sqrt{s_{\Lambda^{-1}}}\|v_S\|\leq 4\sqrt{s_{\Lambda^{-1}}}\|v\|.
    \end{align*}
    Next by Assumption \ref{covariate diversity} and following the similar arguments in \cite{ren2024dynamic}, 
    \begin{align*}
        v^{\top}\widehat{\Lambda}_{T} v\geq \lambda_0\|v\|^2.
    \end{align*}
    Combine the above inequalities, we can obtain
    \begin{align*}
        \frac{1}{2}\lambda_0\|v\|^2\leq \frac{3}{2}\mu_{T_2}\|v_S\|_{\ell 1}\leq \frac{3}{2}\mu_{T_2}\sqrt{s_{\Lambda^{-1}}}\|v\|.
    \end{align*}
    Therefore,
    \begin{align*}
        &\|v\|\leq \frac{3\sqrt{s_{\Lambda^{-1}}}}{\lambda_0}\mu_{T_2} \\
        &\|v\|_{\ell 1}\leq \frac{12s_{\Lambda^{-1}}}{\lambda_0}\mu_{T_2}.      
    \end{align*}
\end{proof}

\subsection{Proof of Lemma \ref{lemma: sgd}}
\begin{proof}
    Define $\{e_l\}_{l=1}^d$ as the canonical basis of $\RR^d$. Since
    \begin{align*}
        g_t&=2\widehat{\Sigma}_t\widehat{\beta}_{t-1}-\frac{2}{t}\sum_{\tau=1}^t\frac{\indicator(a_{\tau}=1)}{\pi_{\tau}}X_{\tau}y_{\tau} \\
        &=\frac{2}{t}\sum_{\tau=1}^t\frac{\indicator(a_{\tau}=1)}{\pi_{\tau}}X_{\tau}X_{\tau}^{\top}(\widehat{\beta}_{t-1}-\beta)-\frac{2}{t}\sum_{\tau=1}^t\frac{\indicator(a_{\tau}=1)}{\pi_{\tau}}X_{\tau}\xi_{\tau} \\
        &= 2\widehat{\Sigma}_t(\widehat{\beta}_{t-1}-\beta)-\frac{2}{t}\sum_{\tau=1}^t\frac{\indicator(a_{\tau}=1)}{\pi_{\tau}}X_{\tau}\xi_{\tau},
    \end{align*}
    we have
    \begin{align*}
        |\inp{g_t-\nabla f(\widehat{\beta}_{t-1})}{e_l}| &= \left|\inp{2(\widehat{\Sigma}_t-\Sigma)(\widehat{\beta}_{t-1}-\beta)-\frac{2}{t}\sum_{\tau=1}^t\frac{\indicator(a_{\tau}=1)}{\pi_{\tau}}X_{\tau}\xi_{\tau}}{e_l}\right| \\
        &\leq \underbrace{2\left|\inp{(\widehat{\Sigma}_t-\Sigma)(\widehat{\beta}_{t-1}-\beta)}{e_l}\right|}_{(1)} + \underbrace{2\left|\frac{1}{t}\sum_{\tau=1}^t\frac{\indicator(a_{\tau}=1)}{\pi_{\tau}}X_{\tau}\xi_{\tau}\right|}_{(2)}.
    \end{align*}
    Notice that $\widehat{\beta}_{t-1}-\beta$ is at most $2s$-sparse, $(1)$ can be bounded by
    \begin{align*}
        (1)&\leq 2\max_{l,j\in [d]} |\widehat{\Sigma}_{t(lj)}-\Sigma_{(lj)}|\|\widehat{\beta}_{t-1}-\beta\|_{\ell 1} \\
        &\leq 2\sqrt{2s} \max_{l,j\in [d]} \left|\frac{1}{t}\sum_{\tau=1}^t\frac{\indicator(a_{\tau}=1)}{\pi_{\tau}}X_{\tau(l)}X_{\tau(j)}^{\top} - \Sigma_{lj}\right|\|\widehat{\beta}_{t-1}-\beta\|
    \end{align*}
    where $\widehat{\Sigma}_{t(lj)}$ and $\Sigma_{(lj)}$ represent the $(l,j)$-th entry of $\widehat{\Sigma}$ and $\Sigma$ respectively. By martingale Bernstein inequality, it is easy to derive with probability at least $1-d^{-10}$,
    \begin{align*}
         \left|\frac{1}{t}\sum_{\tau=1}^t\frac{\indicator(a_{\tau}=1)}{\pi_{\tau}}X_{\tau(l)}X_{\tau(j)}^{\top} - \Sigma_{lj}\right|\lesssim \frac{D^2\log^{1/2} d}{t^{(1-\gamma)/2}} + \frac{D^2\log d}{t^{1-\gamma}} \lesssim \frac{D^2\log d}{t^{(1-\gamma)/2}}
    \end{align*}
    for any $\gamma\in [0,1)$. 
    The upper bound for (2) can also be derived by martingale Bernstein inequality. With probability at least $1-d^{-10}$, 
    \begin{align*}
        \max_{l\in [d]} \left|\frac{1}{t}\sum_{\tau=1}^t\frac{\indicator(a_{\tau}=1)}{\pi_{\tau}}X_{\tau(l)}\xi_{\tau}\right| \lesssim  \frac{\sigma D\log^{1/2} d}{t^{(1-\gamma)/2}} + \frac{D\log^{3/2} d}{t^{1-\gamma}} \lesssim \frac{\sigma D\log d}{t^{(1-\gamma)/2}}.
    \end{align*}
    Then we finish the proof. 
\end{proof}

\subsection{Proof of Lemma \ref{lemma:bias1}}
\begin{proof}
    Recall that, for any $l\in [d]$,
    \begin{align*}
        \|R_1+R_2\|_{\max}&\leq \underbrace{\left\|\frac{1}{\sqrt{T^{1+\gamma}}}e_l^{\top}\sum_{t=1}^{T} \frac{\indicator(a_t=1)}{\pi_t}(M_T-\Omega) X_t\xi_t\right\|_{\max}}_{R_1} \\ &\quad + \frac{1}{\sqrt{T^{1+\gamma}}}\underbrace{\left\|e_l^{\top}\sum_{t=1}^{T} \left(I-\frac{\indicator(a_t=1)}{\pi_t}M_TX_tX_t^{\top}\right)(\widehat{\beta}_{T} - \beta)\right\|_{\max}}_{R_2}.
    \end{align*}
    It is obvious that by KKT condition of (\ref{eq:optimization1}),
    \begin{align}\label{eq:kkt1}
        \|\widehat{\Sigma}_{T}m_{l,T} - e_l\|_{\max}\leq \mu_{T_1}.
    \end{align}
    As a result,
    \begin{align*}
        \|R_2\|_{\max}&\leq \sqrt{T^{1-\gamma}}\left\|e_l^{\top} \left(I-M_T\widehat{\Sigma}_T\right)\right\|_{\max}\|\widehat{\beta}_{T} - \beta\|_{\ell 1} \\
        &\lesssim \sqrt{T^{1-\gamma}}\sqrt{\frac{\log d}{T}}s_0\sqrt{\frac{\log d}{T^{1-\gamma}}}\leq s_0\sqrt{\frac{\log^2 d}{T}}.
    \end{align*}
    As long as $s_0=o(\sqrt{\frac{T}{\log^2 d}})$, $\|R_2\|_{\max}=o(1)$.

    Next we analysis $R_1$. Since $\EE\left[\frac{\indicator(a_t=1)}{\pi_t}e_l^{\top}(M_T-\Omega) X_t\xi_t|\mathcal{F}_{t-1}\right]=0$ for any $t$, and the uniform bound is
    \begin{align*}
        \left|\frac{1}{\sqrt{T^{1+\gamma}}}\frac{\indicator(a_t=1)}{\pi_t}e_l^{\top}(M_T-\Omega) X_t\xi_t\right|\lesssim \frac{t^{\gamma}}{\sqrt{T^{1+\gamma}}}\|e_l^{\top}(M_T-\Omega)\|_{\ell 1}\sigma\log^{1/2}d\lesssim \frac{s_{\Omega}}{\phi_{\min}(8s_{\Omega})}\sqrt{\frac{\log^2 d}{T^{2-\gamma}}}.
    \end{align*}
    Moreover, the conditional variance is
    \begin{align*}
        &\frac{1}{T^{1+\gamma}}\sum_{t=1}^{T}\EE\left[\frac{\indicator(a_t=1)}{\pi_t^2}(e_l^{\top}(M_T-\Omega) X_t)^2\xi_t^2|\mathcal{F}_{t-1}\right] \\
        &\quad \leq \frac{1}{T^{1+\gamma}}\|e_l^{\top}(M_T-\Omega)\|_{\ell 1}^2\sum_{t=1}^{T}\|X_t\|_{\max}^2\EE\left[\frac{\indicator(a_t=1)}{\pi_t^2}\xi_t^2|\mathcal{F}_{t-1}\right] \\
        &\quad \lesssim \frac{1}{T^{1+\gamma}}\frac{s_\Omega^2}{\phi_{\min}^2(8s_{\Omega})}\frac{\log d}{T}\sum_{t=1}^{T}t^{\gamma}\sigma^2 \lesssim s_\Omega^2\frac{\log d}{T}.
    \end{align*}
    By martingale Bernstein inequality, with probability at least $1-d^{-10}$, 
    \begin{align*}
        \|R_1\|_{\max}\lesssim s_\Omega\sqrt{\frac{\log^2 d}{T}} + s_{\Omega}\sqrt{\frac{\log^4 d}{T^{2-\gamma}}}.
    \end{align*}
    Since $\gamma\leq 1$, as long as $s_{\Omega}=o(\sqrt{\frac{T}{\log^2 d}})$, $\|R_1\|_{\max}=o(1)$.
\end{proof}

\subsection{Proof of Lemma \ref{lambdaconvergence1}}
\begin{proof}
    Without loss of generality, we only prove $i=1$ case. We have
    \begin{align*}
        \Lambda_{1,t}=\EE\left[\indicator(a_t=1)X_tX_t^{\top}|\mathcal{F}_{t-1}\right]&=\EE\left[X_tX_t^{\top}\indicator(\widehat{\beta}_{1,t-1} - \widehat{\beta}_{0,t-1}>0)|\mathcal{F}_{t-1}\right] \\
        &= \EE\left[X_tX_t^{\top}\indicator\left((\inp{\widehat{\beta}_{1,t-1} - \widehat{\beta}_{0,t-1}}{X_t}>0)\cap (\inp{\beta_1-\beta_0}{X_t}>0)\right)|\mathcal{F}_{t-1}\right] \\ &\quad + \EE\left[X_tX_t^{\top}\indicator\left((\inp{\widehat{\beta}_{1,t-1} - \widehat{\beta}_{0,t-1}}{X_t}>0)\cap (\inp{\beta_1-\beta_0}{X_t}<0)\right)|\mathcal{F}_{t-1}\right] 
    \end{align*}
    and
    \begin{align*}
        \Lambda_1&=\EE\left[\indicator(\inp{\beta_1-\beta_0}{X}>0)XX^{\top}|\mathcal{F}_{t-1}\right] \\
        &= \EE\left[XX^{\top}\indicator\left((\inp{\widehat{\beta}_{1,t-1} - \widehat{\beta}_{0,t-1}}{X}>0)\cap (\inp{\beta_1-\beta_0}{X}>0)\right)|\mathcal{F}_{t-1}\right] \\ &\quad + \EE\left[XX^{\top}\indicator\left((\inp{\widehat{\beta}_{1,t-1} - \widehat{\beta}_{0,t-1}}{X}<0)\cap (\inp{\beta_1-\beta_0}{X}>0)\right)|\mathcal{F}_{t-1}\right]. 
    \end{align*}
    We first prove a key Lemma.
    \begin{Lemma}
        \label{probright}
        Denote $\Delta_{X_t}=\inp{\beta_1-\beta_0}{X_t}$ and $\widehat{\Delta}_{X_t}=\inp{\widehat{\beta}_{1,t-1}-\widehat{\beta}_{0,t-1}}{X_t}$. Under Assumption \ref{assump:basic}, the event that successfully choosing the optimal action, i.e. $\{a_t=a^{*}(X_t)\}$ happens if and only if
        \begin{align*}
            |\Delta_{X_t}|>|\Delta_{X_t} - \widehat{\Delta}_{X_t}|.
        \end{align*}
        As a result,
        \begin{align*}
            \PP\left(\indicator(a_t\neq a^{*}(X_t))|\mathcal{F}_{t-1}\right)=\PP\left(|\Delta_{X_t}|\leq |\Delta_{X_t} - \widehat{\Delta}_{X_t}||\mathcal{F}_{t-1}\right).
       \end{align*} 
    \end{Lemma}
    Then for any $l,j\in [d]$, 
    \begin{align*}
        |e_l^{\top}(\Lambda_{1,t}-\Lambda)e_j|&\leq \EE\left[e_l^{\top}XX^{\top}e_j\indicator\left((\inp{\widehat{\beta}_{1,t-1} - \widehat{\beta}_{0,t-1}}{X}>0)\cap (\inp{\beta_1-\beta_0}{X}<0)\right)|\mathcal{F}_{t-1}\right] \\ &\quad + \EE\left[e_l^{\top}XX^{\top}e_j\indicator\left((\inp{\widehat{\beta}_{1,t-1} - \widehat{\beta}_{0,t-1}}{X}<0)\cap (\inp{\beta_1-\beta_0}{X}>0)\right)|\mathcal{F}_{t-1}\right] \\
        &\leq D^2\PP\left((\inp{\widehat{\beta}_{1,t-1} - \widehat{\beta}_{0,t-1}}{X}>0)\cap (\inp{\beta_1-\beta_0}{X}<0)|\mathcal{F}_{t-1}\right) \\ &\quad + D^2\PP\left((\inp{\widehat{\beta}_{1,t-1} - \widehat{\beta}_{0,t-1}}{X}<0)\cap (\inp{\beta_1-\beta_0}{X}>0)|\mathcal{F}_{t-1}\right) \\
        &\leq 4D^2\PP\left(a_t\neq a^{*}(X_t)|\mathcal{F}_{t-1}\right)\lesssim \left(\frac{s_0\log d}{(t-1)^{1-\gamma}}\right)^{\nu/2},
    \end{align*}
    where the last inequality comes from Lemma \ref{probright} and Assumption \ref{assump:margincondition}, that holds with probability at least $1-d^{-10}$. Then we prove the desired result.  

    Following the same arguments, for any $l\in [d]$,
    \begin{align*}
        |m_l^{\top}(\Lambda_{1,t}-\Lambda)m_l|&\leq \EE\left[m_l^{\top}XX^{\top}m_l\indicator\left((\inp{\widehat{\beta}_{1,t-1} - \widehat{\beta}_{0,t-1}}{X}>0)\cap (\inp{\beta_1-\beta_0}{X}<0)\right)|\mathcal{F}_{t-1}\right] \\ &\quad + \EE\left[m_l^{\top}XX^{\top}m_l\indicator\left((\inp{\widehat{\beta}_{1,t-1} - \widehat{\beta}_{0,t-1}}{X}<0)\cap (\inp{\beta_1-\beta_0}{X}>0)\right)|\mathcal{F}_{t-1}\right] \\
        &\leq |m_l^{\top}X_t|^2\PP\left((\inp{\widehat{\beta}_{1,t-1} - \widehat{\beta}_{0,t-1}}{X}>0)\cap (\inp{\beta_1-\beta_0}{X}<0)|\mathcal{F}_{t-1}\right) \\ &\quad + |m_l^{\top}X_t|^2\PP\left((\inp{\widehat{\beta}_{1,t-1} - \widehat{\beta}_{0,t-1}}{X}<0)\cap (\inp{\beta_1-\beta_0}{X}>0)|\mathcal{F}_{t-1}\right) \\
        &\leq 4\|m_l\|^2\phi_{\max}^2(s_{\Omega})\log d\PP\left(a_t\neq a^{*}(X_t)|\mathcal{F}_{t-1}\right)\lesssim \left(\frac{s_0\log^3 d}{(t-1)^{1-\gamma}}\right)^{\nu/2},
    \end{align*}
    where the second last inequality comes from Assumption \ref{assump:basic}(c), and $\|m_l\|\leq C_{\min}$ is bounded.
\end{proof}

\subsection{Proof of Lemma \ref{lemma:bias2}}
\begin{proof}
    This proof follows almost the same arguments as the proof of Lemma \ref{lemma:bias1}. Recall that, for any $l\in [d]$,
    \begin{align*}
        \|R_1+R_2\|_{\max}&\leq \frac{1}{\sqrt{T}}\underbrace{\left\|e_l^{\top}\sum_{t=1}^{T} \indicator(a_t=1)(M_T-\Lambda^{-1}) X_t\xi_t\right\|_{\max}}_{R_1} \\ &\quad + \frac{1}{\sqrt{T}}\underbrace{\left\|e_l^{\top}\sum_{t=1}^{T} \left(I-\indicator(a_t=1)M_TX_tX_t^{\top}\right)(\widehat{\beta}_{T} - \beta)\right\|_{\max}}_{R_2}.
    \end{align*}
    It is obvious that by KKT condition of (\ref{eq:optimization2}),
    \begin{equation}
        \|\widehat{\Lambda}_{i,T}m_{l}^{(i)} - e_l\|_{\max}\leq \mu_{T_2} \label{kkt2}.
    \end{equation}
    As a result,
    \begin{align*}
        \|R_2\|_{\max}&\leq \sqrt{T}\left\|e_l^{\top} \left(I-M_T\widehat{\Lambda}\right)\right\|_{\max}\|\widehat{\beta}_{T} - \beta\|_{\ell 1} \\
        &\lesssim \sqrt{T}\left(\frac{s_0\log d}{T}\right)^{\nu/2}s_0\sqrt{\frac{\log d}{T}}\leq \sqrt{\frac{s_0^{2+\nu}\log^{1+\nu} d}{T^\nu}}.
    \end{align*}
    As long as $s_0=o(\frac{T^{\nu/(2+\nu)}}{\log d})$, $\|R_2\|_{\max}=o(1)$.

    Next we analysis $R_1$. Since $\EE\left[\indicator(a_t=1)e_l^{\top}(M_T-\Lambda^{-1}) X_t\xi_t|\mathcal{F}_{t-1}\right]=0$ for any $t$, and the uniform bound is
    \begin{align*}
        \left|\frac{1}{\sqrt{T}}\indicator(a_t=1)e_l^{\top}(M_T-\Lambda^{-1}) X_t\xi_t\right|\lesssim \frac{1}{\sqrt{T}}\|e_l^{\top}(M_T-\Lambda^{-1})\|_{\ell 1}\sigma\log^{1/2}d\lesssim \frac{s_{\Lambda^{-1}}}{\lambda_0}\sqrt{\frac{s_0^{\nu}\log^{1+\nu} d}{T^{1+\nu}}}.
    \end{align*}
    Moreover, the conditional variance is
    \begin{align*}
        &\frac{1}{T}\sum_{t=1}^{T}\EE\left[\indicator(a_t=1)(e_l^{\top}(M_T-\Lambda^{-1}) X_t)^2\xi_t^2|\mathcal{F}_{t-1}\right] \\
        &\quad \leq \frac{1}{T}\|e_l^{\top}(M_T-\Lambda^{-1})\|_{\ell 1}^2\sum_{t=1}^{T}\|X_t\|_{\max}^2\EE\left[\indicator(a_t=1)\xi_t^2|\mathcal{F}_{t-1}\right] \\
        &\quad \lesssim \frac{1}{T}\frac{s_{\Lambda^{-1}}^2s_0^{\nu}}{\phi_{\min}^2(8s_{\Lambda^{-1}})}\frac{\log^{\nu} d}{T^{\nu}}T\sigma^2 \lesssim s_{\Lambda^{-1}}^2s_0^{\nu}\frac{\log^{\nu} d}{T^{\nu}}.
    \end{align*}
    By martingale Bernstein inequality, with probability at least $1-d^{-10}$, 
    \begin{align*}
        \|R_1\|_{\max}\lesssim  s_{\Lambda^{-1}}\sqrt{\frac{s_0^{\nu}\log^{3+\nu} d}{T^{1+\nu}}} + \sqrt{s_{\Lambda^{-1}}^2s_0^{\nu}\frac{\log^{\nu+1} d}{T^{\nu}}}.
    \end{align*}
    Then as long as $s_{\Lambda^{-1}}^2s_0^{\nu}=o(\frac{T^{\nu}}{\log d})$, $\|R_1\|_{\max}=o(1)$.
\end{proof}

\subsection{Proof of Lemma \ref{lambdaconvergence2}}
\begin{proof}
    This proof follows almost the same arguments as the proof of Lemma \ref{lambdaconvergence1}. The only difference is the probability of $\{a_t\neq a^{*}(X_t)\}$. Recall in the proof of Lemma \ref{lambdaconvergence1}, we have derived for any $l,j\in [d]$, 
    \begin{align*}
        |e_l^{\top}(\Lambda_{1,t}-\Lambda)e_j|\leq  4D^2\PP\left(a_t\neq a^{*}(X_t)|\mathcal{F}_{t-1}\right)
    \end{align*}
    and 
    \begin{align*}
        |m_l^{\top}(\Lambda_{1,t}-\Lambda)m_l|\leq 4\|m_l\|^2\phi_{\max}^2(s_{\Omega})\log d\PP\left(a_t\neq a^{*}(X_t)|\mathcal{F}_{t-1}\right).
    \end{align*}
    By Assumption \ref{assump:margincondition} and Theorem \ref{thm:estimation-dc}, $\PP\left(a_t\neq a^{*}(X_t)|\mathcal{F}_{t-1}\right)\lesssim (\frac{s_0\log d}{t-1})^{\nu/2}$. Then $|m_l^{\top}(\Lambda_{1,t}-\Lambda)m_l|\lesssim \phi_{\max}^2(s_{\Lambda^{-1}})(\frac{s_0\log^3 d}{t-1})^{\nu/2}$ and $|e_l^{\top}(\Lambda_{1,t}-\Lambda)e_j|\lesssim (\frac{s_0\log d}{t-1})^{\nu/2}$. 
\end{proof}

\subsection{Proof of Lemma \ref{lemma:neg1}}
\begin{proof}
    Without loss of generality, we only prove $i=1$ case and omit all the subscripts and superscripts $i$. Notice that
    \begin{align*}
        \left\|M_{T}\widehat{\Lambda} - M_{T}\Lambda \right\|_{\max}\leq \left\|(M_{T}-\Lambda^{-1})(\widehat{\Lambda}_T -\Lambda)\right\|_{\max} + \left\|\Lambda^{-1}(\widehat{\Lambda} -\Lambda)\right\|_{\max}.
    \end{align*}
    Note that $\|\widehat{\Lambda}_T -\Lambda\|_{\max}\leq \|\widehat{\Lambda}_T - \frac{1}{T}\sum_{t=1}^{T}\Lambda_t \|_{\max} + \|\frac{1}{T}\sum_{t=1}^{T}\Lambda_t -\Lambda\|_{\max}$, where $\Lambda_t=\EE[\indicator(a_t=1)X_tX_t^{\top}|\mathcal{F}_{t-1}]$.
    
    By Lemma \ref{lambdaconvergence2}, $\|\frac{1}{T}\sum_{t=1}^{T}\Lambda_t -\Lambda\|_{\max}\lesssim \sqrt{\frac{s_0\log d}{T}}$. To prove the upper bound for $\|\widehat{\Lambda} - \frac{1}{T}\sum_{t=1}^{T}\Lambda_k \|_{\max}$, note that for any $l,j\in [d]$, $|\indicator(a_t=1)e_l^{\top}X_tX_t^{\top}e_j|\leq D^2$ and $\EE[\indicator(a_t=1)(e_l^{\top} X_tX_t^{\top}e_j)^2|\mathcal{F}_{t-1}]\leq D^4$. Then by Bernstein inequality, $\|\widehat{\Lambda}_T - \frac{1}{T}\sum_{t=1}^{T}\Lambda_t \|_{\max}\lesssim \sqrt{\frac{\log d}{T}}$ with probability at least $1-d^{-10}$. 
    
    Combine with Lemma \ref{lemma:mlconvergence2}, $\|(M_T-\Lambda^{-1})(\widehat{\Lambda}_T -\Lambda)\|_{\max}\leq \|M_{T}-\Lambda^{-1}\|_{\infty}\|\widehat{\Lambda}_T -\Lambda\|_{\max}\lesssim \frac{s_0^{\nu}s_{\Lambda^{-1}}\log^{2\nu} d}{T^\nu} + \frac{s_{\Lambda^{-1}}s_0^{\nu/2}\log^{\nu} d}{T}\lesssim \frac{s_0^{\nu}s_{\Lambda^{-1}}\log^{2\nu} d}{T^\nu}$. 
    
    Moreover, by Lemma \ref{lemma:feasible2},  $\|I-\Lambda^{-1}\widehat{\Lambda}_T\|_{\max}\lesssim \left(\frac{s_0\log d}{T}\right)^{\nu/2}$. Then we prove the desired result.
\end{proof}

\subsection{Proof of Lemma \ref{lambdaconvergence3}}
\begin{proof}
    The proof follows a similar argument with Lemma \ref{lambdaconvergence1}. Without loss of generality, we only prove $i=1$ case and omit all the subscript. We have
    \begin{align*}
        \Lambda^{-1}\Lambda_{t}&=\EE\left[\indicator(a_t=1)\Lambda^{-1}X_tX_t^{\top}|\mathcal{F}_{t-1}\right]=\EE\left[\Lambda^{-1}X_tX_t^{\top}\indicator(\widehat{\beta}_{1,t-1} - \widehat{\beta}_{0,t-1}>0)|\mathcal{F}_{t-1}\right] \\
        &= \EE\left[\Lambda^{-1}X_tX_t^{\top}\indicator\left((\inp{\widehat{\beta}_{1,t-1} - \widehat{\beta}_{0,t-1}}{X_t}>0)\cap (\inp{\beta_1-\beta_0}{X_t}>0)\right)|\mathcal{F}_{t-1}\right] \\ &\quad + \EE\left[\Lambda^{-1}X_tX_t^{\top}\indicator\left((\inp{\widehat{\beta}_{1,t-1} - \widehat{\beta}_{0,t-1}}{X_t}>0)\cap (\inp{\beta_1-\beta_0}{X_t}<0)\right)|\mathcal{F}_{t-1}\right]. 
    \end{align*}
    and
    \begin{align*}
        I=\Lambda^{-1}\Lambda&=\Lambda^{-1}\EE\left[\indicator(\inp{\beta_1-\beta_0}{X}>0)XX^{\top}|\mathcal{F}_{t-1}\right] \\
        &= \EE\left[\Lambda^{-1}XX^{\top}\indicator\left((\inp{\widehat{\beta}_{1,t-1} - \widehat{\beta}_{0,t-1}}{X}>0)\cap (\inp{\beta_1-\beta_0}{X}>0)\right)|\mathcal{F}_{t-1}\right] \\ &\quad + \EE\left[\Lambda^{-1}XX^{\top}\indicator\left((\inp{\widehat{\beta}_{1,t-1} - \widehat{\beta}_{0,t-1}}{X}<0)\cap (\inp{\beta_1-\beta_0}{X}>0)\right)|\mathcal{F}_{t-1}\right]. 
    \end{align*}
    Then for any $l,j\in [d]$ and $t\geq T_0$, 
    \begin{align*}
        |e_l^{\top}(I-\Lambda^{-1}\Lambda_t)e_j|&\leq \EE\left[e_l^{\top}\Lambda^{-1}XX^{\top}e_j\indicator\left((\inp{\widehat{\beta}_{1,t-1} - \widehat{\beta}_{0,t-1}}{X}>0)\cap (\inp{\beta_1-\beta_0}{X}<0)\right)|\mathcal{F}_{t-1}\right] \\ &\quad + \EE\left[e_l^{\top}\Lambda^{-1}XX^{\top}e_j\indicator\left((\inp{\widehat{\beta}_{1,t-1} - \widehat{\beta}_{0,t-1}}{X}<0)\cap (\inp{\beta_1-\beta_0}{X}>0)\right)|\mathcal{F}_{t-1}\right] \\
        &\leq |e_l^{\top}\Lambda^{-1}X||X^{\top}e_j|\PP\left((\inp{\widehat{\beta}_{1,t-1} - \widehat{\beta}_{0,t-1}}{X}>0)\cap (\inp{\beta_1-\beta_0}{X}<0)|\mathcal{F}_{t-1}\right) \\ &\quad + |e_l^{\top}\Lambda^{-1}X||X^{\top}e_j|\PP\left((\inp{\widehat{\beta}_{1,t-1} - \widehat{\beta}_{0,t-1}}{X}<0)\cap (\inp{\beta_1-\beta_0}{X}>0)|\mathcal{F}_{t-1}\right) \\
        &\leq \phi_{\max}(s_{\Lambda^{-1}})\|e_l^{\top}\Lambda^{-1}\|\log^{1/2}dD\PP\left(a_t\neq a^{*}(X_t)|\mathcal{F}_{t-1}\right)\lesssim \left(\frac{s_0\log^{2} d}{t-1}\right)^{\nu/2},
    \end{align*}
    where the second last inequality comes from Assumption \ref{assump:basic}(c) and the last inequality comes from Lemma \ref{probright} and Assumption \ref{assump:margincondition}. Then we prove the desired result. 
\end{proof}

\subsection{Proof of Lemma \ref{probright}}
\begin{proof}
    We first prove the first claim. Notice that
    \begin{align*}
        \indicator(\widehat{\Delta}_{X_t}>0)=\indicator(\widehat{\Delta}_{X_t} + \Delta_{X_t} - \Delta_{X_t}>0)= \indicator(\Delta_{X_t} > \Delta_{X_t} - \widehat{\Delta}_{X_t}),
    \end{align*}
    which implies
    \begin{align*}
        \indicator(\widehat{\Delta}_{X_t}>0)=\indicator(\Delta_{X_t}>0)
    \end{align*}
    if and only if
    \begin{align*}
        |\Delta_{X_t}|>|\Delta_{X_t} - \widehat{\Delta}_{X_t}|.
    \end{align*}
   As a result,
   \begin{align*}
        \PP\left(\indicator(a_t\neq a^{*}(X_t))|\mathcal{F}_{t-1}\right)&= \PP\left(\indicator(\widehat{\Delta}_{X_t}>0)\neq \indicator(\Delta_{X_t}>0)|\mathcal{F}_{t-1}\right)\\
        &= \PP\left(|\Delta_{X_t}|\leq |\Delta_{X_t} - \widehat{\Delta}_{X_t}||\mathcal{F}_{t-1}\right).
   \end{align*} 
   Then we proved the second claim.
\end{proof}

\end{document}